\documentclass[10pt,journal,compsoc]{IEEEtran}

\usepackage{times}
\usepackage{epsfig}
\usepackage{graphicx}
\usepackage{amsmath}
\usepackage{amssymb}
\usepackage{rotating}
\usepackage{multirow}
\usepackage{lipsum}
\usepackage{enumitem}
\usepackage{algorithm}
\usepackage{algorithmicx}
\usepackage[noend]{algpseudocode}
\usepackage{color}
\usepackage{longtable}  % to split table across multiple pages

\usepackage[T1]{fontenc}
\usepackage[a]{esvect}
\usepackage{amsmath}
\usepackage{xcolor}     % for colour
\usepackage{ntheorem}
\usepackage{mdframed}

\newcommand{\parallelsum}{\mathbin{\!/\mkern-5mu/\!}}

\theoremstyle{break}
\theoremheaderfont{\bfseries}
\newmdtheoremenv[%
linecolor=gray,leftmargin=20,%
rightmargin=20,
backgroundcolor=gray!20,%
innertopmargin=5pt,%
ntheorem]{observation}{Observation}

\ifCLASSOPTIONcompsoc
  \usepackage[nocompress]{cite}
\else
  \usepackage{cite}
\fi

\usepackage{fancybox}
\usepackage{amsfonts,amsmath,amssymb,latexsym}
\usepackage{array}
\usepackage{graphicx}

\newenvironment{separate}{\begin{quote}\begin{sffamily}\footnotesize}{\end{sffamily}\end{quote}}

\newcommand{\beq}{\begin{equation}}
\newcommand{\eeq}{\end{equation}}
\newcommand{\beqa}{\begin{eqnarray}}
\newcommand{\eeqa}{\end{eqnarray}}
\newcommand{\beqas}{\begin{eqnarray*}}
\newcommand{\eeqas}{\end{eqnarray*}}

\newcommand{\tightlist}{\setlength{\itemsep}{0pt}
\setlength{\parsep}{0pt} \setlength{\parskip}{0pt}}

\newlength{\boxwidth}
\newlength{\narrowboxwidth}
\usepackage{upquote}

\usepackage[pagebackref=true,breaklinks=true,colorlinks,bookmarks=false]{hyperref}

\newcommand{\drive}{\texttt{DR(eye)VE}}
\newcommand{\etal}{~\emph{et al.}}
\newcommand{\quotes}[1]{``#1''}

\newcommand\given[1][]{\:#1\vert\:}

\begin{document}

%%%%%%%%% TITLE
\title{Predicting the Driver's Focus of Attention:\\ the DR(eye)VE Project}

\author{Andrea Palazzi$^\ast$, Davide Abati$^\ast$, Simone Calderara, Francesco Solera, %~\IEEEmembership{Student Member,~IEEE,} %~\IEEEmembership{Fellow,~OSA,}
        and~Rita~Cucchiara%,~\IEEEmembership{Life~Fellow,~IEEE}% <-this % stops a space
\IEEEcompsocitemizethanks{\IEEEcompsocthanksitem
All authors are with the Department
of Engineering ``Enzo Ferrari'', University of Modena and Reggio Emilia, Italy.\protect\\
E-mail: name.surname@unimore.it\protect\\
$^\ast$ indicates equal contribution.
}
%\IEEEcompsocthanksitem J. Doe and J. Doe are with Anonymous University.}% <-this % stops an unwanted space
\thanks{}
}
\IEEEtitleabstractindextext{%
\begin{abstract}
In this work we aim to predict the driver's focus of attention. The goal is to estimate what a person would pay attention to while driving, and which part of the scene around the vehicle is more critical for the task. To this end we propose a new computer vision model based on a multi-branch deep architecture that integrates three sources of information: raw video, motion and scene semantics. We also introduce \drive, the largest dataset of driving scenes for which eye-tracking annotations are available. This dataset features more than 500,000 registered frames, matching ego-centric views (from glasses worn by drivers) and car-centric views (from roof-mounted camera), further enriched by other sensors measurements. 
Results highlight that several attention patterns are shared across drivers and can be reproduced to some extent. 
The indication of which elements in the scene are likely to capture the driver's attention may benefit several applications in the context of human-vehicle interaction and driver attention analysis.
\end{abstract}

}

\maketitle
\IEEEdisplaynontitleabstractindextext
\IEEEpeerreviewmaketitle

\maketitle

%%%%%%%%%%%%%%%%%%%%%%%%%%%%%%%%%%%%%%%%%%%%%%%%%%%%%%%%%%%%%%%%%%%%%%
%%%%%%%%%%%%%%%%%%%%%% INTRODUCTION  %%%%%%%%%%%%%%%%%%%%%%%%%%%%%%%%%
%%%%%%%%%%%%%%%%%%%%%%%%%%%%%%%%%%%%%%%%%%%%%%%%%%%%%%%%%%%%%%%%%%%%%%

\section{Introduction}
According to the J3016 SAE international Standard, which defined the five levels of autonomous driving \cite{Grubm2017}, cars will provide a  fully autonomous journey only at the fifth level. At lower levels of autonomy, computer vision and other sensing systems will still support humans in the driving task. Human-centric Advanced Driver Assistance Systems (ADAS) have significantly improved safety and comfort in driving (\emph{e.g.}~collision avoidance systems, blind spot control, lane change assistance etc.). 
Among ADAS solutions, the most ambitious examples are related to monitoring systems~\cite{jain2015car,frolich2014will, kumar2013learning, morris2011lane}: they parse the attention behavior of the driver together with the road scene to predict potentially unsafe manoeuvres and act on the car in order to avoid them -- either by signaling the driver or braking.
However, all these approaches suffer from the complexity of capturing the true driver's attention and rely on a limited set of fixed safety-inspired rules.
Here, we shift the problem from a personal level (\emph{what the driver is looking at}) to a task-driven level (\emph{what most drivers would look at}) introducing a computer vision model able to to replicate the human attentional behavior during the driving task.

We achieve this result in two stages: First, we conduct a data-driven study on drivers' gaze fixations under different circumstances and scenarios. The study concludes that the semantic of the scene, the speed and bottom-up features all influence the driver's gaze. Second, we advocate for the existence of common gaze patterns that are shared among different drivers. We empirically demonstrate the existence of such patterns by developing a deep learning model that can profitably learn to predict where a driver would be looking at in a specific situation.\\
\begin{figure}[t]
    \centering
    \begin{tabular}{cc}
    \hspace{-0.2cm}\includegraphics[width=0.23\textwidth]{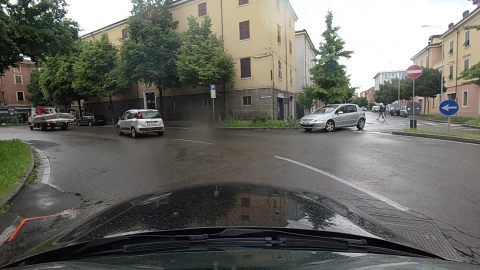} &\hspace{-0.3cm}\includegraphics[width=0.23\textwidth]{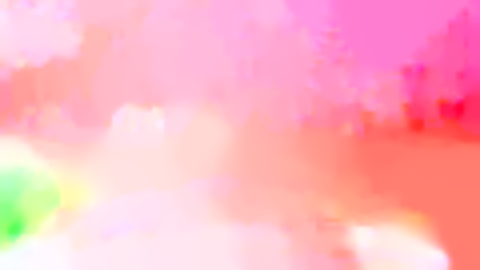}\\
    \hspace{-0.2cm}(a) RGB frame & \hspace{-0.3cm}(b) optical flow \\
    \hspace{-0.2cm}\includegraphics[width=0.23\textwidth]{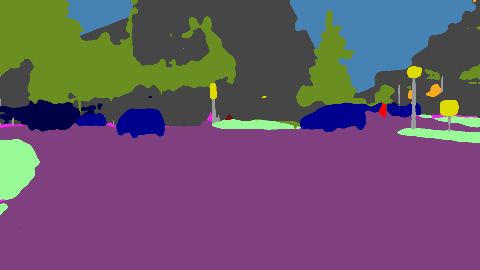} &\hspace{-0.3cm}\includegraphics[width=0.23\textwidth]{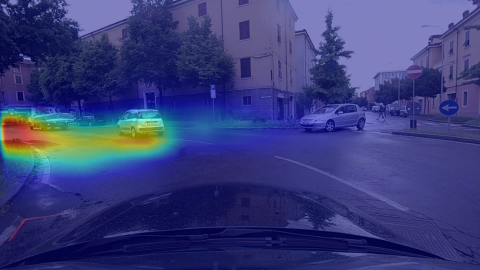}\\
    \hspace{-0.2cm}(c) semantic segmentation & \hspace{-0.3cm}(d) predicted map
    \end{tabular}
    \caption{An example of visual attention while driving (d), estimated from our deep model using (a) raw video, (b) optical flow and (c) semantic segmentation.}
    \label{fig:overview}
\end{figure}
To this aim we recorded and annotated 555,000 frames (approx. 6 hours) of driving sequences in different traffic and weather conditions: the \drive~dataset. For every frame we acquired the driver's gaze through an accurate eye tracking device and registered such data to the external view recorded from a roof-mounted camera.
The \drive~data richness enables us to train an end-to-end deep network that predicts salient regions in car-centric driving videos. The network we propose is based on three branches which estimate attentional maps from a) visual information of the scene, b) motion cues (in terms of optical flow) and c) semantic segmentation (Fig.~\ref{fig:overview}). 
In contrast to the majority of experiments, which are conducted in controlled laboratory settings or employ sequences of unrelated images~\cite{TJunct,mit-saliency-benchmark,jiang2015salicon}, we train our model on data acquired on the field.
Final results demonstrate the ability of the network to generalize across different day times, different weather conditions, different landscapes and different drivers. \\
Eventually, we believe our work can be complementary to the current semantic segmentation and object detection literature\cite{yu2015multi, wang2017understanding, nilsson2016semantic, 3dopNIPS15, mousavian20163d} by providing a diverse set of information.
%by providing the driver with a more balanced set of information that can help reducing his cognitive load.
According to~\cite{Tatler}, the act of driving combines complex attention mechanisms guided by the driver's past experience, short reactive times and strong contextual constraints. Thus, very little information is needed to drive if guided by a strong focus of attention (FoA) on a limited set of targets: our model aims at predicting them.
\\
\\
\noindent The paper is organized as follows. In Sec.~\ref{sec:related}, related works about computer vision and gaze prediction are provided to frame our work in the current state-of-the-art scenario. Sec.~\ref{sec:dataset} describes the \drive~dataset and some insights about several attention patterns that human drivers exhibit. Sec.~\ref{sec:models} illustrates the proposed deep network to replicate such human behavior, and Sec.~\ref{sec:exp1} reports the performed experiments.

%%%%%%%%%%%%%%%%%%%%%%%%%%%%%%%%%%%%%%%%%%%%%%%%%%%%%%%%%%%%%%%%%%%%%%
%%%%%%%%%%%%%%%%%%%%%%%%% RELATED WORK  %%%%%%%%%%%%%%%%%%%%%%%%%%%%%%
%%%%%%%%%%%%%%%%%%%%%%%%%%%%%%%%%%%%%%%%%%%%%%%%%%%%%%%%%%%%%%%%%%%%%%
\section{Related Work}
\label{sec:related}

\begin{figure*}[t]
\centering
\bgroup
\setlength{\tabcolsep}{.16667em}
\begin{tabular}{cccc}
\includegraphics[width=0.22\textwidth]{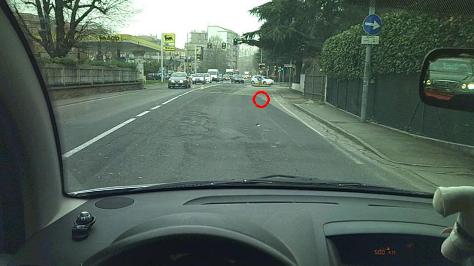} &
\includegraphics[width=0.22\textwidth]{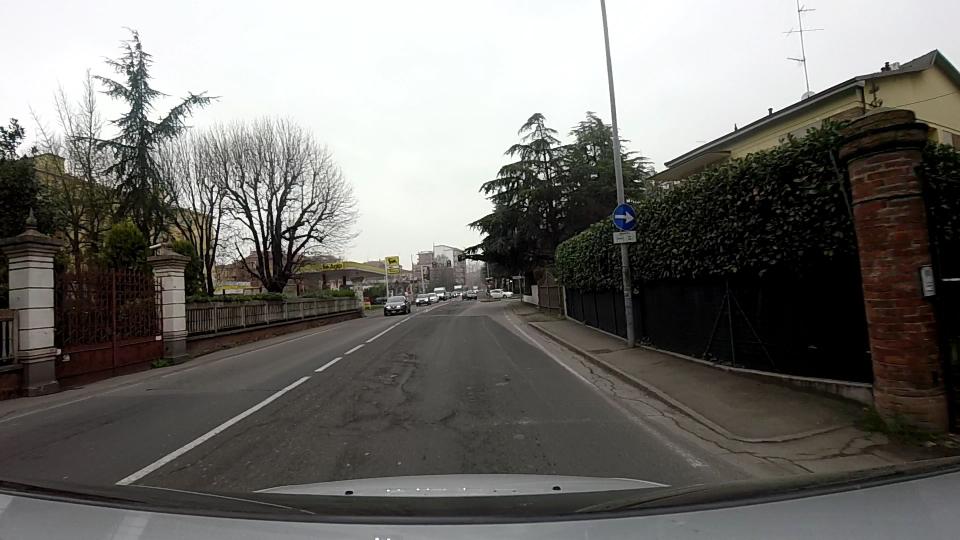} & 
\includegraphics[width=0.22\textwidth]{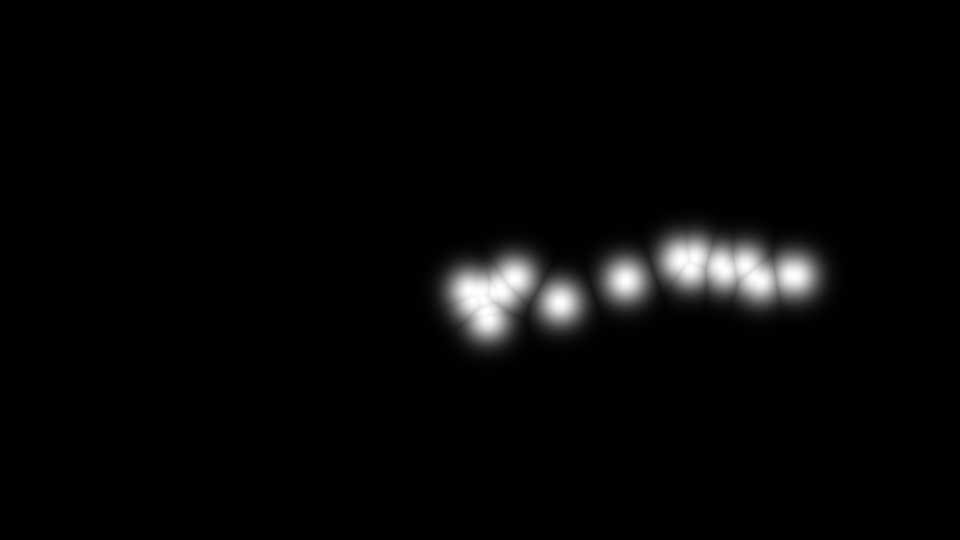} &
\includegraphics[width=0.22\textwidth]{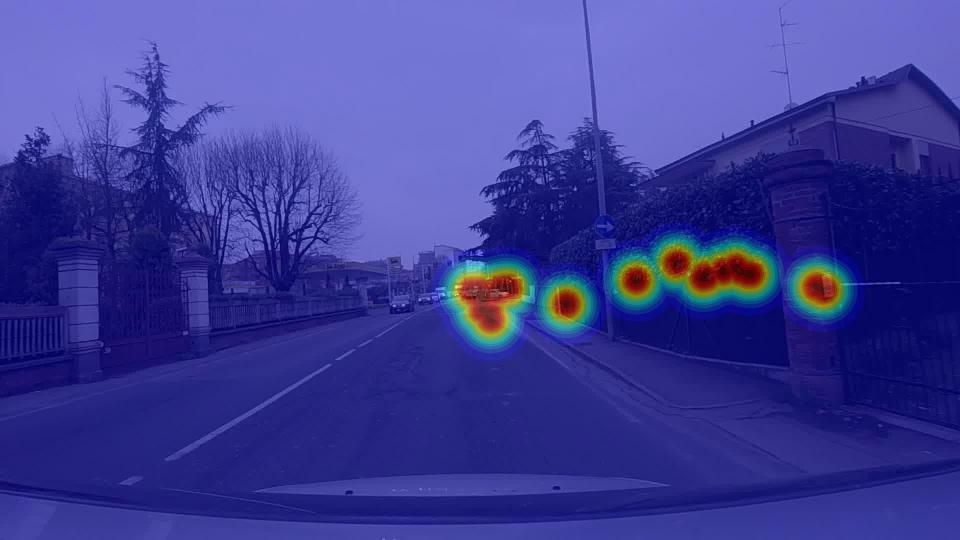}\\ 
\includegraphics[width=0.22\textwidth]{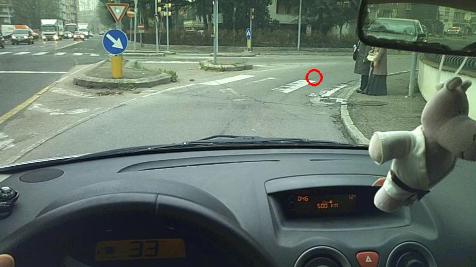} &
\includegraphics[width=0.22\textwidth]{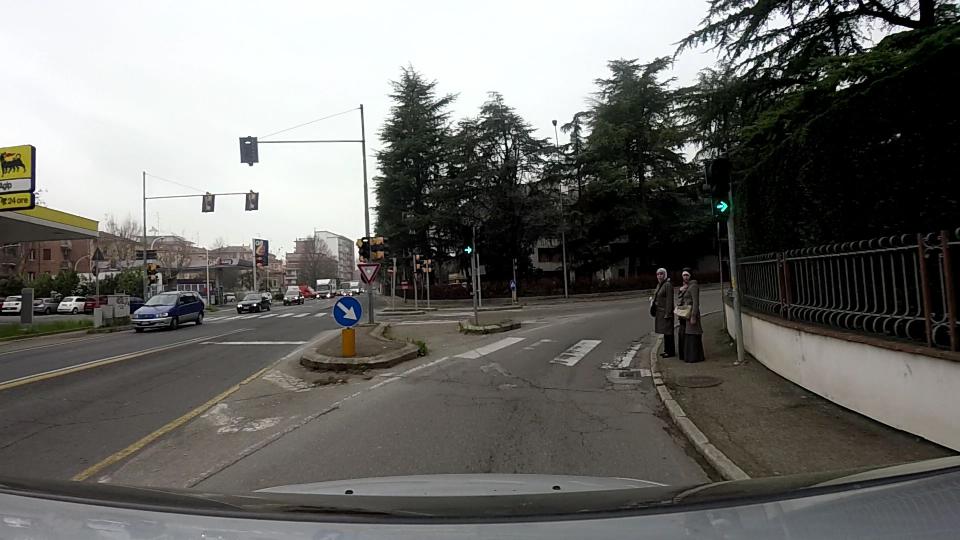} & 
\includegraphics[width=0.22\textwidth]{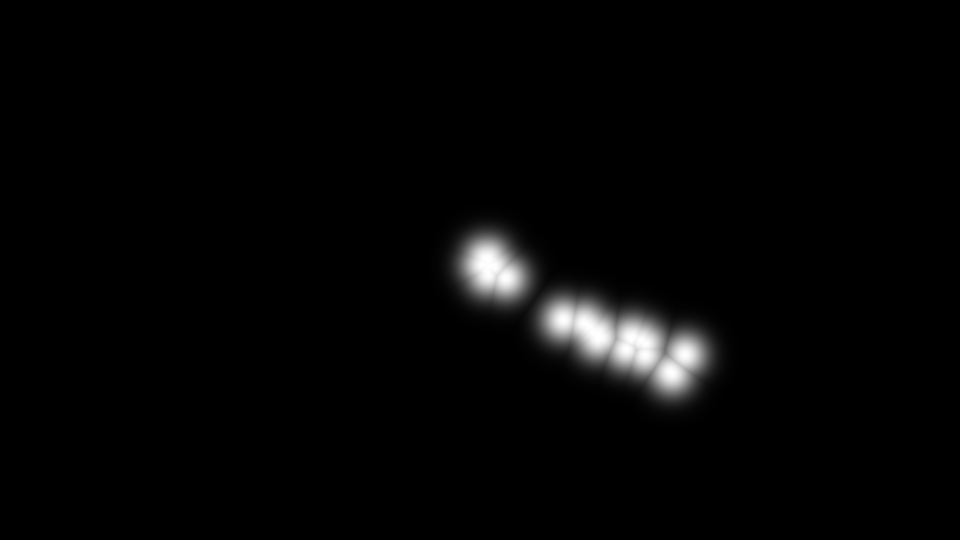} &
\includegraphics[width=0.22\textwidth]{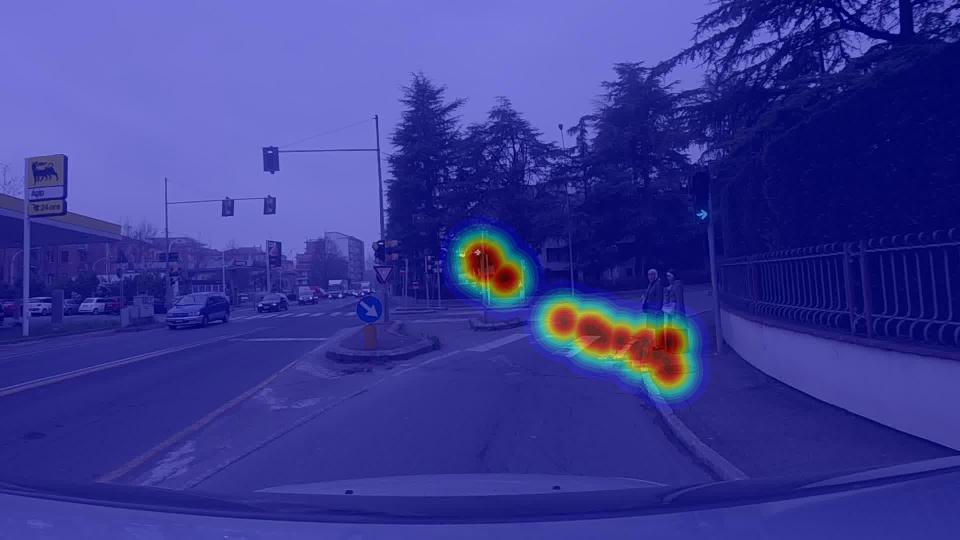}\\ 
\includegraphics[width=0.22\textwidth]{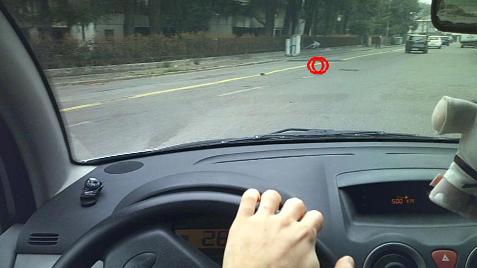} & \includegraphics[width=0.22\textwidth]{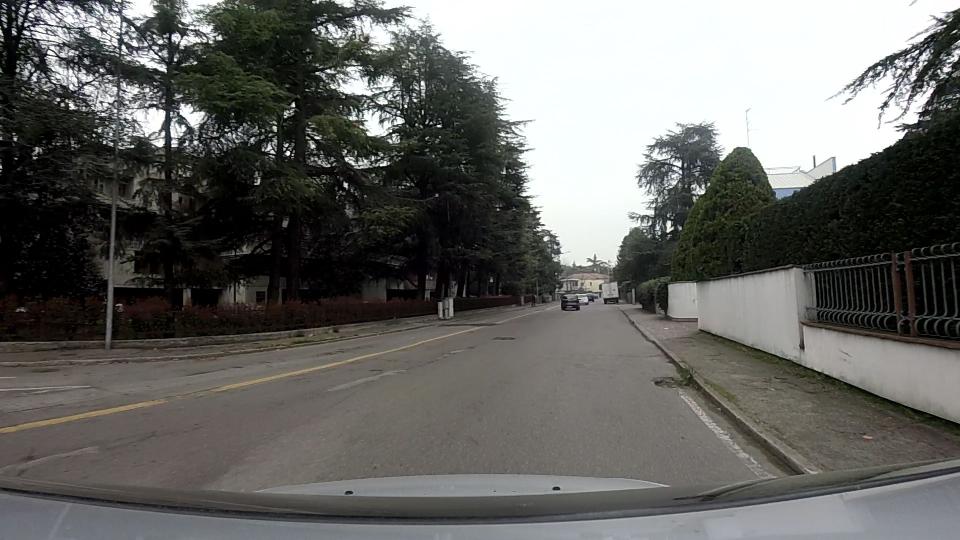} &
\includegraphics[width=0.22\textwidth]{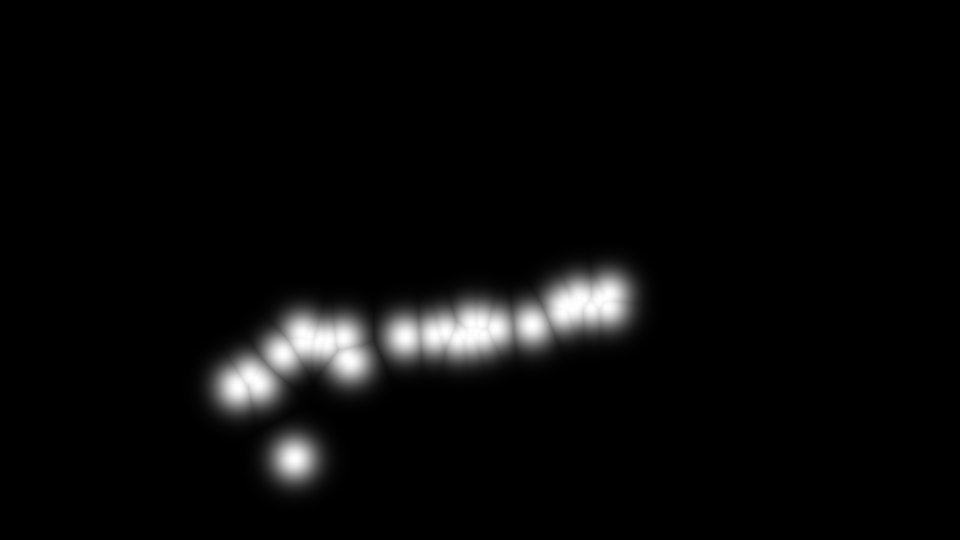} & \includegraphics[width=0.22\textwidth]{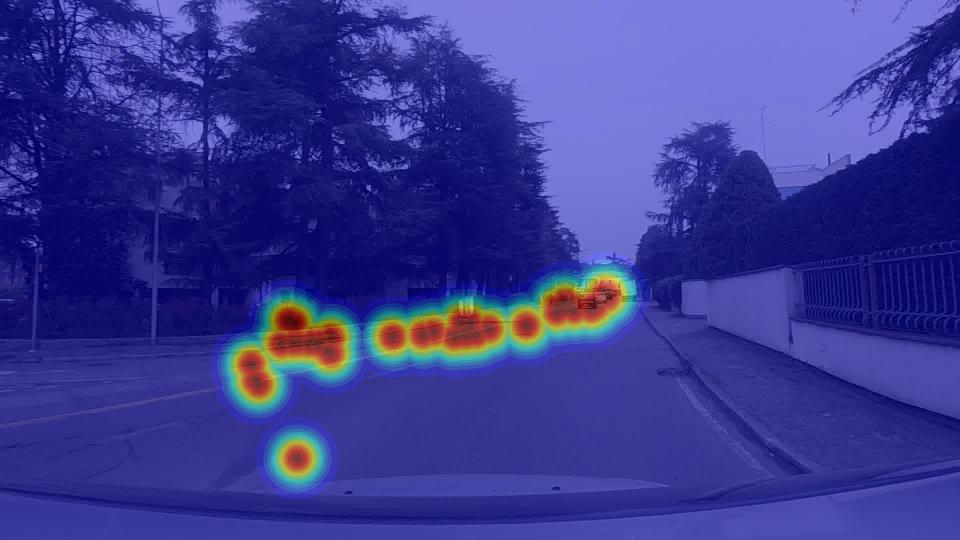} \\ 
\end{tabular}
\egroup
\label{fig:sequence}
\caption{Examples taken from a random sequence of \drive. From left to right: frames from the eye tracking glasses with gaze data, from the roof-mounted camera, temporal aggregated fixation maps (as defined in Sec.~\ref{sec:dataset}) and overlays between frames and fixation maps.}
\end{figure*}
The way humans favor some entities in the scene, along with key factors guiding eye fixations in presence of a given task (e.g. visual search) has been extensively studied for decades~\cite{treisman1980feature,wolfe1998visual}. 
The main difficulty that rises when approaching the subject is the variety of perspectives under which it can be cast.
Indeed, visual attention has been approached by psychologists, neurobiologists and computer scientists, making the field highly interdisciplinary~\cite{frintrop2010computational}.
We are particularly interested in the computational perspective, in which predicting human attention is often formalized as an estimation task delivering the probability of each point in a given scene to attract the observer's gaze.
\\
\\
{\bf Attention in images and videos.} 
Coherently with psychological literature, that identifies two distinct mechanisms guiding human eye fixations~\cite{theeuwes2010top}, computational models for FoA prediction branch into two families: top-down and bottom-up strategies.
% Literature concerning human attention prediction on images branches in two families: top-down and bottom-up strategies. 
Former approaches aim at highlighting objects and cues that could be meaningful in the context of a given task. For this reason, such methods are also known as task-driven. Usually, top-down computer vision models are built to integrate semantic contextual information in the attention prediction process~\cite{torralba2006contextual}. This can be achieved by either merging estimated maps at different levels of scale and abstraction~\cite{Goferman:2012:CSD:2360766.2361199}, or including a-priori cues about relevant objects for the task at hand~\cite{Wolfe89guidedsearch,Gao08onthe,Elazary20101338}. Human focus in complex interactive environments (\textit{e.g.} while playing videogames)~\cite{peters2007beyond,peters2008applying,borji2014look} follows task-driven behaviors as well.

Conversely, bottom-up models capture salient objects or events naturally popping out in the image, independently of the observer, the undergoing task and other external factors. This task is widely known in literature as \emph{visual saliency prediction}. In this context, computational models focus on spotting visual discontinuities, either by clustering features or considering the rarity of image regions, locally~\cite{6287326,Ma} or globally~\cite{5206596,Zhai,6871397}.
For a comprehensive review of visual attention prediction methods, we refer the reader to~\cite{borji2013state}.
Recently, the success of deep networks involved both task-driven attention and saliency prediction, as models have become more powerful in both paradigms, achieving state-of-the-art results on public benchmarks~\cite{deepgaze,dn2,dn3,mlnet2016,cornia2016predicting}.\\
In video, attention prediction and saliency estimation are more complex with respect to still images since motion heavily affects human gaze. Some models merge bottom-up saliency with motion maps, either by means of optical flow~\cite{zhong2013video} or feature tracking~\cite{Zhai:2006}.
Other methods enforce temporal dependencies between bottom-up features in successive frames. Both supervised~\cite{zhong2013video,MatheSminchisescuPAMI2015} and unsupervised~\cite{mauthner2015encoding, wang2015saliency, wang2015consistent} feature extraction can be employed, and temporal coherence can be achieved either by conditioning the current prediction on information from previous frames~\cite{rudoy2013learning} or by capturing motion smoothness with optical flow~\cite{zhong2013video,MatheSminchisescuPAMI2015}. While deep video saliency models still lack, an interesting work is~\cite{bazzani2017recurrent}, which relies on a recurrent architecture fed with clip encodings to predict the fixation map by means of a Gaussian Mixture Model (GMM).
Nevertheless, most methods limit to bottom-up features accounting for just visual discontinuities in terms of textures or contours. Our proposal, instead, is specifically tailored to the driving task and fuses the bottom-up information with semantics and motion elements that have emerged as attention factors from the analysis of the \drive~dataset.
\\
{\bf Attention and driving.} 
Prior works addressed the task of detecting saliency and attention in the specific context of assisted driving.
In such cases, however, gaze and attentive mechanisms have been mainly studied for some driving sub-tasks only, often acquiring gaze maps from on-screen images. Bremond~\emph{et al.}~\cite{bremond1} presented a model that exploits visual saliency with a non-linear SVM classifier for the detection of traffic signs. The validation of this study was performed in a laboratory non-realistic setting, emulating an in-car driving session. A more realistic experiment~\cite{bremond2} was then conducted with a larger set of targets, \emph{e.g.} including pedestrians and bicycles. \\
Driver's gaze has also been studied in a pre-attention context, by means of intention prediction relying only on fixation maps~\cite{preattention}. The study in~\cite{TJunct} inspects the driver's attention at T junctions, in particular towards pedestrians and motorbikes, and exploits object saliency to avoid the \emph{looked-but-failed-to-see} effect. In absence of eye tracking systems and reliable gaze data,~\cite{fridman2015driver,head2,head3,borghi2016poseidon} focus on drivers' head, detecting facial landmarks to predict head orientation. Such mechanisms are more robust to varying lighting conditions and occlusions, but there is no certainty about the adherence of predictions to the true gaze during the driving task.
\\
\\
{\bf Datasets.} Many image saliency datasets have been released in the past few years, improving the understanding of the human visual attention and pushing computational models forward. Most of these datasets include no motion information, as saliency ground truth maps are built by aggregating fixations of several users within the same still image. Usually, a Gaussian filtering post-processing step is employed on recorded data, in order to smooth such fixations and integrate their spatial locations. Some datasets, such as the MIT saliency benchmark~\cite{mit-saliency-benchmark}, were labeled through an eye tracking system, while others, like the SALICON dataset~\cite{jiang2015salicon} relied on users clicking on salient image locations. We refer the reader to~\cite{CAT2000} for a comprehensive list of  available datasets.
On the contrary, datasets addressing human attention prediction in video still lack. Up to now, \emph{Action in the Eye}~\cite{6942210} represents the most important contribution, since it consists in the largest video dataset accompanied by gaze and fixation annotations. That information, however, is collected in the context of action recognition, so it is heavily task-driven. A few datasets address directly the study of attention mechanisms while driving, as summarized in Tab.~\ref{tab:dreyeve_features}. However, these are mostly restricted to limited settings and are not publicly available. In some of them~\cite{bremond1,TJunct} fixation and saliency maps are acquired during an in-lab simulated driving experience. In-lab experiments enable several attention drifts that are influenced by external factors (\emph{e.g.} monitor distance and others) rather than the primary task of driving~\cite{Tatler}. A few in-car datasets exist~\cite{bremond2,preattention}, but were precisely tailored to force the driver to fulfill some tasks, such as looking at people or traffic signs. Coarse gaze information is also available in~\cite{fridman2015driver}, while the external road scene images are not acquired. We believe that the dataset presented in~\cite{preattention} is, among the others, the closer to our proposal. Yet, video sequences are collected from one driver only it is not publicly available. Conversely, our \drive~dataset is the first dataset addressing driver's focus of attention prediction that is made publicly available. Furthermore, it includes sequences from several different drivers and presents a high variety of landscapes (\emph{i.e.} highway, downtown and countryside), lighting and weather conditions. 

%%%%%%%%%%%%%%%%%%%%%%%%%%%%%%%%%%%%%%%%%%%%%%%%%%%%%%%%%%%%%%%%%%%%%%
%%%%%%%%%%%%%%%%%% DATASET  %%%%%%%%%%%%%%%%%%%%%%%%%%%%%%%%%%%%%%%%%%
%%%%%%%%%%%%%%%%%%%%%%%%%%%%%%%%%%%%%%%%%%%%%%%%%%%%%%%%%%%%%%%%%%%%%%
\section{The \drive~project}
\label{sec:dataset}
\begin{figure}[b]
\centering
\includegraphics[width=\columnwidth]{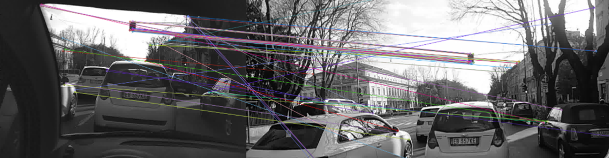}
\caption{Registration between the egocentric and roof-mounted camera views by means of SIFT descriptor matching.}
\label{fig:sift}
\end{figure}

\begin{figure}[t]
\centering
\includegraphics[width=0.9\columnwidth]{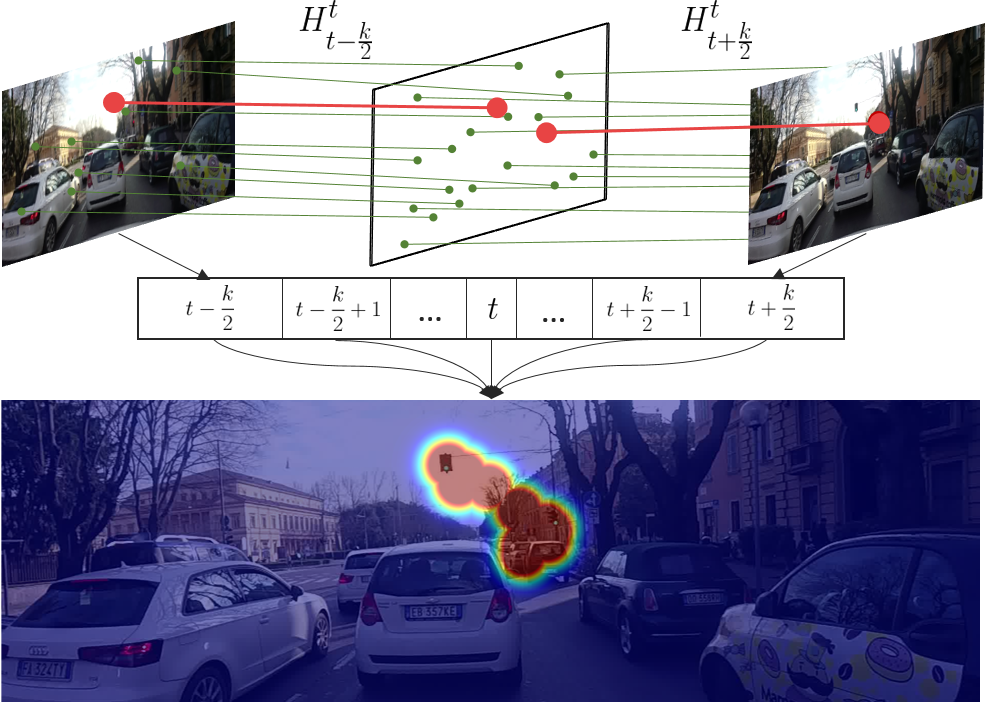}
\caption{Resulting fixation map from a 1 second integration (25 frames). The adoption of the \emph{max} aggregation of equation~\ref{eq:projection} allows to account in the final map two brief glances towards traffic lights.}
\label{fig:schema_mappa}
\end{figure}

In this section we present the \drive~dataset (Fig.~\ref{fig:sequence}), the protocol adopted for video registration and annotation, the automatic processing of eye-tracker data and the analysis of the driver's behavior in different conditions.
\\
\\
{\bf The dataset.}
The \drive~dataset consists of 555,000 frames divided in 74 sequences, each of which is 5 minutes long. Eight different drivers of varying age from 20 to 40, including 7 men and a woman, took part to the driving experiment, that lasted more than two months.
Videos were recorded in different contexts, both in terms of landscape (downtown, countryside, highway) and traffic condition, ranging from traffic-free to highly cluttered scenarios. They were recorded in diverse weather conditions (sunny, rainy, cloudy) and at different hours of the day (both daytime and night). Tab.~\ref{tab:dreyeve_features} recaps the dataset features and Tab.~\ref{tab:datasets} compares it with other related proposals. \drive~ is currently the largest publicly available dataset including gaze and driving behavior in automotive settings.
\begin{table*}[tb]
\centering
\caption{Summary of the \drive~dataset characteristics. The dataset was designed to embody the most possible diversity in the combination of different features. The reader is referred to either the additional material or to the dataset presentation~\cite{alletto2016dr} for details on each sequence.}
\label{tab:dreyeve_features}
\begin{tabular}{|c|c|c|c|c|c|c|c|}
\hline
\textbf{\# Videos}  & \textbf{\# Frames}      & \textbf{Drivers}   & \textbf{Weather conditions} & \textbf{Lighting} & \textbf{Gaze Info} & \textbf{Metadata} & \textbf{Camera Viewpoint} \\ \hline
\multirow{3}{*}{74} & \multirow{3}{*}{555,000} & \multirow{3}{*}{8} & sunny                       & day               & raw fixations      & GPS               & driver (720p)        \\ \cline{4-8} 
                    &                         &                    & cloudy                      & evening           & gaze map           & car speed         & car (1080p)          \\ \cline{4-8} 
                    &                         &                    & rainy                       & night             & pupil dilation     & car course        &                      \\ \hline
\end{tabular}
\end{table*}
\begin{table*}[tb]
\centering
\caption{A comparison between \drive~and other datasets.}
\label{tab:datasets}
\begin{tabular}{l c c l l c c}
\hline
\textbf{Dataset}                                & \textbf{Frames}    & \textbf{Drivers} & \textbf{Scenarios}                                                                & \textbf{Annotations}                                                                                              & \textbf{Real-world} & \textbf{Public} \\ \hline
Pugeault~\etal~\cite{preattention}     & 158,668   & --    & \begin{tabular}[c]{@{}l@{}}Countryside, Highway\\  Downtown\end{tabular} & \begin{tabular}[c]{@{}l@{}}Gaze Maps\\Driver's Actions\end{tabular} & Yes        & No     \\ \hline
Simon \etal \cite{bremond1}            & 40        & 30      & Downtown                                                                 & Gaze Maps                                                                                                & No         & No     \\ \hline
Underwood \etal \cite{TJunct}          & 120       & 77      & Urban Motorway                                                           & --                                                                                                     & No         & No     \\ \hline
Fridman \etal \cite{fridman2015driver} & 1,860,761 & 50      & Highway                                                                  & 6 Gaze Location Classes                                                                                  & Yes        & No     \\ \hline
\drive\cite{alletto2016dr}                              & 555,000   & 8       & \begin{tabular}[c]{@{}l@{}}Countryside, Highway\\  Downtown\end{tabular} & \begin{tabular}[c]{@{}l@{}}Gaze Maps\\GPS, Speed, Course\end{tabular} & Yes        & Yes    \\ \hline
\end{tabular}
\end{table*}
\\
\\
{\bf The Acquisition System.} The driver's gaze information was captured using the commercial \emph{SMI ETG 2w} Eye Tracking Glasses (ETG). ETG capture attention dynamics also in presence of head pose changes, which occur very often during the task of driving. While a frontal camera acquires the scene at 720p/30fps, users pupils are tracked at 60Hz. Gaze information are provided in terms of eye fixations and saccade movements. ETG was manually calibrated before each sequence for every driver.\\
Simultaneously, videos from the car perspective were acquired using the \emph{GARMIN VirbX} camera mounted on the car roof (RMC, Roof-Mounted Camera).
Such sensor captures frames at 1080p/25fps, and includes further information such as GPS data, accelerometer and gyroscope measurements.
\\
\\
{\bf Video-gaze registration.}
The dataset has been processed to move the acquired gaze from the egocentric (ETG) view to the car (RMC) view.
The latter features a much wider field of view (FoV), and can contain fixations that are out of the egocentric view.
For instance, this can occur whenever the driver takes a peek at something at the border of this FoV, but doesn't move his head.
For every sequence, the two videos were manually aligned to cope with the difference in sensors framerate.
Videos were then registered frame-by-frame through a homographic transformation that projects fixation points across views.
More formally, at each timestep $t$ the RMC frame $I_{RMC}^t$ and the ETG frame $I_{ETG}^t$ are registered by means of a homography matrix $H_{ETG\rightarrow RMC}$, computed by matching SIFT descriptors~\cite{lowe1999object} from one view to the other (see~Fig.~\ref{fig:sift}).
A further RANSAC~\cite{fischler1981random} procedure ensures robustness to outliers. While homographic mapping is theoretically sound only across planar views - which is not the case of outdoor environments - we empirically found that projecting an object from one image to another always recovered the correct position. This makes sense if the distance between the projected object and the camera is far greater than the distance between the object and the projective plane. In Sec.~\ref{sup:bound} of the supplementary material, we derive formal bounds to explain this phenomena.
\\
\\
{\bf Fixation map computation.}
The pipeline discussed above provides a frame-level annotation of the driver's fixations.
In contrast to image saliency experiments~\cite{mit-saliency-benchmark}, there is no clear and indisputable protocol for obtaining continuous maps from raw fixations when acquired in task-driven real-life scenarios. This is even more evident when fixations are collected in task-driven real-life scenarios. The main motivation resides in the fact that observer's subjectivity cannot be removed by averaging different observers' fixations. Indeed two different observers cannot experience the same scene at the same time (\textit{e.g.} two drivers cannot be at the same time in the same point of the street). The only chance to average among different observers would be the adoption of a simulation environment, but it has been proved that the cognitive load in controlled experiments is lower than in real test scenarios and it effects the true attention mechanism of the observer~\cite{cogni}. In our preliminary \drive~release \cite{alletto2016dr}, fixation points were aggregated and smoothed by means of a temporal sliding window. In such a way, temporal filtering discarded momentary glimpses that contain precious information about the driver's attention. Following the psychological protocol in \cite{mannan1997fixation}~and~\cite{groner1984looking}, this limitation was overcome in the current release where the new fixation maps were computed without temporal smoothing.
Both~\cite{mannan1997fixation}~and~\cite{groner1984looking} highlight the high degree of subjectivity of scene scanpaths in short temporal windows ($<1$ sec) and suggest to neglect the fixations pop-out order within such windows. This mechanism also ameliorates the \emph{inhibition of return} phenomenon that may prevent interesting objects to be observed twice in short temporal intervals~\cite{posner1985inhibition, henderson2003human}, leading to the underestimation of their importance.\\
More formally, the \emph{fixation map} $F_t$ for a frame at time $t$ is built by accumulating projected gaze points in a temporal sliding window of $k=25$ frames, centered in $t$. For each time step $t+i$ in the window, where $i \in \{-\frac{k}{2}, -\frac{k}{2} + 1, \ldots, \frac{k}{2}-1, \frac{k}{2}\}$, gaze points projections on $F_t$ are estimated through the homography transformation $H_{t+i}^t$ that projects points from the image plane at frame $t+i$, namely $p_{t+i}$, to the image plane in $F_t$. A continuous fixation map is obtained from the projected fixations by centering on each of them a multivariate Gaussian having a diagonal covariance matrix $\Sigma$ (the spatial variance of each variable is set to $\sigma_s^2 = 200$ pixels) and taking the \emph{max} value along the time axis:
\begin{equation}
F_t(x,y) = \max_{i \in (-\frac{k}{2},..., \frac{k}{2})} \mathcal{N}((x,y)\given H_{t+i}^t \cdot p_{t+i}, \Sigma)  
\label{eq:projection}
\end{equation}
The Gaussian variance has been computed by averaging the ETG spatial acquisition errors on 20 observers looking at calibration patterns at different distances from 5 to 15 meters.
The described process can be appreciated in Fig.~\ref{fig:schema_mappa}. Eventually, each map $F_t$ is normalized to sum to 1, so that it can be considered a probability distribution of fixation points.
\begin{figure}[bt]
    \centering
    \begin{tabular}{cc}
    \hspace{-0.2cm}\includegraphics[width=0.21\textwidth]{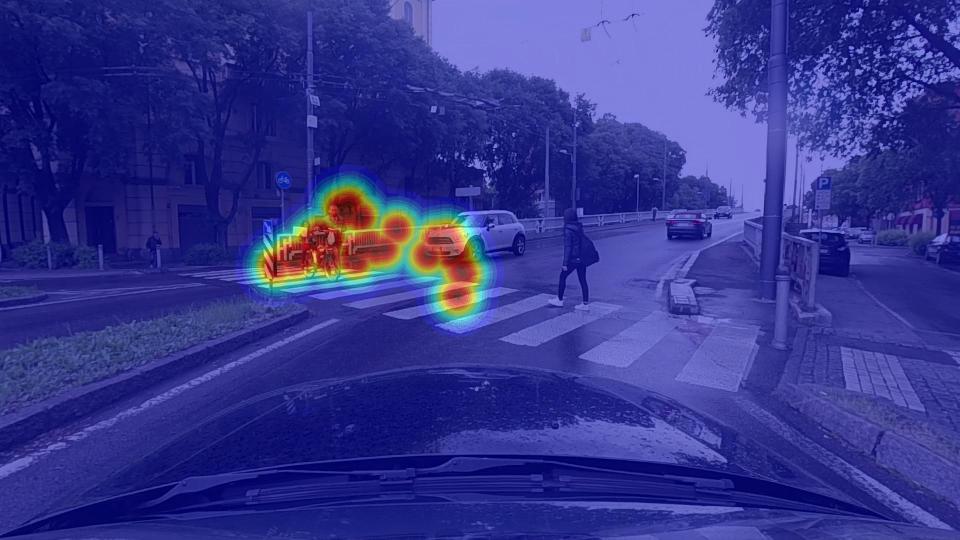} &\hspace{-0.3cm}\includegraphics[width=0.21\textwidth]{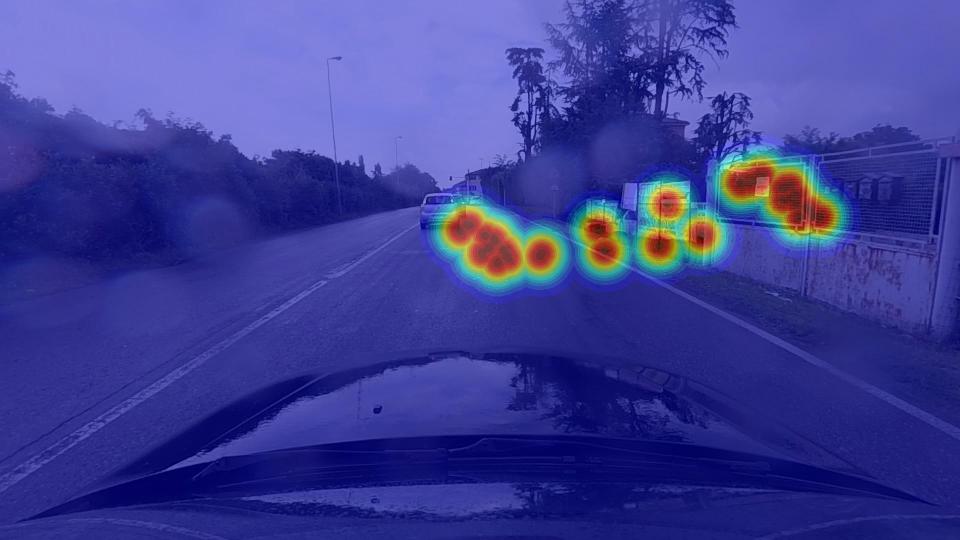}\\
    \hspace{-0.2cm}(a) Acting - 69\,719 & \hspace{-0.3cm}(b) Inattentive - 12\,282 \\
    \hspace{-0.2cm}\includegraphics[width=0.21\textwidth]{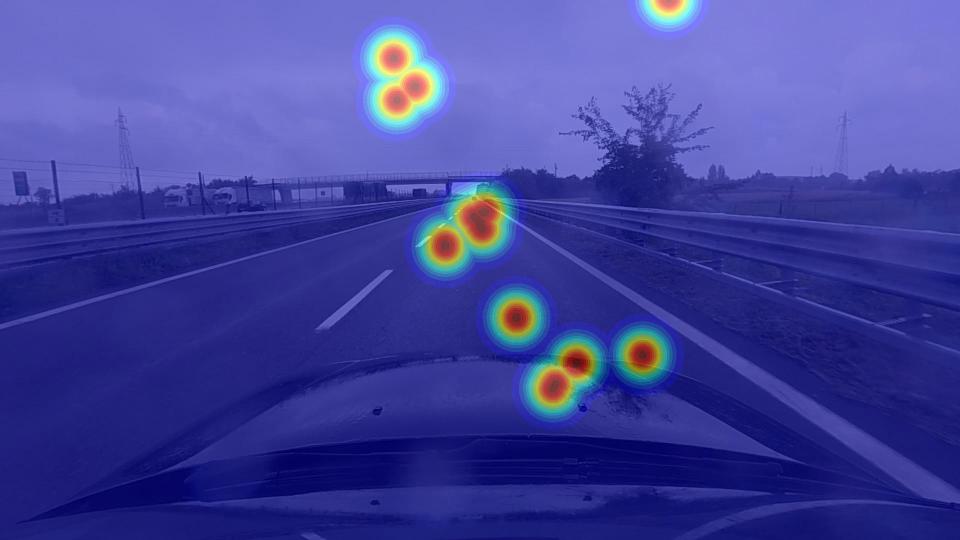} &\hspace{-0.3cm}\includegraphics[width=0.21\textwidth]{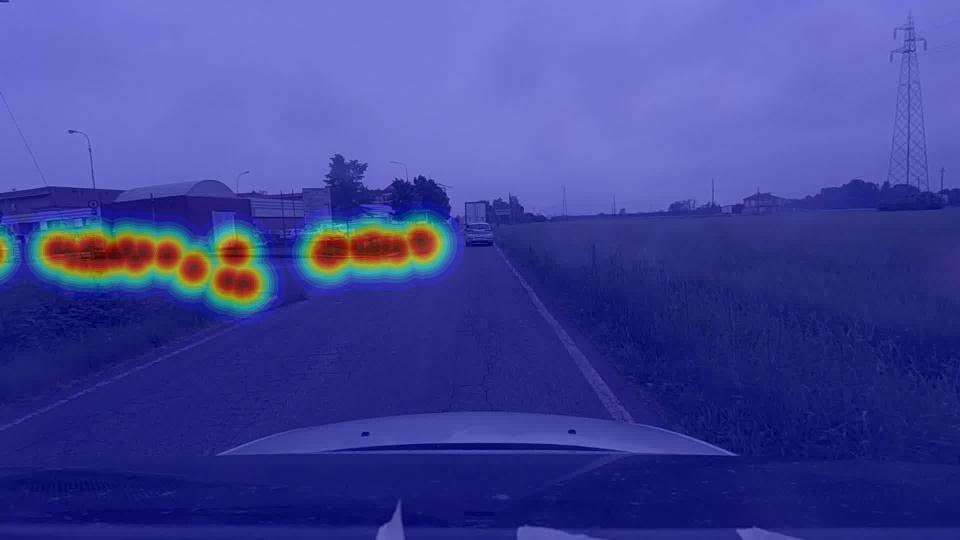}\\
    \hspace{-0.2cm}(c) Error - 22\,893 & \hspace{-0.3cm}(d) Subjective - 3\,166
    \end{tabular}
    \caption{Examples of the categorization of frames where gaze is far from the mean. Overall, 108\,060 frames ($\sim$20\% of \drive) were extended with this type of information.}
    \label{fig:att_vs_inatt}
\end{figure}
\\
\\
\noindent {\bf Labeling attention drifts.} Fixation maps exhibit a very strong central bias. This is common in saliency annotations~\cite{tatler2007central} and even more in the context of driving.
For these reasons, there is a strong unbalance between lots of easy-to-predict scenarios and unfrequent but interesting hard-to-predict events.\\
\begin{figure}[b]
    \centering
    \begin{tabular}{cc}
        \hspace{-0.2cm}\includegraphics[width=0.21\textwidth]{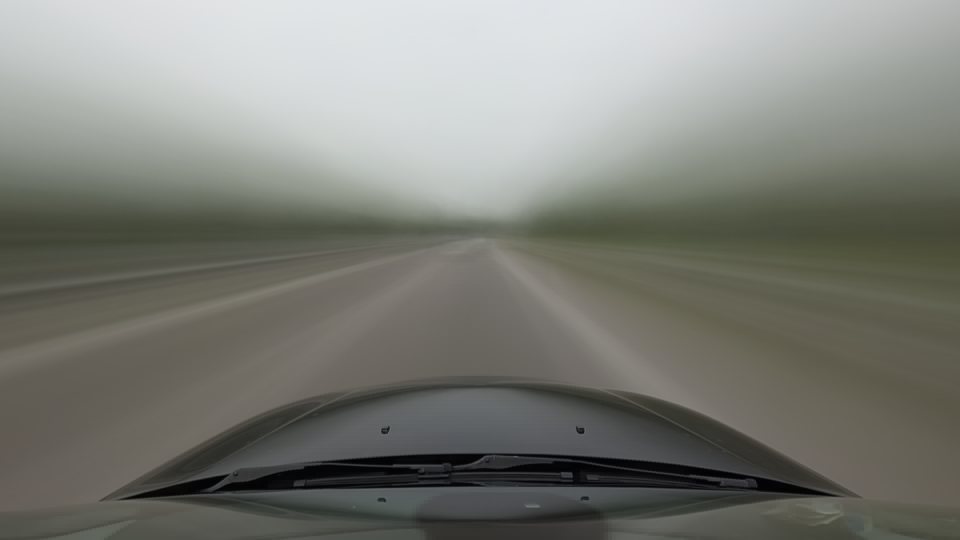} & \hspace{-0.2cm} \includegraphics[width=0.21\textwidth]{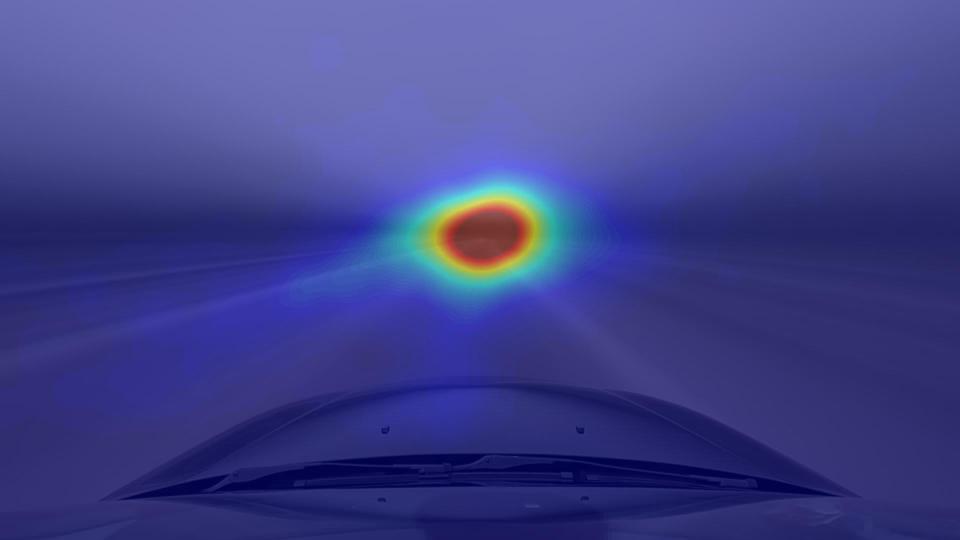} \\
        (a) & (b)
    \end{tabular}
    \caption{Mean frame (a) and fixation map (b) averaged across the whole sequence \texttt{02}, highlighting the link between driver's focus and the vanishing point of the road.}
    \label{fig:vanishing_point}
\end{figure}
\begin{figure*}[ht]
\centering
    \begin{tabular}{ccccc}
    $|\Sigma|=2.49\times 10^9$&
    $|\Sigma|=2.03\times 10^9$&
    $|\Sigma|=8.61\times 10^8$&
    $|\Sigma|=1.02\times 10^9$&
    $|\Sigma|=9.28\times 10^8$\\
    \includegraphics[width=0.18\textwidth]{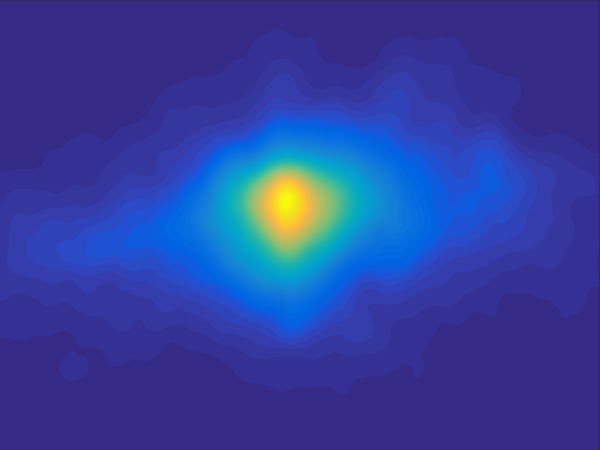} &
    \includegraphics[width=0.18\textwidth]{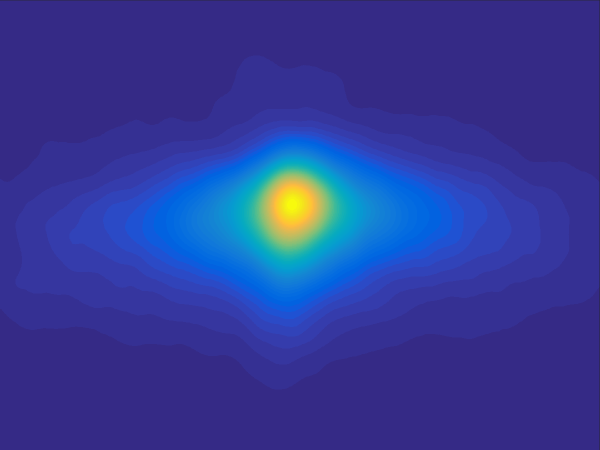} &
    \includegraphics[width=0.18\textwidth]{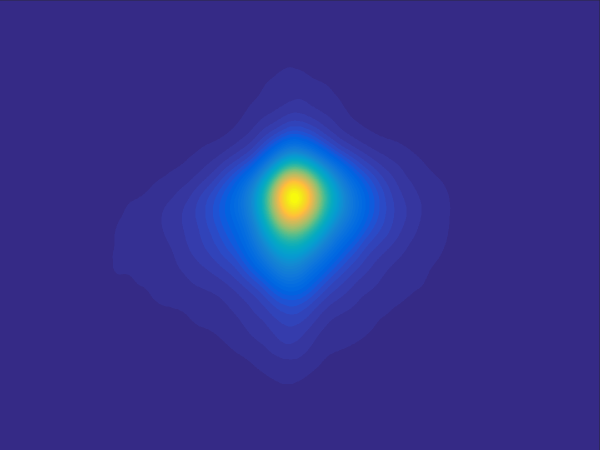} &
    \includegraphics[width=0.18\textwidth]{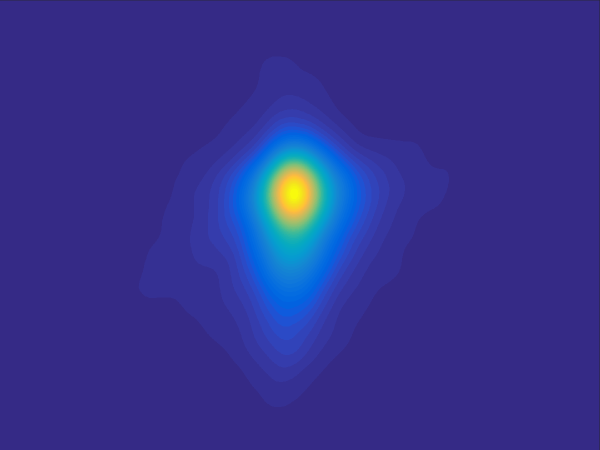} &
    \includegraphics[width=0.18\textwidth]{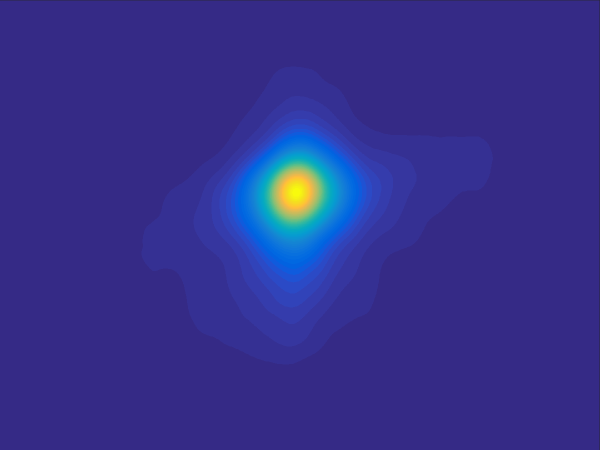}\\
    \includegraphics[width=0.18\textwidth, height=0.25cm]{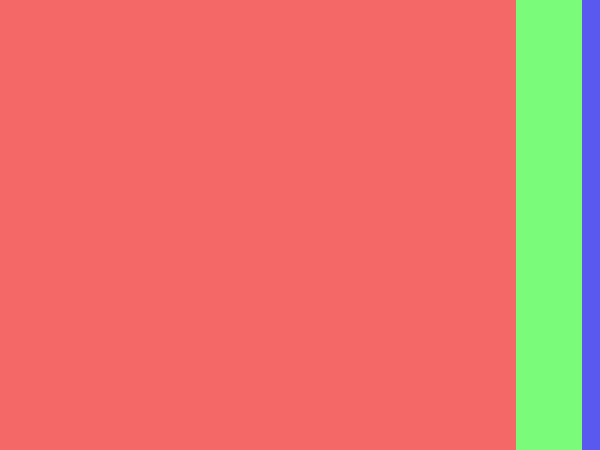} &
    \includegraphics[width=0.18\textwidth, height=0.25cm]{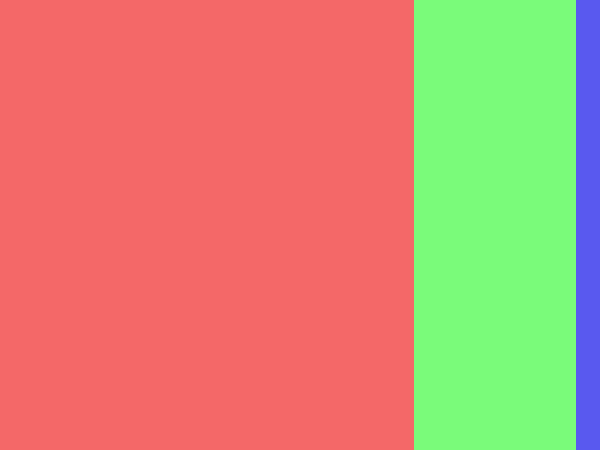} &
    \includegraphics[width=0.18\textwidth, height=0.25cm]{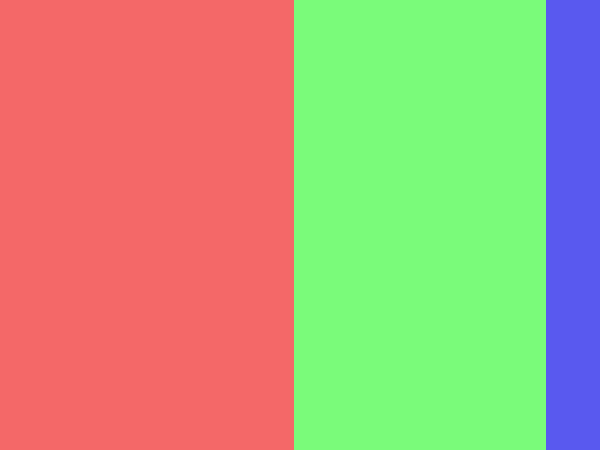} &
    \includegraphics[width=0.18\textwidth, height=0.25cm]{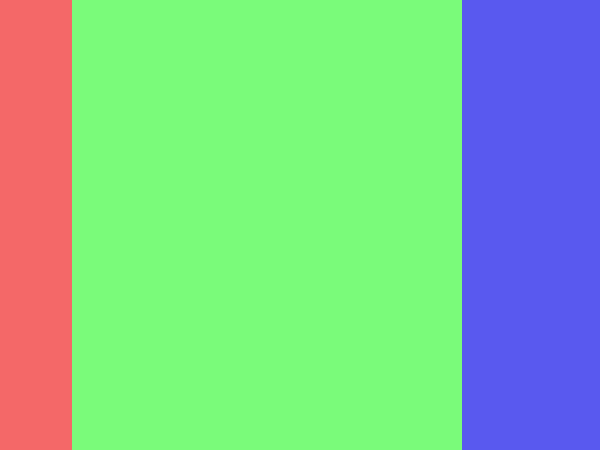} &
    \includegraphics[width=0.18\textwidth, height=0.25cm]{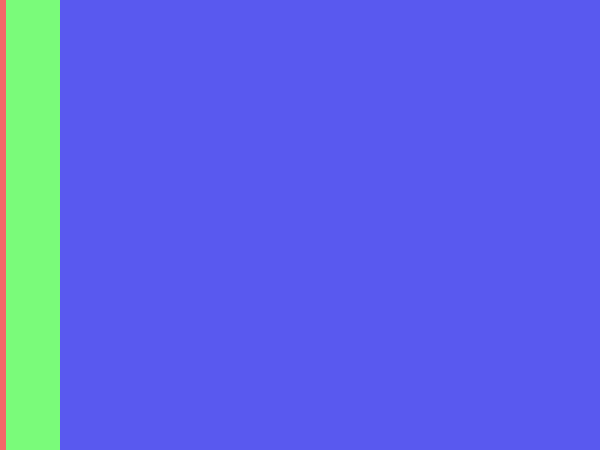}\\
    (a) $0\leq \mathrm{km/h}\leq10$ &
    (b) $10\leq \mathrm{km/h}\leq30$ &
    (c) $30\leq \mathrm{km/h}\leq50$ &
    (d) $50\leq \mathrm{km/h}\leq70$ &
    (e) $70\leq \mathrm{km/h}$
    \end{tabular}
    \caption{As speed gradually increases, driver's attention converges towards the vanishing point of the road. (a) When the car is approximately stationary, the driver is distracted by many objects in the scene. (b-e) As the speed increases, the driver's gaze deviates less and less from the vanishing point of the road. To measure this effect quantitatively, a two-dimensional Gaussian is fitted to approximate the mean map for each speed range, and the determinant of the covariance matrix $\Sigma$ is reported as an indication of its spread (the determinant equals the product of eigenvalues, each of which measures the spread along a different data dimension). The bar plots illustrate the amount of downtown (red), countryside (green) and highway (blue) frames that concurred to generate the average gaze position for a specific speed range. Best viewed on screen.}
    \label{fig:gt_across_speed}
\end{figure*}
\begin{figure*}[ht]
    %\centering
    \begin{tabular}{cccccccccc}
    \begin{turn}{90}{\small {\fontfamily{phv}\selectfont Probability of fixation}}\end{turn}
    \includegraphics[height=3cm]{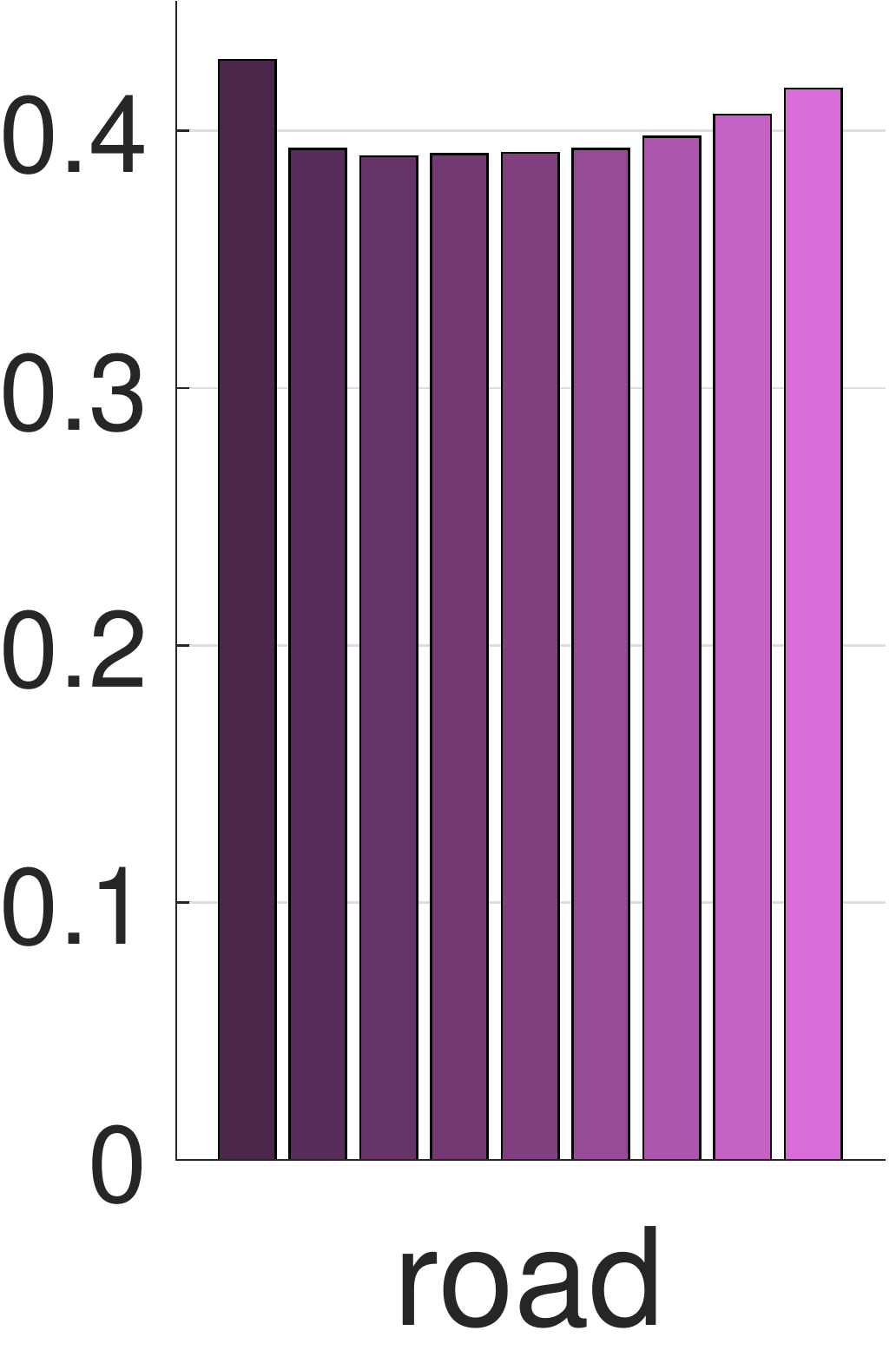} &
    \hspace{-0.5cm}\includegraphics[height=3cm]{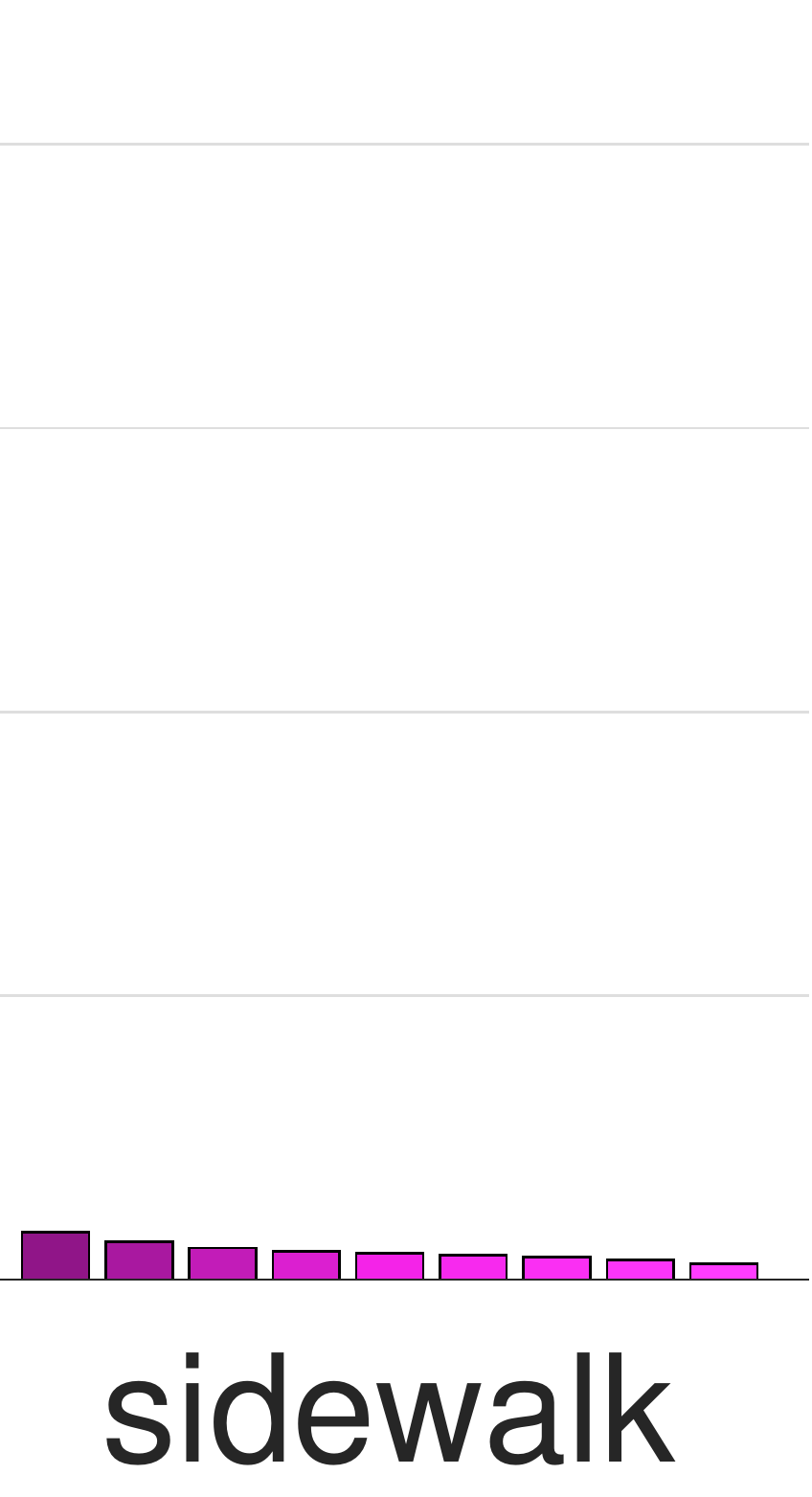} &
    \hspace{-0.45cm}\includegraphics[height=3cm]{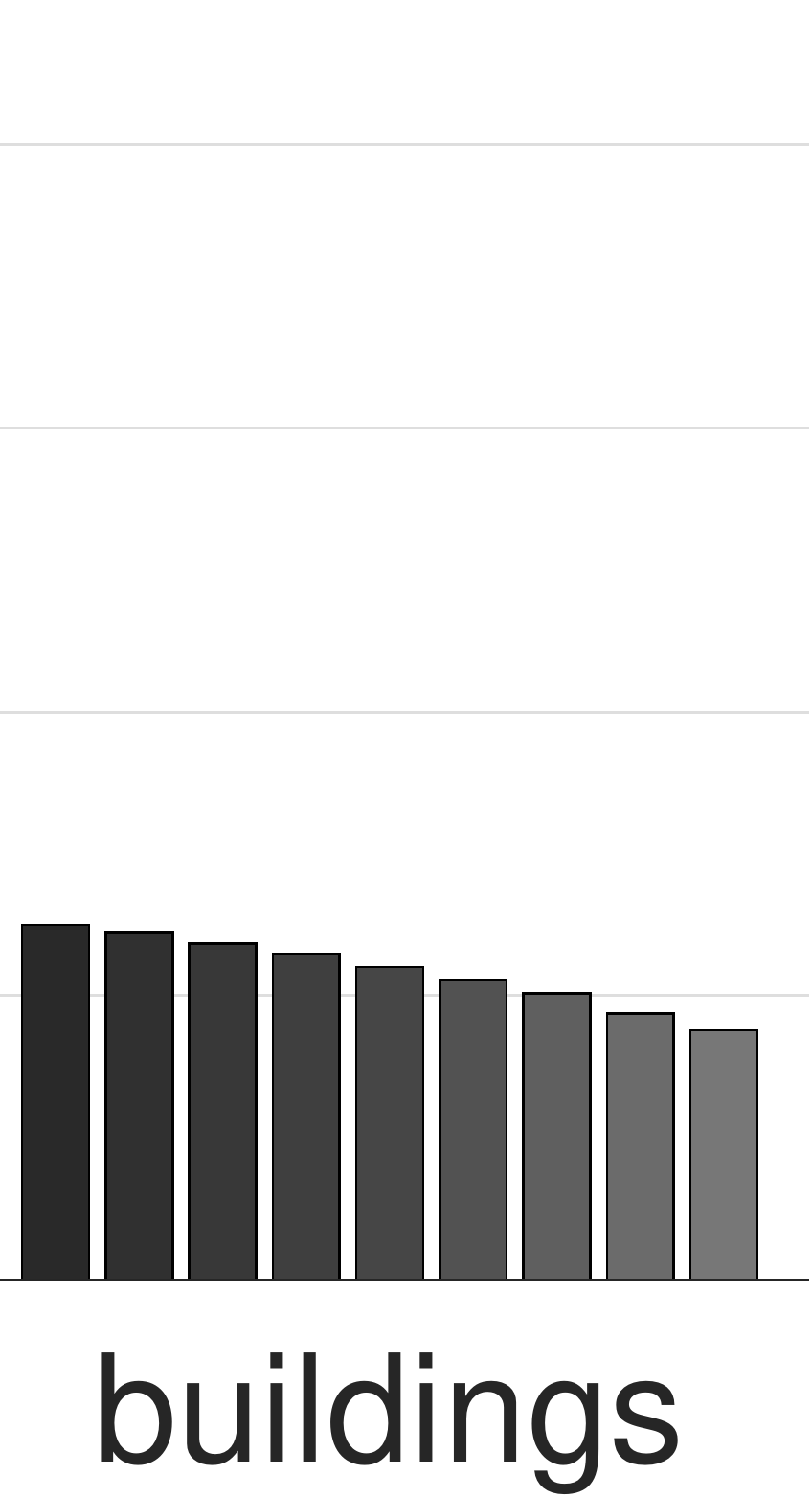} &
    \hspace{-0.45cm}\includegraphics[height=3cm]{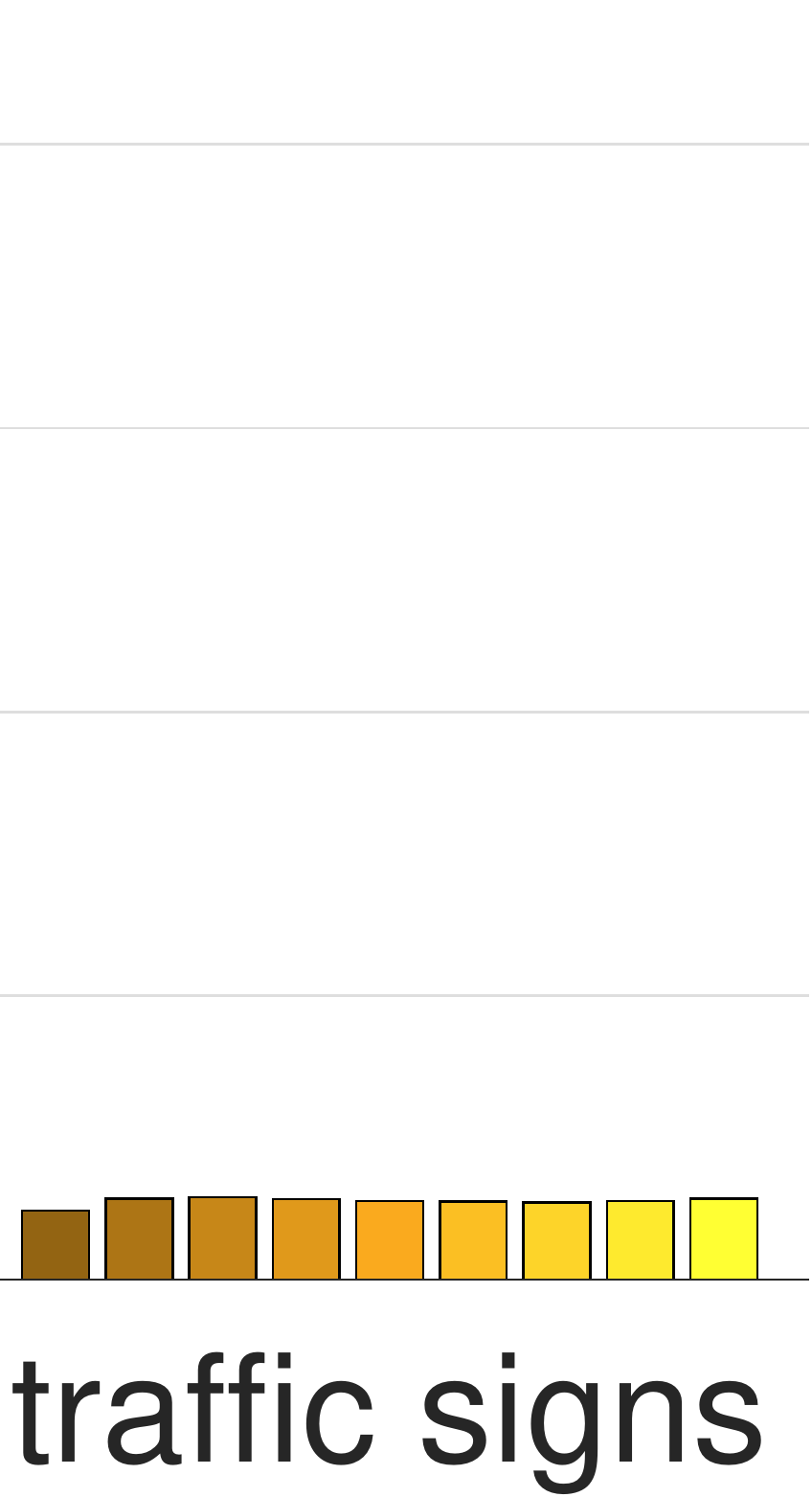} &
    \hspace{-0.45cm}\includegraphics[height=3cm]{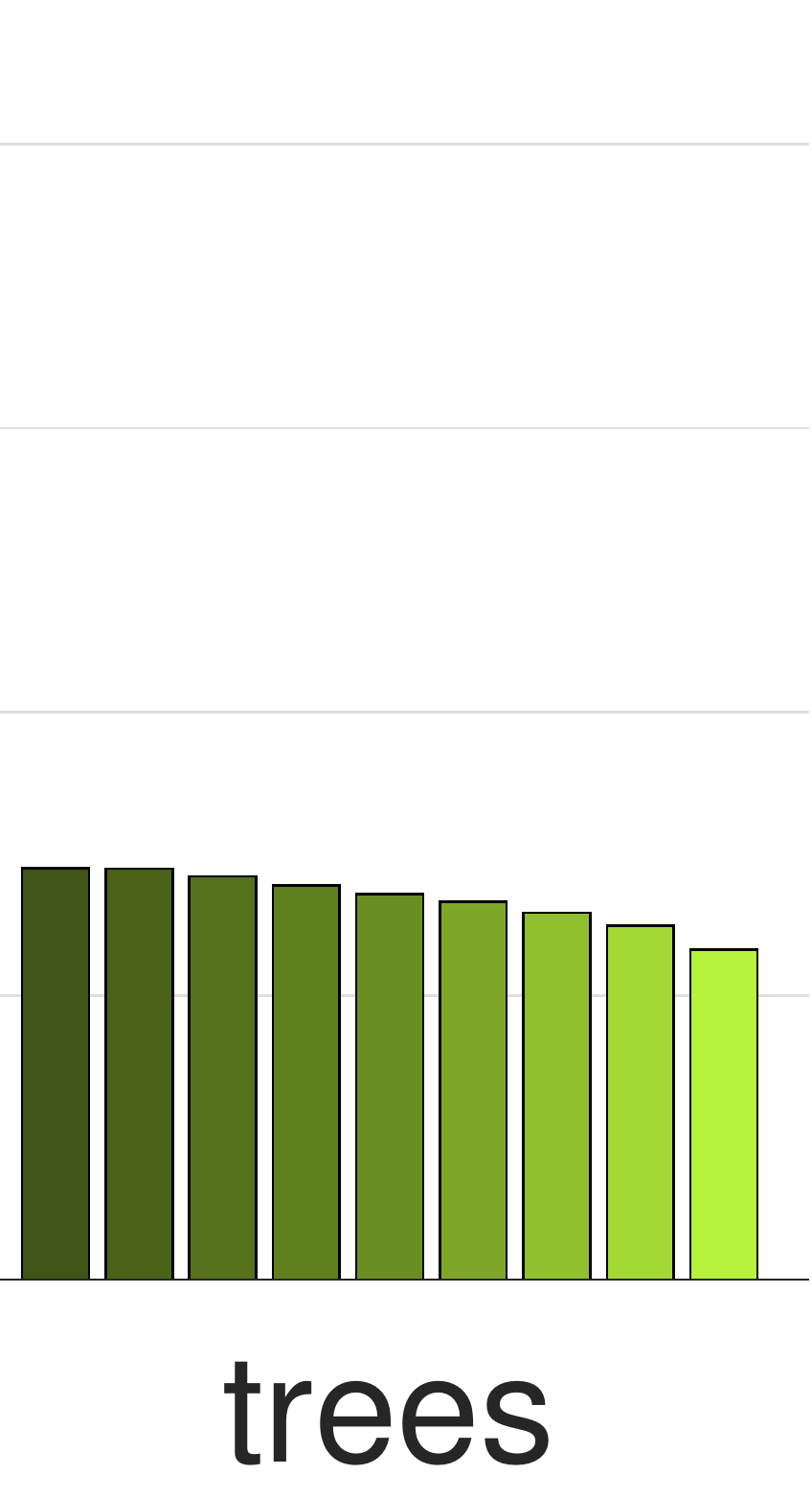} &
    \hspace{-0.45cm}\includegraphics[height=3cm]{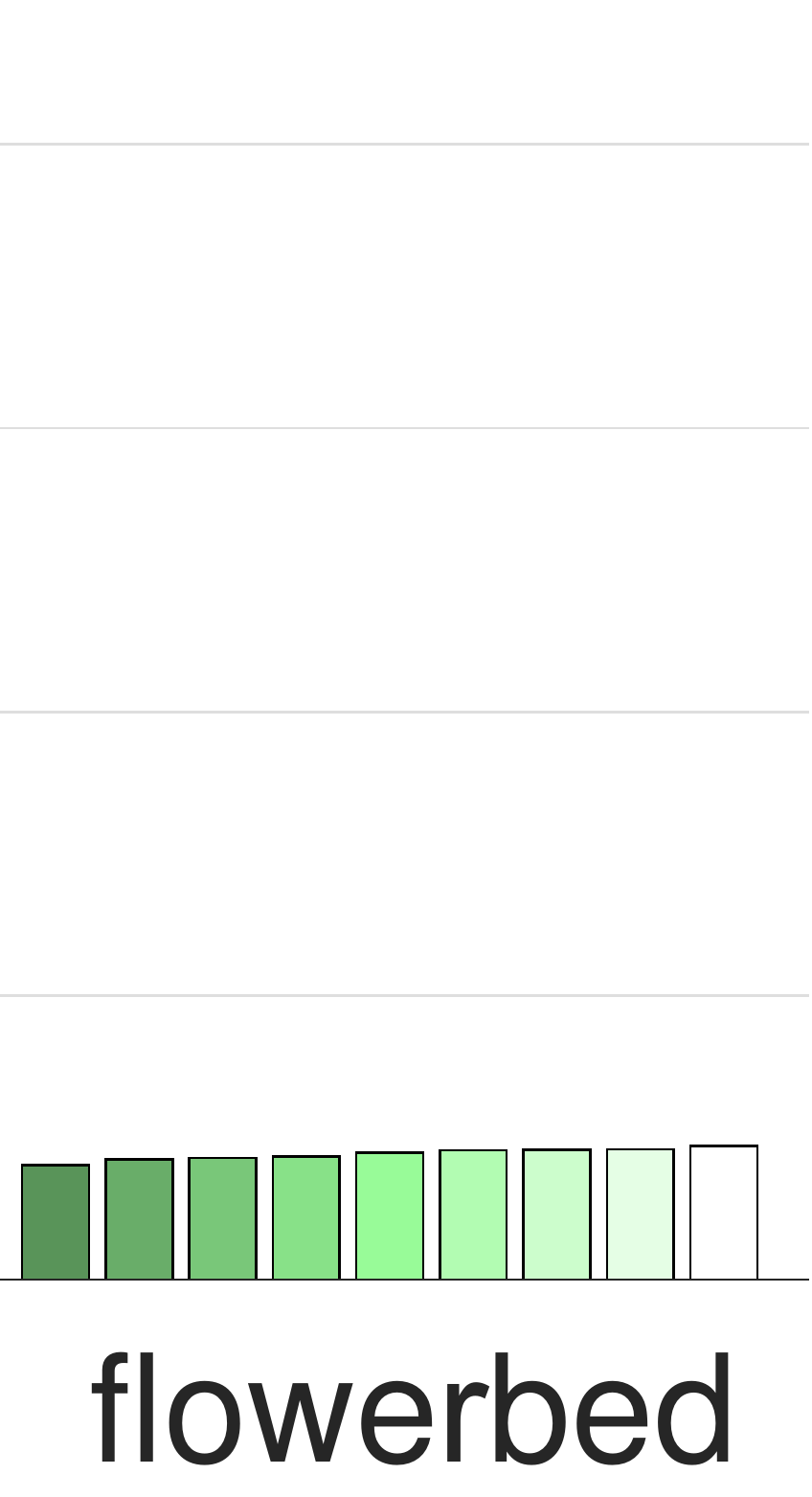} &
    \hspace{-0.45cm}\includegraphics[height=3cm]{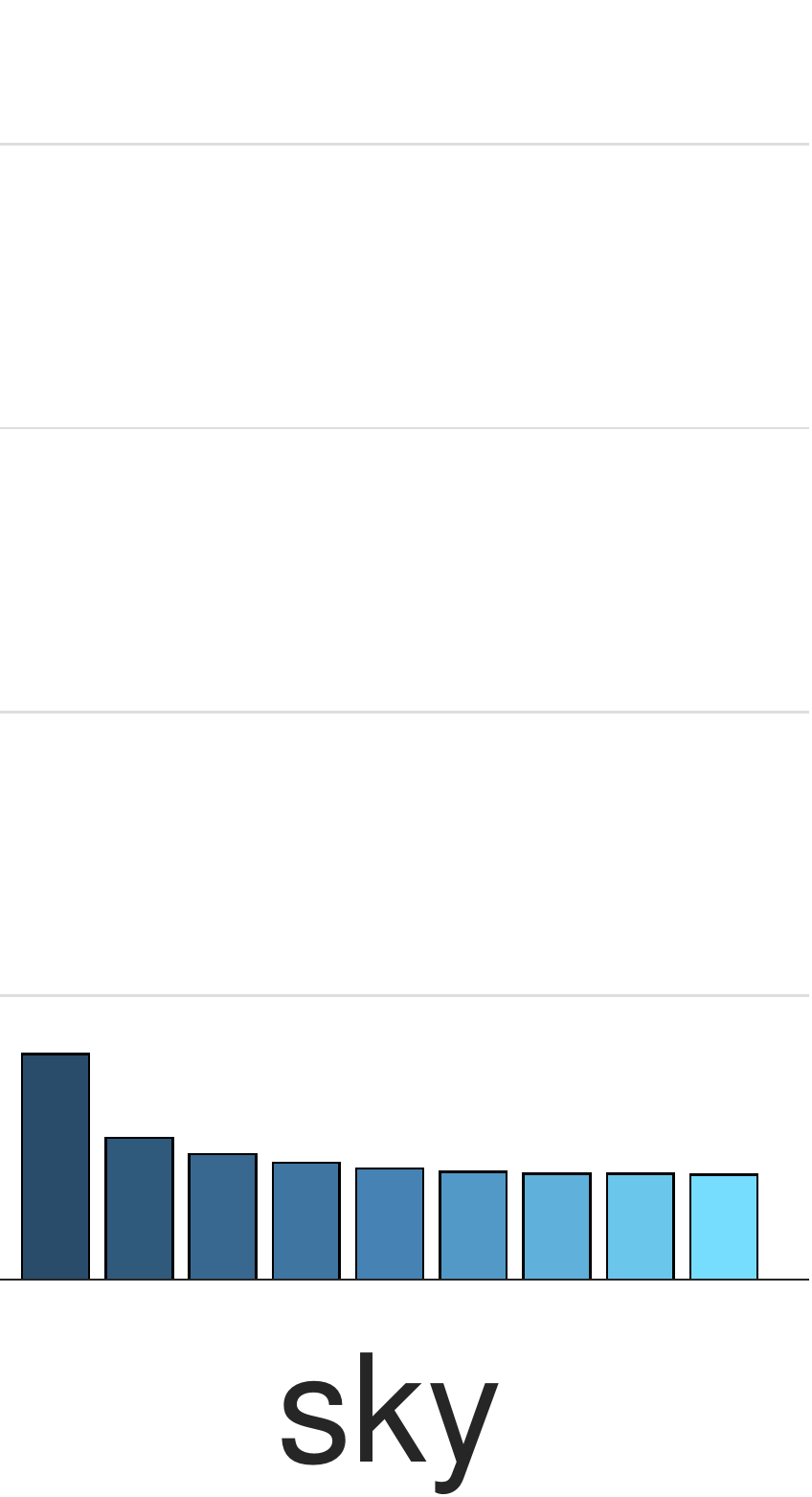} &
    \hspace{-0.45cm}\includegraphics[height=3cm]{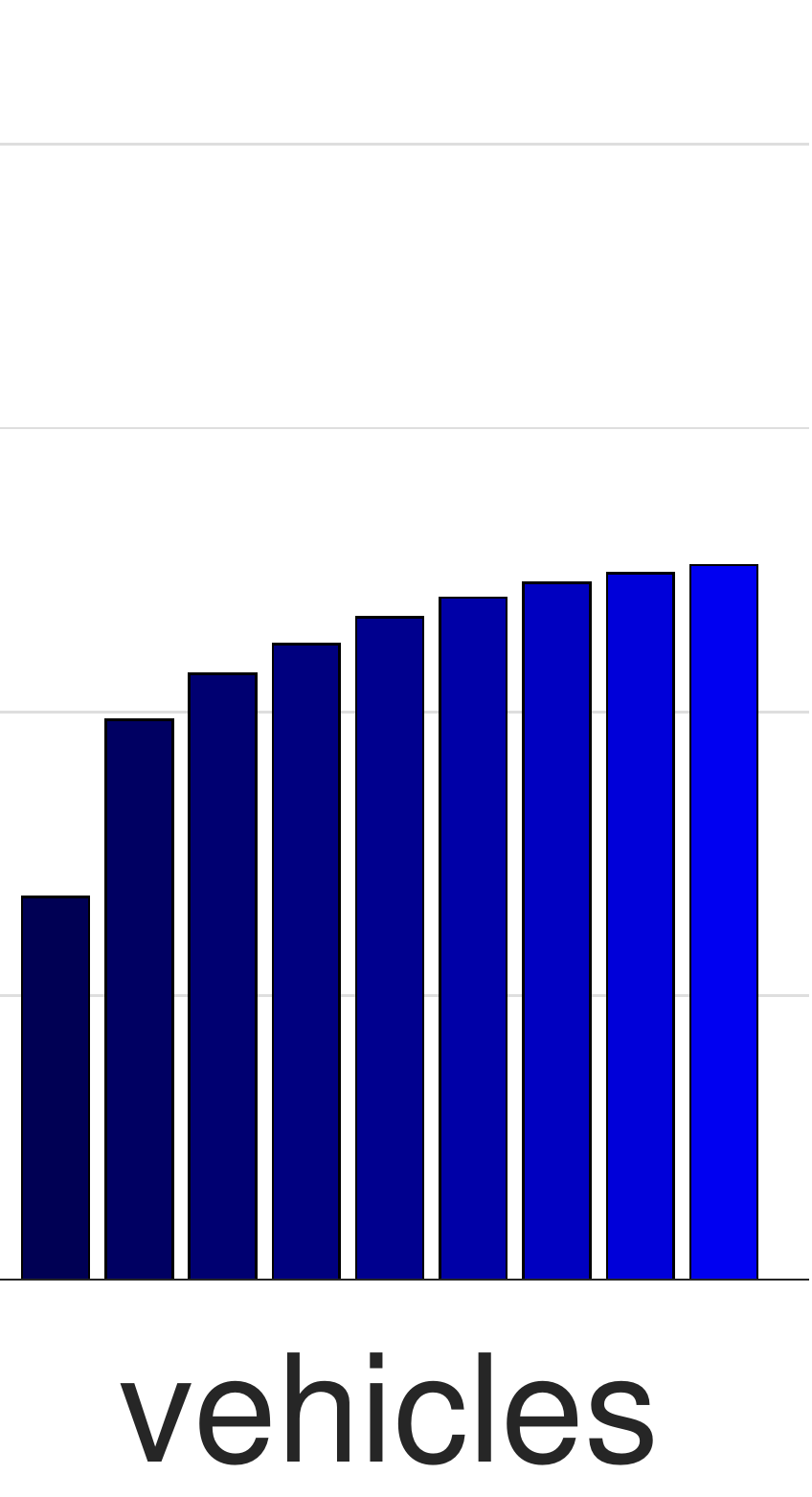} &
                     \includegraphics[height=3cm]{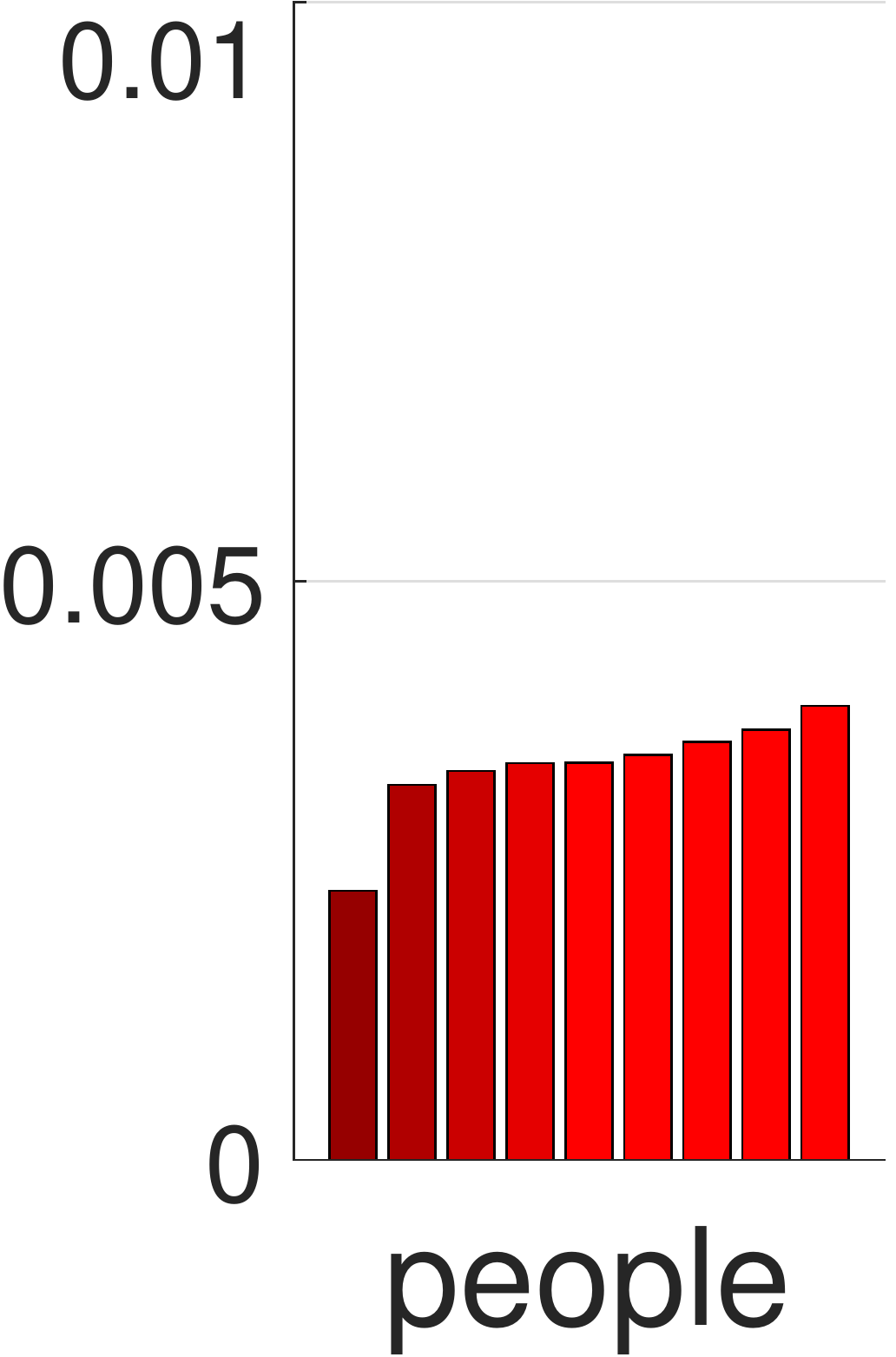} &
    \hspace{-0.45cm}\includegraphics[height=3cm]{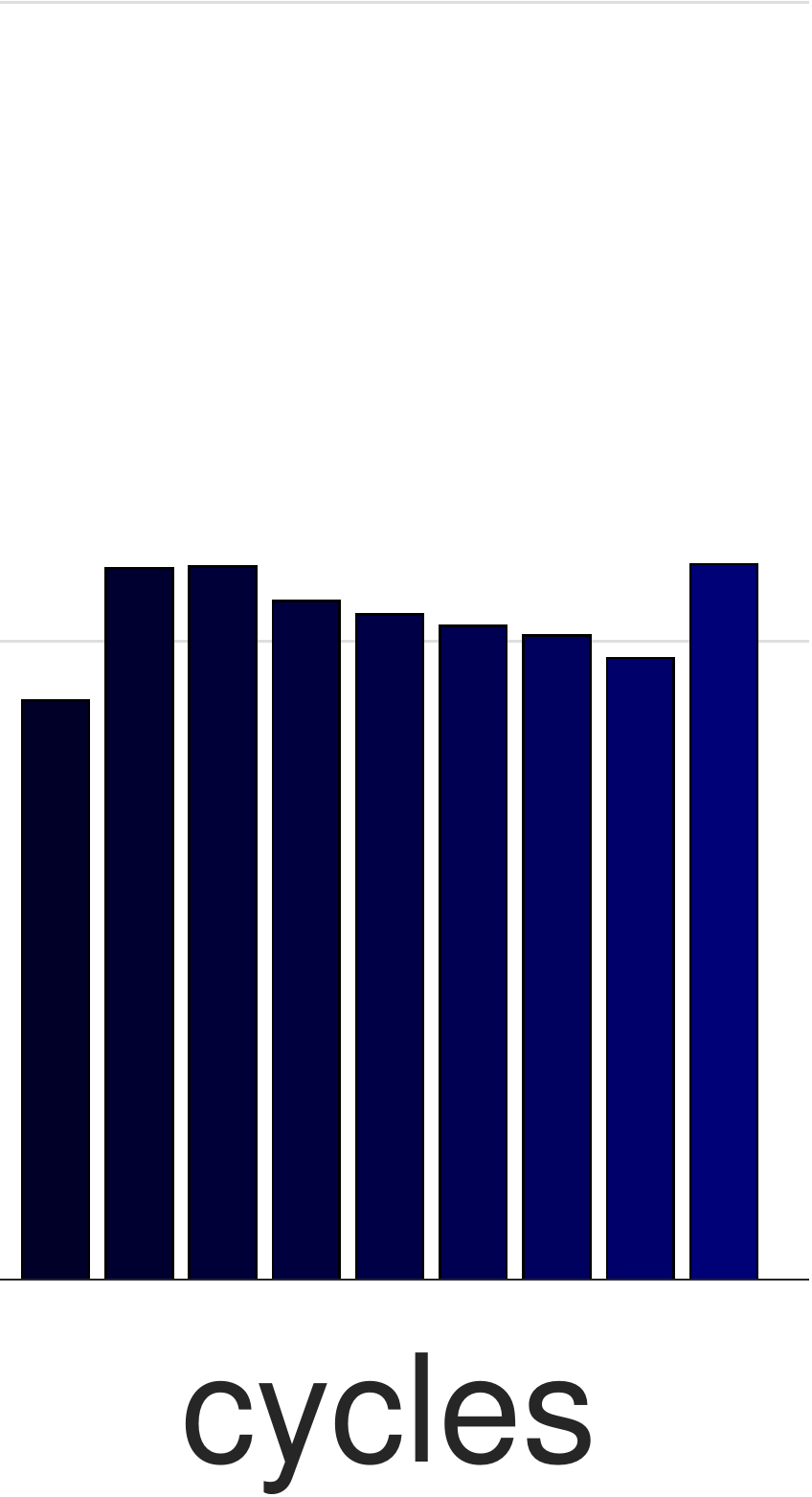}\\
    \end{tabular}
    \caption{Proportion of semantic categories that fall within the driver's fixation map when thresholded at increasing values (from left to right). Categories exhibiting a positive trend (\emph{e.g.} road and vehicles) suggest a real attention focus, while a negative trend advocates for an awareness of the object which is only circumstantial. See Sec.~\ref{sec:dataset_analysis} for details.}
    \label{fig:what_gt}
\end{figure*}
To enable the evaluation of computational models under such circumstances, the \drive~dataset has been extended with a set of further annotations. For each video, subsequences whose ground truth poorly correlates with the average ground truth of that sequence are selected. We employ Pearson's Correlation Coefficient ($CC$) and select subsequences with $\text{CC}<0.3$. This happens when the attention of the driver focuses far from the vanishing point of the road.
Examples of such subsequences are depicted in Fig.~\ref{fig:att_vs_inatt}. Several human annotators inspected the selected frames and manually split them into (a) acting, (b) inattentive, (c) errors and (d) subjective events:
\begin{itemize}
\item \emph{errors} can happen either due to failures in the measuring tool (\emph{e.g.} in extreme lighting conditions) or in the successive data processing phase (\emph{e.g.} SIFT matching);
\item \emph{inattentive} subsequences occur when the driver focuses his gaze on objects unrelated to the driving task (\emph{e.g.} looking at an advertisement);
\item \emph{subjective} subsequences describe situations in which the attention is closely related to the individual experience of the driver, \emph{e.g.} a road sign on the side might be an interesting element to focus for someone that has never been on that road before but might be safely ignored by someone who drives that road every day.
\item \emph{acting} subsequences include all the remaining ones. 
\end{itemize}
\emph{Acting} subsequences are particularly interesting as the deviation of driver's attention from the common central pattern denotes an intention linked to task-specific actions (\emph{e.g.} turning, changing lanes, overtaking \ldots). For these reasons, subsequences of this kind will have a central role in the evaluation of predictive models in Sec.~\ref{sec:exp1}.
\subsection{Dataset analysis}
\label{sec:dataset_analysis}
By analyzing the dataset frames, the very first insight is the presence of a strong attraction of driver's focus towards the vanishing point of the road, that can be appreciated in Fig.~\ref{fig:vanishing_point}. 
The same phenomenon was observed in previous studies~\cite{ueda2017eye, borji2016vanishing} in the context of visual search tasks. 
We observed indeed that drivers often tend to disregard road signals, cars coming from the opposite direction and  pedestrians on sidewalks.
This is an effect of human peripheral vision~\cite{sardegna2002encyclopedia}, that allows observers to still perceive and interpret stimuli out of - but sufficiently close to - their focus of attention (FoA). A driver can therefore achieve a larger area of attention by focusing on the road's vanishing point: due to the geometry of the road environment, many of the objects worth of attention are coming from there and have already been perceived when distant.\\
\begin{figure*}[ht]
\begin{center}
\includegraphics[width=0.85\textwidth]{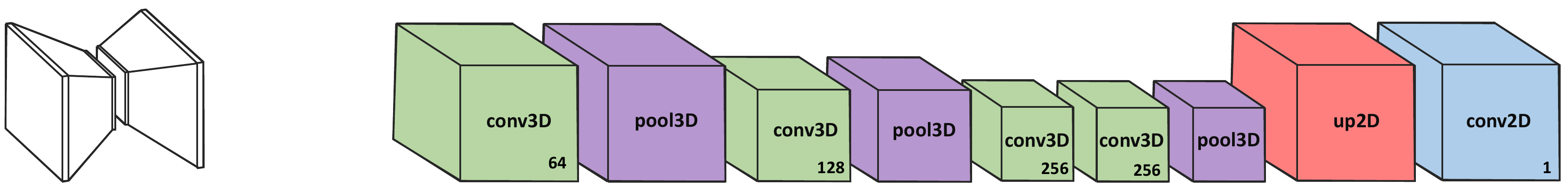}
\end{center}
   \caption{The \texttt{COARSE} module is made of an encoder based on C3D network~\cite{tran2014learning} followed by a bilinear upsampling (bringing representations back to the resolution of the input image) and a final 2D convolution. During feature extraction, the temporal axis is lost due to 3D pooling. All convolutional layers are preceded by zero paddings in order keep borders, and all kernels have size 3 along all dimensions. Pooling layers have size and stride of (1, 2, 2, 4) and (2, 2, 2, 1) along temporal and spatial dimensions respectively. All activations are ReLUs.}
\label{fig:coarse_module}
\end{figure*}
\begin{figure*}[ht]
\begin{center}
\includegraphics[width=\textwidth]{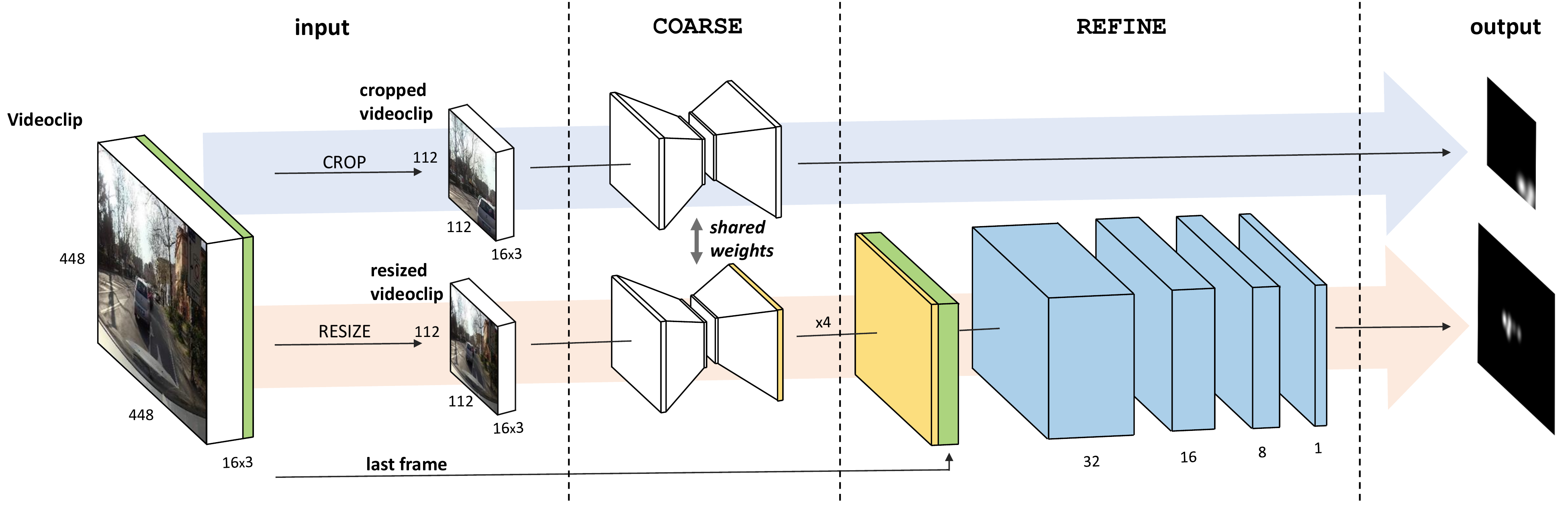}
\end{center}
   \caption{A single FoA branch of our prediction architecture. The \texttt{COARSE} module (see Fig.~\ref{fig:coarse_module}) is applied to both a cropped and a resized version of the input tensor, which is a videoclip of 16 consecutive frames. The cropped input is used during training to augment the data and the variety of ground truth fixation maps. The prediction of the resized input is stacked with the last frame of the videoclip and fed to a stack of convolutional layers (refinement module) with the aim of refining the prediction. Training is performed end-to-end and weights between \texttt{COARSE} modules are shared. At test time, only the refined predictions are used. Note that the complete model is composed of three of these branches (see Fig.~\ref{fig:multi_branch_model}), each of which predicting visual attention for different inputs (namely image, optical flow and semantic segmentation). All activations in the refinement module are LeakyReLU with $\alpha=10^{-3}$, except for the last single channel convolution that features ReLUs. Crop and resize streams are highlighted by light blue and orange arrows respectively.}
\label{fig:foa_branch}
\end{figure*}
Moreover, the gaze location tends to drift from this central attractor when the context changes in terms of car speed and landscape. Indeed~\cite{roge2004influence} suggests that our brain is able to compensate spatially or temporally dense information by reducing the visual field size. In particular, as the car travels at higher speed the temporal density of information (\textit{i.e.} the amount of information that the driver needs to elaborate per unit of time) increases: this causes the useful visual field of the driver to shrink~\cite{roge2004influence}. We also observe this phenomenon in our experiments, as shown in~ Fig.~\ref{fig:gt_across_speed}.\\
\drive~data also highlight that the driver's gaze is attracted towards specific semantic categories.
To reach the above conclusion, the dataset is analysed by means of the semantic segmentation model in \cite{yu2015multi} and the distribution of semantic classes within the fixation map evaluated.
More precisely, given a segmented frame and the corresponding fixation map, the probability for each semantic class to fall within the area of attention is computed as follows: First, the fixation map (which is continuous in $[0, 1]$) is normalized such that the maximum value equals 1. Then, nine binary maps are constructed by thresholding such continuous values linearly in the interval $[0, 1]$. As the threshold moves towards 1 (the maximum value), the area of interest shrinks around the real fixation points (since the continuous map is modeled by means of several Gaussians centered in fixation points, see previous section). For every threshold, a histogram over semantic labels within the area of interest is built, by summing up occurrences collected from all \drive~frames. Fig.~\ref{fig:what_gt} displays the result: for each class, the probability of a pixel to fall within the region of interest is reported for each threshold value. The figure provides insight about which categories represent the real focus of attention and which ones tend to fall inside the attention region just by proximity with the formers. Object classes that exhibit a positive trend, such as road, vehicles and people, are the real focus of the gaze, since the ratio of pixels classified accordingly increases when the observed area shrinks around the fixation point.
In a broader sense, the figure suggests that despite while driving our focus is dominated by road and vehicles, we often observe specific objects categories even if they contain little information useful to drive.
%%%%%%%%%%%%%%%%%%%%%%%%%%%%%%%%%%%%%%%%%%%%%%%%%%%%%%%%%%%%%%%%%%%%%%%%%%%%%%%%%%%%%%%%%%%%%%%%
%%%%%%%%%%%%%%%%%%%%%% MODEL %%%%%%%%%%%%%%%%%%%%%%%%%%%%%%%%%%%%%%%%%%%%%%%%%%%%%%%%%%%%%%%%%%%
%%%%%%%%%%%%%%%%%%%%%%%%%%%%%%%%%%%%%%%%%%%%%%%%%%%%%%%%%%%%%%%%%%%%%%%%%%%%%%%%%%%%%%%%%%%%%%%%
%
\section{Multi-Branch deep architecture for\\focus of attention prediction}
\label{sec:models}
The \drive~dataset is sufficiently large to allow the construction of a deep architecture to model common attentional patterns. Here, we describe our neural network model to predict human FoA while driving.
\\
\\
{\bf Architecture design.} In the context of high level video analysis (\emph{e.g.} action recognition and video classification), it has been shown that a method leveraging single frames can be outperformed if a sequence of frames is used as input instead~\cite{tran2014learning, karpathy2014large}. Temporal dependencies are usually modeled either by 3D convolutional layers~\cite{tran2014learning}, tailored to capture short range correlations, or by recurrent architectures (\emph{e.g.} LSTM, GRU), that can model longer term dependencies~\cite{baraldi2016hierarchical, pan2016hierarchical}.
Our model follows the former approach, relying on the assumption that a small time window (\emph{e.g.} half a second) holds sufficient contextual information for predicting where the driver would focus in that moment. Indeed, human drivers can take even less time to react to an unexpected stimulus. Our architecture takes a sequence of 16 consecutive frames ($\approx$ 0.65s) as input (called \emph{clips} from now on) and predicts the fixation map for the last frame of such clip.
\\
Many of the architectural choices made to design the network come from insights from the dataset analysis presented in Sec.\ref{sec:dataset_analysis}. In particular, we rely on the following results:
\begin{itemize}
    \item the drivers' FoA exhibits consistent patterns, suggesting that it can be reproduced by a computational model;
    \item the drivers' gaze is affected by a strong prior on objects semantics, \emph{e.g.} drivers tend to focus on items lying on the road;
    \item motion cues, like vehicle speed, are also key factors that influence gaze.
\end{itemize}
Accordingly, the model output merges three branches with identical architecture, unshared parameters and different input domains: the RGB image, the semantic segmentation and the optical flow field.
We call this architecture \texttt{multi-branch} model. 
%The whole architecture is called \texttt{multi-branch} model from now on and it is illustrated in Fig.~\ref{fig:multi_branch_model}. 
Following a bottom-up approach, in Sec.~\ref{sec:single_foa_branch} the building blocks of each branch are motivated and described. Later, in Sec.~\ref{sec:multi_branch} it will be shown how the branches merge into the final model.
%
%%%%%%%%%%%%%%%%%%%%%%%%%%%%%%%%%%%%%%%%%%%%%%%%%%%%%%%%%%%%%%%%%%%%%%%%%%%%%%%%%%%%%%%%%%%%%%%%
%%%%%% SUBSECTION - Single FoA branch
%%%%%%%%%%%%%%%%%%%%%%%%%%%%%%%%%%%%%%%%%%%%%%%%%%%%%%%%%%%%%%%%%%%%%%%%%%%%%%%%%%%%%%%%%%%%%%%%
\subsection{Single FoA branch}
\label{sec:single_foa_branch}
Each branch of the \texttt{multi-branch} model is a two-input two-output architecture composed of two intertwined streams.
The aim of this peculiar setup is to prevent the network from learning a central bias, that would otherwise stall the learning in early training stages~\footnote{For further details the reader can refer to Sec.~\ref{sup:shifts} and Sec.~\ref{sup:padding} of the supplementary material.}.
% The aim of this peculiar setup is to ameliorate the effect of the central bias, that would otherwise affect the model training.
% ADD SOMETHING LIKE THIS despite its fully convolutional architecture, the network can still learn an internal biased representation due to x1, x2 and x3.\footnote{see supplementary for details
To this end, one of the streams is given as input (output) a severely cropped portion of the original image (ground truth), ensuring a more uniform distribution of the true gaze, and runs through the \texttt{COARSE} module, described below.
Similarly, the other stream uses the \texttt{COARSE} module to obtain a rough prediction over the full resized image and then refines it through a stack of additional convolutions called \texttt{REFINE} model.
At test time, only the output of the \texttt{REFINE} stream is considered.
Both streams rely on the \texttt{COARSE} module, the convolutional backbone (with shared weights) which provides the rough estimate of the attentional map corresponding to a given clip. This component is detailed in Fig.~\ref{fig:coarse_module}.
\\
The \texttt{COARSE} module is based on the C3D architecture~\cite{tran2014learning} that encodes video dynamics by applying a 3D convolutional kernel on the 4D input tensor. 
As opposed to 2D convolutions that stride along the width and height dimension of the input tensor, a 3D convolution also strides along time. Formally, the $j$-th feature map in the
$i$-th layer at position $(x,y)$ at time $t$ is computed as: 
\begin{equation}
v_{i,j}^{x,y,t} = b_{i,j} + \sum_m \sum_{p=0}^{P_{i-1}} \sum_{q=0}^{Q_{i-1}} \sum_{r=0}^{R_{i-1}} w_{i,j,m}^{p,q,r}v_{i-1,m}^{x+p,y+q,t+r}  
\label{eq:c3d}
\end{equation}
where $m$ indexes different input feature maps, $w^{p,q,r}_{i,j,m}$ is the value at the position $(p, q)$ at time $r$ of the kernel connected to the $m$-th feature map, and $P_i$, $Q_i$ and $R_i$ are the dimensions of the kernel along width, height and temporal axis respectively; $b_{i,j}$ is the bias from layer $i$ to layer $j$.\\
From C3D, only the most general-purpose features are retained by removing the last convolutional layer and the fully connected layers which are strongly linked to the original action recognition task.
The size of the last pooling layer is also modified in order to cover the remaining temporal dimension entirely. This collapses the tensor from 4D to 3D, making the output independent of time.
Eventually, a bilinear upsampling brings the tensor back to the input spatial resolution and a 2D convolution merges all features into one channel. See Fig.~\ref{fig:coarse_module} for additional details on the \texttt{COARSE} module.
\\
\\
{\bf Training the two streams together} The architecture of a single FoA branch is depicted in Fig.~\ref{fig:foa_branch}. During training, the first stream feeds the \texttt{COARSE} network with random crops, forcing the model to learn the current focus of attention given visual cues rather than prior spatial location. 
The C3D training process described in~\cite{tran2014learning}, employs a $128\times128$ image resize, and then a $112\times112$ random crop. However, the small difference in the two resolutions limits the variance of gaze position in ground truth fixation maps and is not sufficient to avoid the attraction towards the center of the image.
For this reason, training images are resized to $256\times256$ before being cropped to $112\times112$. This crop policy generates samples that cover less than a quarter of the original image thus ensuring a sufficient variety in prediction targets. This comes at the cost of a coarser prediction: as crops get smaller, the ratio of pixels in the ground truth covered by gaze increases, leading the model to learn larger maps.\\
%
%%%%%%%%%%%%%%%%%%%%%%%%%%%%%%%%%%%%%%%%%%%%%%%%%%%%%%%%%%%%%%%%
%%%%%%%%%%%%%%%%%  Training Algorithm  %%%%%%%%%%%%%%%%%%%%%%%%%
%%%%%%%%%%%%%%%%%%%%%%%%%%%%%%%%%%%%%%%%%%%%%%%%%%%%%%%%%%%%%%%%
\algnewcommand{\LineComment}[1]{\Statex \(\triangleright\) #1}
\definecolor{comment_color}{rgb}{0.59, 0.78, 0.64}
\begin{algorithm*}
\caption{TRAINING. The model is trained in two steps: first each branch is trained separately through iterations detailed in \textbf{procedure} \textsc{single\_branch\_training\_iteration}, then the three branches are fine-tuned altogether as shown by \textbf{procedure} \textsc{multi\_branch\_fine-tuning\_iteration}. For clarity, we omit from notation: i) the subscript $b$ denoting the current domain in all $X$, $x$ and $\hat{y}$ variables in the single branch iteration and ii) the normalization of the sum of the outputs from each branch in line 13.}
\label{alg:train_algorithm}
\begin{algorithmic}[1]
%%%% Procedure A
\vspace{0.1cm}
\Procedure{{\bf A:} single\_branch\_training\_iteration}{}
\Statex \hspace{0.4cm} {\bf input:} domain data $X=\{x_1, x_2, \dots, x_{16}\}$, true attentional map $y$ of last frame $x_{16}$ of videoclip $X$
\Statex \hspace{0.4cm} {\bf output:} branch loss $\mathcal{L}_b$ computed on input sample $(X, y)$
\State $X_\text{res}$ $\gets$ \texttt{resize}($X$, (112, 112)) 
\State $X_\text{crop}$,~$y_\text{crop}$ $\gets$ \texttt{get\_crop}(($X$,~$y$), (112, 112))
\State $\hat{y}_\text{crop}$ $\gets$ \texttt{COARSE}($X_\text{crop}$) \hspace{5.60cm} {\color{comment_color} \# get coarse prediction on uncentered crop} 
\State $\hat{y}$ $\gets$ \texttt{REFINE}(\texttt{stack}($x_{16}$, \texttt{upsample}(\texttt{COARSE}($X_\text{res}$)))) \hspace{1.0cm} {\color{comment_color} \# get refined prediction over whole image}
\State $\mathcal{L}_b(X, Y) \gets D_{KL}(y_\text{crop} \| \hat{y}_\text{crop}) + D_{KL}(y \| \hat{y})$  \hspace{2.55cm} {\color{comment_color} \# compute branch loss as in Eq. \ref{eq:loss_foa_branch}}
\EndProcedure
\vspace{0.3cm}
%
%%%% Procedure B
\Procedure{{\bf B:} multi\_branch\_fine-tuning\_iteration}{}
\Statex \hspace{0.4cm} {\bf input:} data $X=\{x_{1}, x_{2}, \dots, x_{16}\}$ for all domains, true attentional map $y$ of last frame $x_{16}$ of videoclip $X$
\Statex \hspace{0.4cm} {\bf output:} overall loss $\mathcal{L}$ computed on input sample $(X, y)$
\State $X_\text{res} \gets \texttt{resize}(X, (112, 112))$
\State $X_\text{crop},~y_\text{crop} \gets \texttt{get\_crop}((X,~y), (112, 112))$
\For{branch $b \in \{\text{RGB}, \text{flow}, \text{seg}\}$}
\State $\hat{y}_{b_\text{crop}} \gets$ \texttt{COARSE}($X_{b_\text{crop}}$) \hspace{5.1cm} {\color{comment_color} \# as in line 4 of the above procedure} 
\State $\hat{y}_b$ $\gets$ \texttt{REFINE}(\texttt{stack}($x_{b_{16}}$, \texttt{upsample}(\texttt{COARSE}($X_{b_\text{res}}$)))) \hspace{0.25cm} {\color{comment_color} \# as in line 5 of the above procedure}
\EndFor
\State $\mathcal{L}(X, Y) \gets D_{KL}(y_\text{crop} \| \sum_b \hat{y}_{b_\text{crop}}) + D_{KL}(y \| \sum_b\hat{y}_b)$  \hspace{1.35cm} {\color{comment_color} \# compute overall loss as in Eq. \ref{eq:loss_finetuning}}
\EndProcedure
\end{algorithmic}
\end{algorithm*}
In contrast, the second stream feeds the same \texttt{COARSE} model with the same images, this time \emph{resized} to $112\times112$ -- and not cropped. The coarse prediction obtained from the \texttt{COARSE} model is then concatenated with the final frame of the input clip, \emph{i.e.} the frame corresponding to the final prediction. Eventually, the concatenated tensor goes through the \texttt{REFINE} module to obtain a higher resolution prediction of the FoA.\\
\noindent The overall two-stream training procedure for a single branch is summarized in Algorithm~\ref{alg:train_algorithm}.
\\
\\
{\bf Training objective} Prediction cost can be minimized in terms of Kullback-Leibler divergence: 
\begin{equation}
D_{KL}(Y\|\hat{Y}) = \sum_i Y(i)~\log\left(\epsilon + \frac{Y(i)}{\epsilon + \hat{Y}(i)}\right)
\end{equation}
\noindent where $Y$ is the ground truth distribution, $\hat{Y}$ is the prediction, the summation index $i$ spans across image pixels and $\epsilon$ is a small constant that ensures numerical stability\footnote{Please note that $D_{KL}$ inputs are always normalized to be a valid probability distribution despite this may be omitted in notation to improve equations readability.}.
Since each single FoA branch computes an error on both the cropped image stream and the resized image stream, the branch loss can be defined as:
\begin{equation}
\label{eq:loss_foa_branch}
\begin{aligned}
\mathcal{L}_{b}(\mathcal{X}_b, \mathcal{Y}) = \sum_m \biggl(&D_{KL}(\phi(Y^{m}) \| \mathcal{C}(\phi(X^{m}_b)))\ +\\
&D_{KL}(Y^{m}\|\mathcal{R}(\mathcal{C}(\psi(X^{m}_b)), X^m_b))) \biggr)
\end{aligned}
\end{equation}
where $\mathcal{C}$ and $\mathcal{R}$ denote \texttt{COARSE} and \texttt{REFINE} modules, $({X^{m}_b, Y^{m}})\in\mathcal{X}_b\times\mathcal{Y}$ is the $m$-th training example in the $b$-th domain (namely RGB, optical flow, semantic segmentation), and $\phi$ and $\psi$ indicate the crop and the resize functions respectively.
\\
\\
{\bf Inference step} While the presence of the $\mathcal{C}(\phi(X^m_b))$ stream is beneficial in training to reduce the spatial bias, at test time only the $\mathcal{R}(\mathcal{C}(\psi(X^m_b)), X^m_b))$ stream producing higher quality prediction is used. The outputs of such stream from each branch $b$ are then summed together, as explained in the following section.
%
%%%%%%%%%%%%%%%%%%%%%%%%%%%%%%%%%%%%%%%%%%%%%%%%%%%%%%%%%%%%%%%%%%%%%%%%%%%%%%%%%%%%%%%%%%%%%%%%
%%%%%%%%%% SUBSECTION - Multi-Branch model
%%%%%%%%%%%%%%%%%%%%%%%%%%%%%%%%%%%%%%%%%%%%%%%%%%%%%%%%%%%%%%%%%%%%%%%%%%%%%%%%%%%%%%%%%%%%%%%%
\subsection{Multi-Branch model}
\label{sec:multi_branch}
\begin{figure*}[t]
\begin{center}
\includegraphics[width=\textwidth]{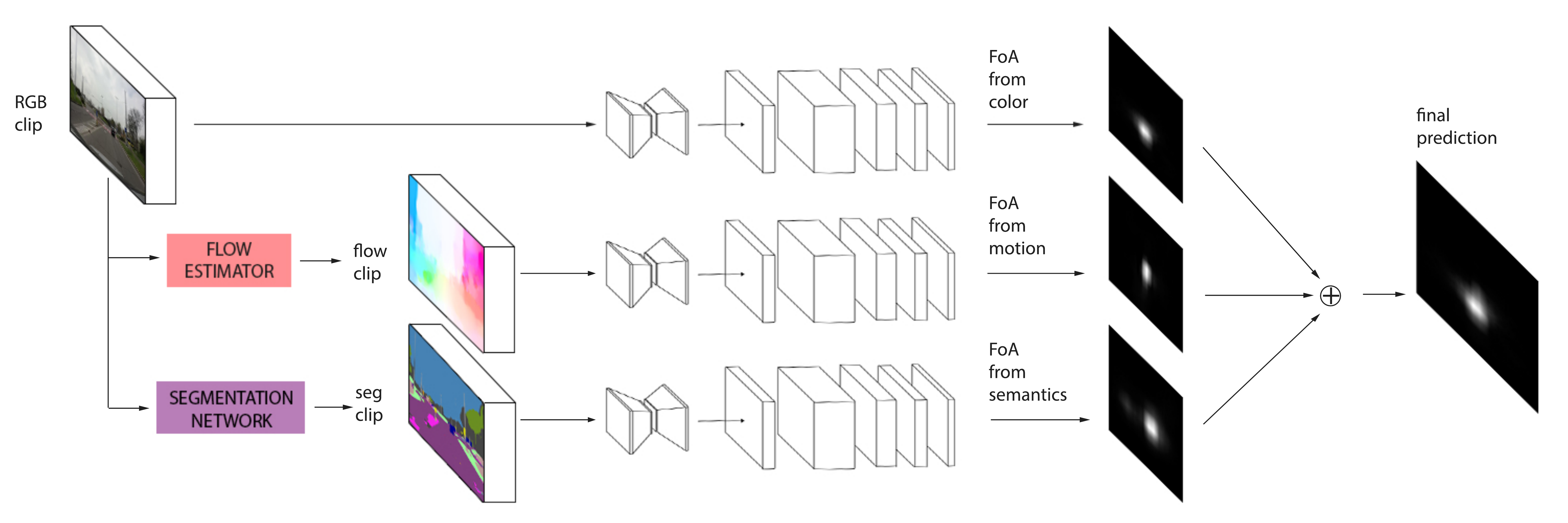}
\end{center}
   \caption{The \texttt{multi-branch} model is composed of three different branches, each of which has its own set of parameters, and their predictions are summed to obtain the final map. Note that in this figure cropped streams are dropped to ease representation, but are employed during training (as discussed in Sec.~\ref{sec:multi_branch} and depicted in Fig.~\ref{fig:foa_branch}.}
\label{fig:multi_branch_model}
\end{figure*}
\begin{algorithm}[b]
\caption{INFERENCE. At test time, the data extracted from the resized videoclip is input to the three branches and their output is summed and normalized to obtain the final FoA prediction.}
\label{alg:test_algorithm}
\begin{algorithmic}[1]
%%%% INFERENCE
\vspace{0.1cm}
\Statex \hspace{-0.5cm} {\bf input:} data $X=\{x_{1}, x_{2}, \dots, x_{16}\}$ for all domains
\Statex \hspace{-0.5cm} {\bf output:} predicted FoA map $\hat{y}$
\State $X_\text{res} \gets \texttt{\small resize}(X, (112, 112))$
\For{branch $b \in \{\text{RGB}, \text{flow}, \text{seg}\}$}
\State $\hat{y}_b \gets$ \texttt{\footnotesize REFINE}(\texttt{\small stack}($x_{b_{16}}$, \texttt{\small upsample}(\texttt{\footnotesize COARSE}($X_{b_\text{res}}$))))
\EndFor
\State $\hat{y} \gets \sum_b \hat{y}_b / \sum_i \sum_b \hat{y}_b(i)$
\end{algorithmic}
\end{algorithm}
As described at the beginning of this section and depicted in Fig.~\ref{fig:multi_branch_model}, the \texttt{multi-branch} model is composed of three identical branches. The architecture of each branch has already been described in Sec.~\ref{sec:single_foa_branch} above. Each branch exploits complementary information from a different domain and contributes to the final prediction accordingly. In detail, the first branch works in the RGB domain and processes raw visual data about the scene $X_\text{RGB}$. The second branch focuses on motion through the optical flow representation $X_\text{flow}$ described in \cite{gkamas2011guiding}. Eventually, the last branch takes as input semantic segmentation probability maps $X_\text{seg}$. For this last branch, the number of input channels depends on the specific algorithm used to extract the results, 19 in our setup (Yu and Koltun \cite{yu2015multi}). The three independent predicted FoA maps are summed and normalized to result in a probability distribution.\\
\noindent To allow for larger batch size, we choose to bootstrap each branch independently by training it according to Eq.~\ref{eq:loss_foa_branch}. Then, the complete \texttt{multi-branch} model which merges the three branches is fine-tuned with the following loss:
\begin{equation}
\label{eq:loss_finetuning}
\begin{aligned}
\mathcal{L}(\mathcal{X}, \mathcal{Y}) = \sum_m \biggl(&D_{KL}(\phi(Y^{m}) \| 
\sum_b \mathcal{C}(\phi(X^{m}_b)))\ +\\
&D_{KL}(Y^{m}\|\sum_b \mathcal{R}(\mathcal{C}(\psi(X^{m}_b)), X^m_b))) \biggr).
\end{aligned}
\end{equation}
\noindent The algorithm describing the complete inference over the \texttt{multi-branch} model in detailed in Alg.~\ref{alg:test_algorithm}.
%
%%%%%%%%%%%%%%%%%%%%%%%%%%%%%%%%%%%%%%%%%%%%%%%%%%%%%%%%%%%%%%%%
%%%%%%%%%%%%%%%%%%  Testing Algorithm  %%%%%%%%%%%%%%%%%%%%%%%%%
%%%%%%%%%%%%%%%%%%%%%%%%%%%%%%%%%%%%%%%%%%%%%%%%%%%%%%%%%%%%%%%%
%
%%%%%%%%%%%%%%%%%%%%%%%%%%%%%%%%%%%%%%%%%%%%%%%%%%%%%%%%%%%%%%%%%%%%%%
%%%%%%%%%%%%%%%%%%%%%%% EXPERIMENTS %%%%%%%%%%%%%%%%%%%%%%%%%%%%%%%%%%
%%%%%%%%%%%%%%%%%%%%%%%%%%%%%%%%%%%%%%%%%%%%%%%%%%%%%%%%%%%%%%%%%%%%%%
\section{Experiments}
\label{sec:exp1}
In this section we evaluate the performance of the proposed \texttt{multi-branch} model. First, we start by comparing our model against some baselines and other methods in literature. Following the guidelines in~\cite{bylinskii2016different}, for the evaluation phase we rely on Pearson's Correlation Coefficient ($CC$) and Kullback--Leibler Divergence ($D_{KL}$) measures. Moreover, we evaluate the Information Gain ($IG$)~\cite{kummerer2015information} measure to assess the quality of a predicted map $P$ with respect to a ground truth map $Y$ in presence of a strong bias, as:
\begin{equation}
IG(P,Y,B)=\frac{1}{N}\sum_{i} Y_i [( \log_2 (\epsilon + P_i ) - \log_2 ( \epsilon + B_i)]
\label{eq:ig}
\end{equation}
where $i$ is an index spanning all the $N$ pixels in the image, $B$ the bias computed as the average training fixation map and $\epsilon$ ensures numerical stability.\\
Furthermore, we conduct an ablation study to investigate how different branches affect the final prediction and how their mutual influence changes in different scenarios. We then study whether our model captures the attention dynamics observed in Sec.~\ref{sec:dataset_analysis}.
Eventually, we assess our model from a human perception perspective.\\
\\
{\bf Implementation details.} The three different pathways of the \texttt{multi-branch} model (namely FoA from color, from motion and from semantics) have been pre-trained independently using the same cropping policy of Sec.~\ref{sec:multi_branch} and minimizing the objective function in Eq.~\ref{eq:loss_foa_branch}. Each branch has been respectively fed with:
\begin{itemize}
    \tightlist
    \item $16$ frames clips in raw RGB color space;
    \item $16$ frames clips with optical flow maps, encoded as color images through the flow field encoding~\cite{gkamas2011guiding};
    \item $16$ frames clips holding semantic segmentation from~\cite{yu2015multi} encoded as $19$ scalar activation maps, one per segmentation class.
\end{itemize}
During individual branch pre-training clips were randomly mirrored for data augmentation. We employ Adam optimizer with parameters as suggested in the original paper~\cite{kingma2014adam}, with the exception of the learning rate that we set to $10^{-4}$. Eventually, batch size was fixed to 32 and each branch was trained until convergence. The \drive~dataset is split into train, validation and test set as follows: sequences 1-38 are used for training, sequences 39-74 for testing. The 500 frames in the middle of each training sequence constitute the validation set.\\
\bgroup
\setlength{\tabcolsep}{.16667em}
\begin{figure*}[htp]
\centering
\begin{tabular}{cccccc}
\textbf{Input frame} & \textbf{GT} & \textbf{\texttt{multi-branch}} & \cite{IV} & \cite{bazzani2017recurrent} & \cite{mlnet2016} \\
\includegraphics[width=0.16\textwidth]{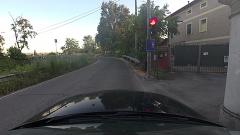}& \includegraphics[width=0.16\textwidth]{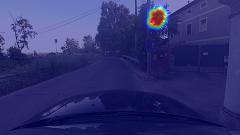}& \includegraphics[width=0.16\textwidth]{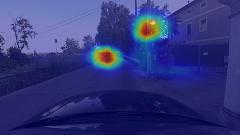}& \includegraphics[width=0.16\textwidth]{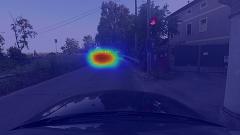}& \includegraphics[width=0.16\textwidth]{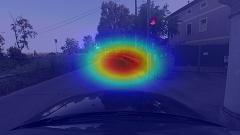}&
\includegraphics[width=0.16\textwidth]{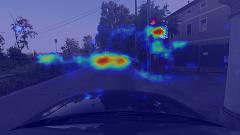}\\
\includegraphics[width=0.16\textwidth]{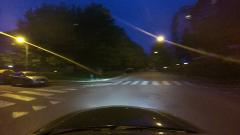}& \includegraphics[width=0.16\textwidth]{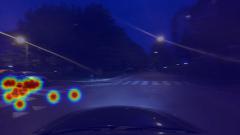}& \includegraphics[width=0.16\textwidth]{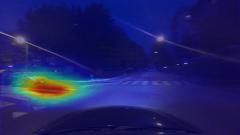}& \includegraphics[width=0.16\textwidth]{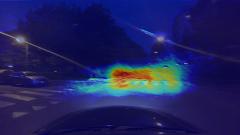}& \includegraphics[width=0.16\textwidth]{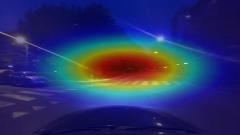}&
\includegraphics[width=0.16\textwidth]{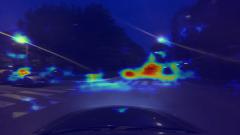}\\
\includegraphics[width=0.16\textwidth]{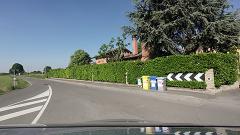}& \includegraphics[width=0.16\textwidth]{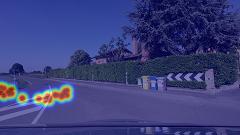}& \includegraphics[width=0.16\textwidth]{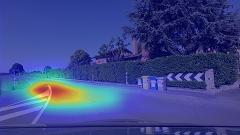}& \includegraphics[width=0.16\textwidth]{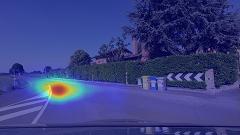}& \includegraphics[width=0.16\textwidth]{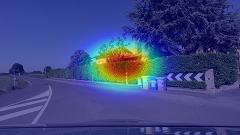}&
\includegraphics[width=0.16\textwidth]{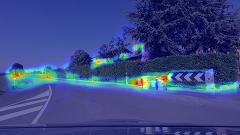}\\
\includegraphics[width=0.16\textwidth]{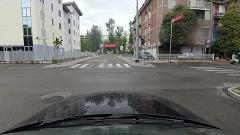}& \includegraphics[width=0.16\textwidth]{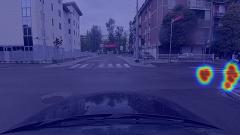}& \includegraphics[width=0.16\textwidth]{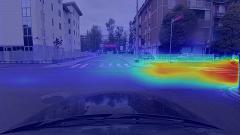}& \includegraphics[width=0.16\textwidth]{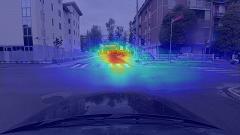}& \includegraphics[width=0.16\textwidth]{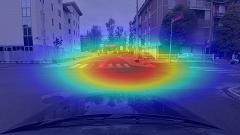}&
\includegraphics[width=0.16\textwidth]{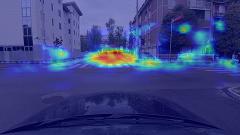}\\
\end{tabular}
\caption{Qualitative assessment of the predicted fixation maps. From left to right: input clip, ground truth map, our prediction, prediction of the previous version of the model \cite{IV}, prediction of RMDN~\cite{bazzani2017recurrent} and prediction of MLNet~\cite{mlnet2016}.}
\label{fig:comparison}
\end{figure*}
\egroup
Moreover, the complete \texttt{multi-branch} architecture was fine-tuned using the same cropping and data augmentation strategies minimizing cost function in Eq.~\ref{eq:loss_finetuning}. In this phase batch size was set to $4$ due to GPU memory constraints and learning rate value was lowered to $10^{-5}$. Inference time of each branch of our architecture is $\approx30$ milliseconds per videoclip on an NVIDIA Titan X.
\subsection{Model evaluation}
\label{sec:comparison}
\begin{table}[b]
\centering
\caption{Experiments illustrating the superior performance of the \texttt{multi-branch} model over several baselines and competitors. We report both the average across the complete test sequences and only the \emph{acting} frames.}
\begin{tabular}{r c c c c c c}
& \multicolumn{3}{c}{Test sequences} & \multicolumn{3}{c}{Acting subsequences} \\%\hline
& \multicolumn{1}{c}{$CC$}
& \multicolumn{1}{c}{$D_{KL}$}
& \multicolumn{1}{c}{$IG$}
& \multicolumn{1}{c}{$CC$}
& \multicolumn{1}{c}{$D_{KL}$} 
& \multicolumn{1}{c}{$IG$}\\
& \multicolumn{1}{c}{$\uparrow$}
& \multicolumn{1}{c}{$\downarrow$}
& \multicolumn{1}{c}{$\uparrow$}
& \multicolumn{1}{c}{$\uparrow$}
& \multicolumn{1}{c}{$\downarrow$}
& \multicolumn{1}{c}{$\uparrow$}\\ \hline
Baseline Gaussian & 0.40 & 2.16 & -0.49 & 0.26 & 2.41 & 0.03  \\
Baseline Mean & 0.51 & 1.60 & 0.00 & 0.22 & 2.35 & 0.00 \\ \hline 
Mathe \emph{et al.}\cite{MatheSminchisescuPAMI2015}  & 0.04 & 3.30 & -2.08 & - & - & -\\ 
Wang \emph{et al.}\cite{wang2015saliency}   & 0.04 & 3.40 & -2.21 & - & - & -\\ 
Wang \emph{et al.}\cite{wang2015consistent}   & 0.11 & 3.06 & -1.72 & - & - & -\\ 
MLNet\cite{mlnet2016}  & 0.44 & 2.00 & -0.88 & 0.32 & 2.35 & -0.36   \\ 
RMDN\cite{bazzani2017recurrent} & 0.41 & 1.77 & -0.06 & 0.31 & 2.13 & 0.31 \\
Palazzi \emph{et al.}\cite{IV}   & 0.55 & 1.48 & -0.21 & 0.37 & 2.00 & 0.20  \\  \hline
\texttt{multi-branch} & \textbf{0.56} & \textbf{1.40} & \textbf{0.04} & \textbf{0.41} & \textbf{1.80} & \textbf{0.51}
\end{tabular}
\label{tab:pred_results}
\end{table}
In Tab.~\ref{tab:pred_results} we report results of our proposal against other state-of-the-art models~\cite{wang2015saliency, MatheSminchisescuPAMI2015, mlnet2016, IV, bazzani2017recurrent, wang2015consistent} evaluated both on the complete test set and on \emph{acting} subsequences only.
All the competitors, with the exception of \cite{IV} are bottom-up approaches and mainly rely on appearance and motion discontinuities. To test the effectiveness of deep architectures for saliency prediction we compare against the Multi-Level Network (MLNet)~\cite{mlnet2016}, which scored favourably in the \texttt{MIT300} saliency benchmark~\cite{mit-saliency-benchmark}, and the Recurrent Mixture Density Network (RMDN)~\cite{bazzani2017recurrent}, which represents the only deep model addressing video saliency. While MLNet works on images discarding the temporal information, RMDN encodes short sequences in a similar way to our \texttt{COARSE} module, and then relies on a LSTM architecture to model long term dependencies and estimates the fixation map in terms of a GMM. To favor the comparison, both models were re-trained on the \drive~dataset.\\
Results highlight the superiority of our \texttt{multi-branch} architecture on all test sequences. The gap in performance with respect to bottom-up unsupervised approaches~\cite{wang2015saliency,wang2015consistent} is higher, and is motivated by the peculiarity of the attention behavior within the driving context, which calls for a task-oriented training procedure. Moreover, MLNet's low performance testifies for the need of accounting for the temporal correlation between consecutive frames that distinguishes the tasks of attention prediction in images and videos. Indeed, RMDN processes video inputs and outperforms MLNet on both $D_{KL}$ and $IG$ metrics, performing comparably on $CC$. 
% Nonetheless, its performance is still limited: this indicates that long term dependencies which RMDN is designed to capture are hardly beneficial for this task.
Nonetheless, its performance is still limited: indeed, qualitative results reported in Fig.~\ref{fig:comparison} suggest that long term dependencies captured by its recurrent module lead the network towards the regression of the mean, discarding contextual and frame-specific variations that would be preferrable to keep. To support this intuition, we measure the average $D_{KL}$ between RMDN predictions and the mean training fixation map (Baseline Mean), resulting in a value of 0.11. Being lower than the divergence measured with respect to groundtruth maps, this value highlights the closer correlation to a central baseline rather than to groundtruth.
Eventually, we also observe improvements with respect to our previous proposal~\cite{IV}, that relies on a more complex backbone model (also including a deconvolutional module) and processes RGB clips only. 
%Notably, the boost is particularly evident in \textit{acting} sub-sequences.
The gap in performance resides in the greater awareness of our \texttt{multi-branch} architecture of the aspects that characterize the driving task as emerged from the analysis in Sec.~\ref{sec:dataset_analysis}. The positive performances of our model are also confirmed when evaluated on the \emph{acting} partition of the dataset. We recall that \emph{acting} indicates sub-sequences exhibiting a significant task-driven shift of attention from the center of the image (Fig.~\ref{fig:att_vs_inatt}). Being able to predict the FoA also on \emph{acting} sub-sequences means that the model captures the strong centered attention bias but is capable of generalizing when required by the context.
% Some successful prediction cases are illustrated in Fig.~\ref{fig:comparison}.
\\
This is further shown by the comparison against a centered Gaussian baseline (BG) and against the average of all training set fixation maps (BM). The former baseline has proven effective on many image saliency detection tasks~\cite{mit-saliency-benchmark} while the latter represents a more task-driven version.
The superior performance of the \texttt{multi-branch} model w.r.t. baselines highlights that despite the attention is often strongly biased towards the vanishing point of the road, the network is able to deal with sudden task-driven changes in gaze direction.
\begin{figure}[th]
    \centering
    \includegraphics[width=\columnwidth]{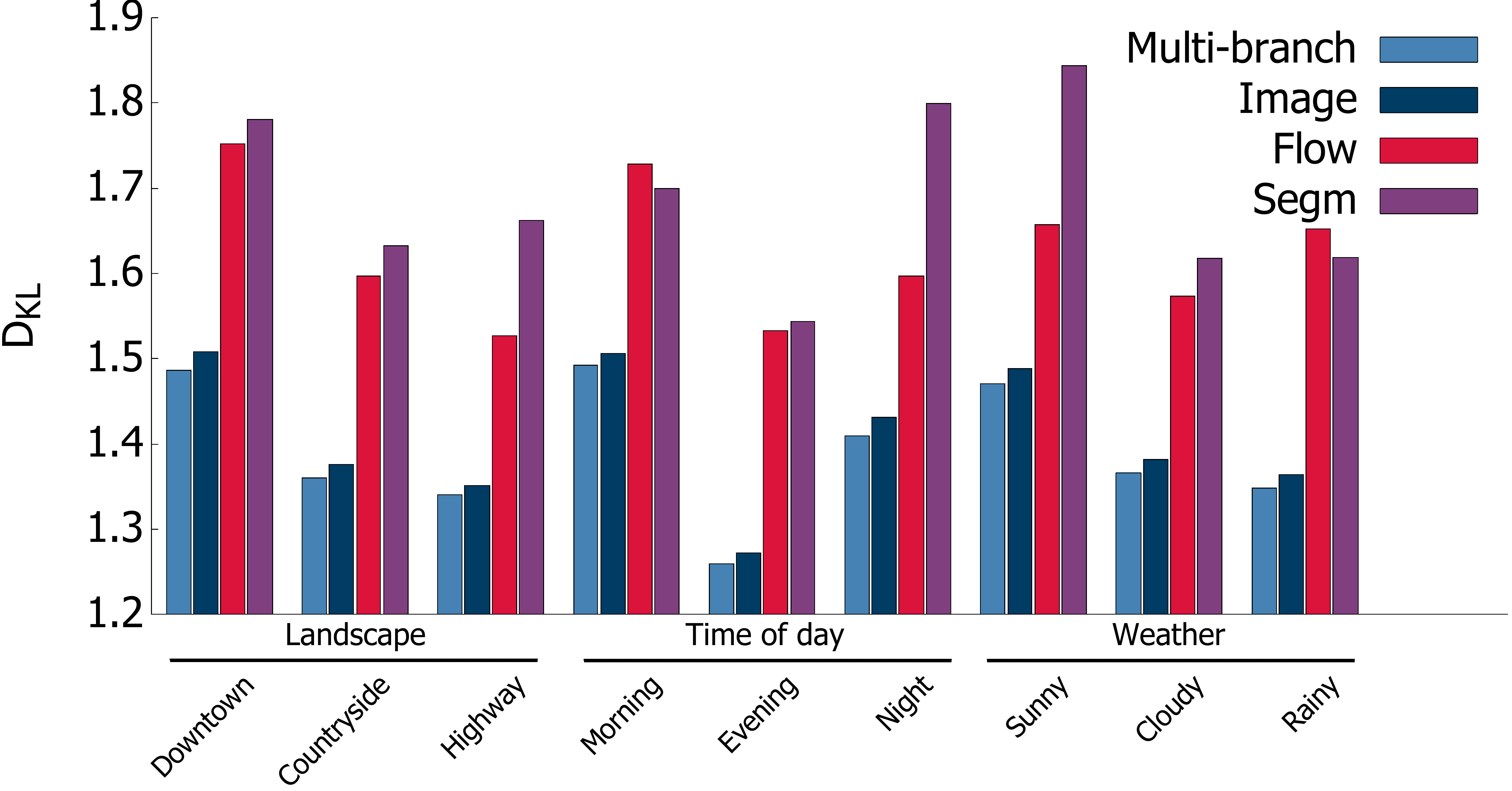}
    \caption{$D_{KL}$ of the different branches in several conditions (from left to right: downtown, countryside, highway, morning, evening, night, sunny, cloudy, rainy). Underlining highlights difference of aggregation in terms of landscape, time of day and weather. Please note that lower $D_{KL}$ indicates better predictions.}
    \label{fig:metric_by_scenario}
\end{figure}
\subsection{Model analysis}
\begin{figure*}[t]
    \centering
    \begin{tabular}{ccccc}
    $|\Sigma|=2.50\times 10^9$&
    $|\Sigma|=1.57\times 10^9$&
    $|\Sigma|=3.49\times 10^8$&
    $|\Sigma|=1.20\times 10^8$&
    $|\Sigma|=5.38\times 10^7$\\
    \includegraphics[width=0.18\textwidth]{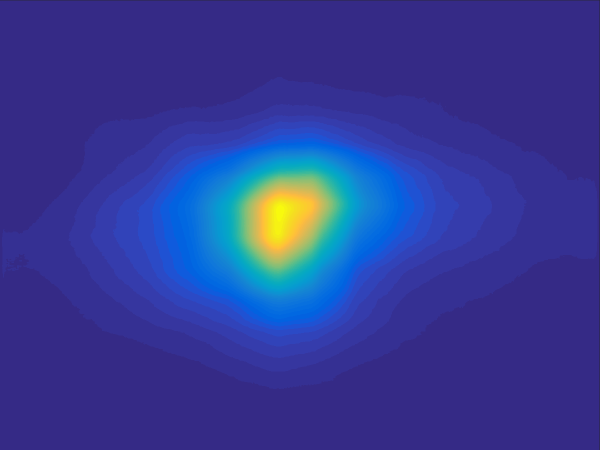} &
    \includegraphics[width=0.18\textwidth]{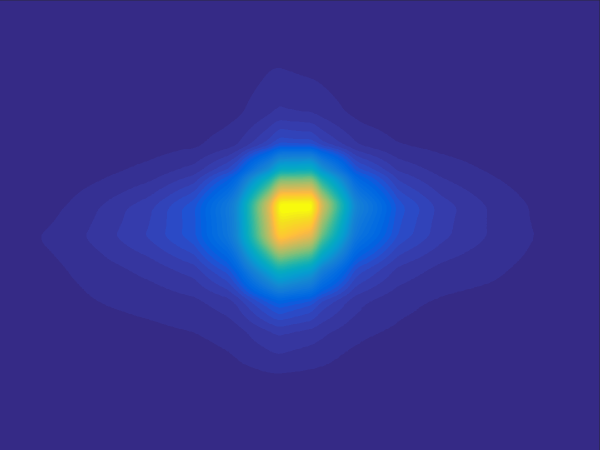} &
    \includegraphics[width=0.18\textwidth]{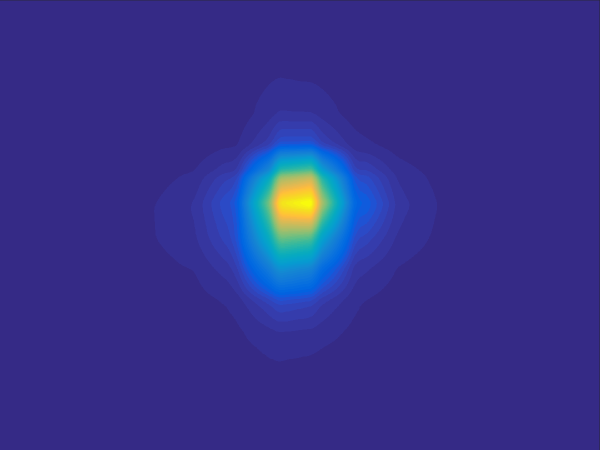} &
    \includegraphics[width=0.18\textwidth]{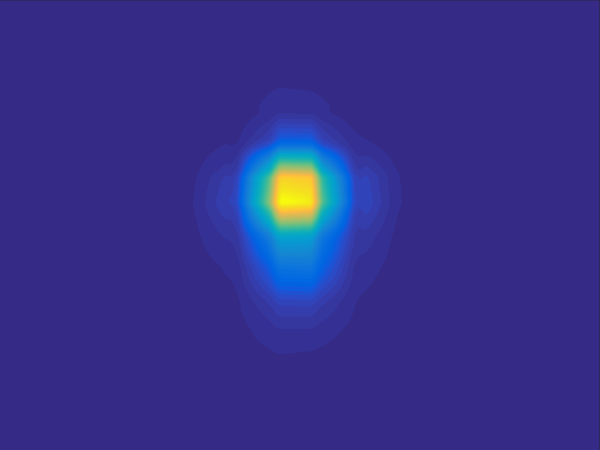} &
    \includegraphics[width=0.18\textwidth]{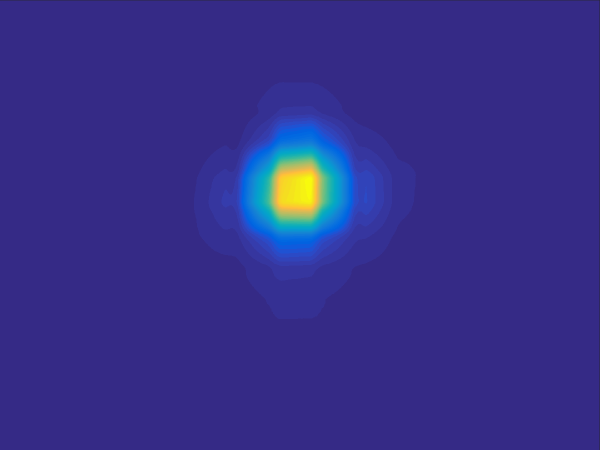}\\
    (a) $0\leq \mathrm{km/h}\leq10$ &
    (b) $10\leq \mathrm{km/h}\leq30$ &
    (c) $30\leq \mathrm{km/h}\leq50$ &
    (d) $50\leq \mathrm{km/h}\leq70$ &
    (e) $70\leq \mathrm{km/h}$
    \end{tabular}
    \hspace{0.2cm}
    \caption{Model prediction averaged across all test sequences and grouped by driving speed. As the speed increases, the area of the predicted map shrinks, recalling the trend observed in ground truth maps. As in Fig.~\ref{fig:gt_across_speed}, for each map a two dimensional Gaussian is fitted and the determinant of its covariance matrix $\Sigma$ is reported as a measure of the spread.}
    \label{fig:pred_across_speed}
\end{figure*}
\begin{figure}[t]
    \centering
	    \includegraphics[width=\columnwidth]{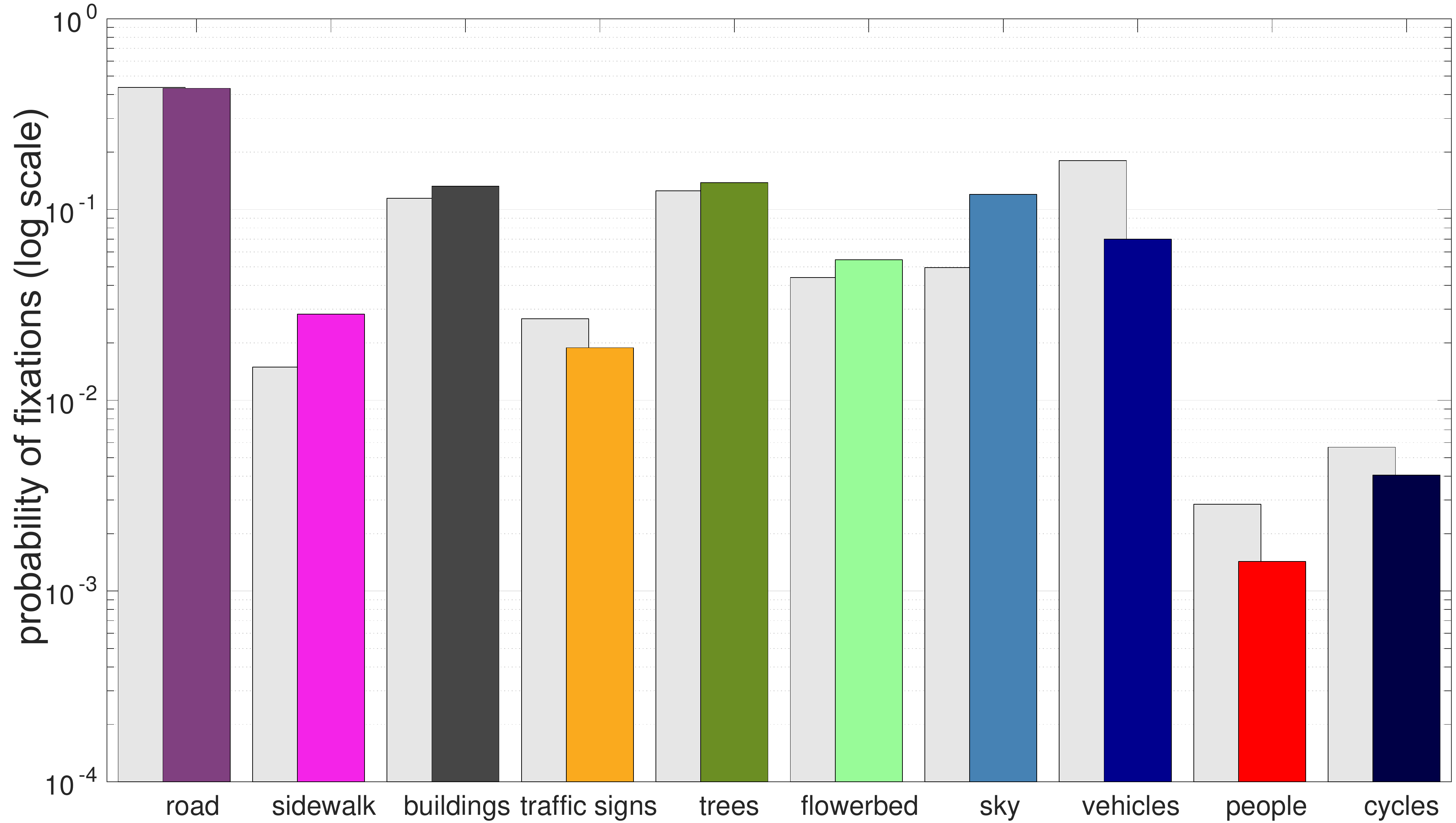}
    \caption{Comparison between ground truth (gray bars) and predicted fixation maps (colored bars) when used to mask semantic segmentation of the scene. The probability of fixation (in log-scale) for both ground truth and model prediction is reported for each semantic class. Despite absolute errors exist, the two bar series agree on the relative importance of different categories.}
    \label{fig:semantic_prediction}
\end{figure}
\label{sec:model_analysis}
In this section we investigate the behavior of our proposed model under different landscapes, time of day and weather (Sec.~\ref{sec:conditions}); we study the contribution of each branch to the FoA prediction task (Sec.~\ref{sec:ablation}); and we compare the learnt attention dynamics against the one observed in the human data (Sec.~\ref{sec:attention-dynamics}).
\subsubsection{Dependency on driving environment}
\label{sec:conditions}
The \drive~data has been recorded under varying landscapes, time of day and weather conditions. We tested our model in all such different driving conditions. As would be expected, Fig.~\ref{fig:metric_by_scenario} shows that the human attention is easier to predict in highways rather than downtown, where the focus can shift towards more distractors. The model seems more reliable in evening scenarios, rather than morning or night, where we observed better lightning conditions and lack of shadows, over-exposure and so on.
Lastly, in rainy conditions we notice that human gaze is easier to model, possibly due to the higher level of awareness demanded to the driver and his consequent inability to focus away from vanishing point. To support the latter intuition, we measured the performance of BM baseline (\emph{i.e.} the average training fixation map), grouped for weather condition. As expected, the $D_{KL}$ value in rainy weather ($1.53$) is significantly lower than the ones for cloudy ($1.61$) and sunny weather ($1.75$), highlighting that when rainy the driver is more focused on the road.

\subsubsection{Ablation study}
\label{sec:ablation}
\begin{table}[b]
\centering
\caption{The ablation study performed on our \texttt{multi-branch} model. I, F and S represent image, optical flow and semantic segmentation branches respectively.}
\begin{tabular}{r c c c c c c}
& \multicolumn{3}{c}{Test sequences} & \multicolumn{3}{c}{Acting subsequences} \\
& $CC$ & $D_{KL}$ & $IG$ & $CC$ & $D_{KL}$ & $IG$ \\
& \multicolumn{1}{c}{$\uparrow$}
& \multicolumn{1}{c}{$\downarrow$}
& \multicolumn{1}{c}{$\uparrow$}
& \multicolumn{1}{c}{$\uparrow$}
& \multicolumn{1}{c}{$\downarrow$}
& \multicolumn{1}{c}{$\uparrow$}\\ \hline
I & 0.554 & 1.415 &-0.008 & 0.403 & 1.826 & 0.458\\
F & 0.516 & 1.616 & -0.137 & 0.368 & 2.010 & 0.349 \\
S & 0.479 & 1.699 & -0.119 & 0.344 & 2.082 & 0.288 \\
I+F & 0.558 & 1.399 & 0.033 & {\bf 0.410} & 1.799 & 0.510\\
I+S & 0.554 & 1.413 & -0.001 & 0.404 & 1.823 & 0.466\\ 
F+S & 0.528 & 1.571 & -0.055 & 0.380 & 1.956 & 0.427 \\\hline
I+F+S & \textbf{0.559} & \textbf{1.398} & \textbf{0.038} & {\bf 0.410} & \textbf{1.797} & \textbf{0.515} \\
\end{tabular}
\label{tab:ablation}
\end{table}
In order to validate the design of the \texttt{multi-branch} model (see Sec.~\ref{sec:multi_branch}), here we study the individual contributions of the different branches by disabling one or more of them.\\
Results in Tab.~\ref{tab:ablation} show that the RGB branch plays a major role in FoA prediction. The motion stream is also beneficial and provides a slight improvement, that becomes clearer in the \emph{acting} subsequences. Indeed, optical flow intrinsically captures a variety of peculiar scenarios that are non-trivial to classify when only color information is provided, \emph{e.g.}~when the car is still at a traffic light or is turning. The semantic stream, on the other hand, provides very little improvement. In particular, from Tab.~\ref{tab:ablation} and by specifically comparing I+F and I+F+S, a slight increase in the $IG$ measure can be appreciated.
Nevertheless, such improvement has to be considered negligible when compared to color and motion, suggesting that in presence of efficiency concerns or real-time constraints the semantic stream can be discarded with little losses in performance.
However, we expect the benefit from this branch to increase as more accurate segmentation models will be released.
\subsubsection{Do we capture the attention dynamics?}
\label{sec:attention-dynamics}
The previous sections validate quantitatively the proposed model. Now, we assess its capability to attend like a human driver by comparing its predictions against the analysis performed in Sec.~\ref{sec:dataset_analysis}.\\
First, we report the average predicted fixation map in several speed ranges in Fig.~\ref{fig:pred_across_speed}. The conclusions we draw are twofold: i) generally, the model succeeds in modeling the behavior of the driver at different speeds, and ii) as the speed increases fixation maps exhibit lower variance, easing the modeling task, and prediction errors decrease.\\
\noindent We also study how often our model focuses on different semantic categories, in a fashion that recalls the analysis of Sec.~\ref{sec:dataset_analysis}, but employing our predictions rather than ground truth maps as focus of attention. More precisely, we normalize each map so that the maximum value equals 1, and apply the same thresholding strategy described in Sec.~\ref{sec:dataset_analysis}. Likewise, for each threshold value a histogram over class labels is built, by accounting all pixels falling within the binary map for all test frames. This results in nine histograms over semantic labels, that we merge together by averaging probabilities belonging to different threshold. 
Fig.~\ref{fig:semantic_prediction} shows the comparison. Color bars represent how often the predicted map focuses on a certain category, while gray bars depict ground truth behavior and are obtained by averaging histograms in Fig.~\ref{fig:what_gt} across different thresholds. Please note that, to highlight differences for low populated categories, values are reported on a logarithmic scale. The plot shows a certain degree of absolute error is present for all categories. However, in a broader sense, our model replicates the relative weight of different semantic classes while driving, as testified by the importance of roads and vehicles, that still dominate, against other categories such as people and cycles that are mostly neglected. This correlation is confirmed by Kendall rank coefficient, which scored $0.51$ when computed on the two bar series.
\subsection{Visual assessment of predicted fixation maps}
\label{sec:qualitative_assessment}
\begin{figure}[tb]
    \centering
    \begin{tabular}{c}
    \hspace{-0.0cm}\includegraphics[width=0.95\columnwidth]{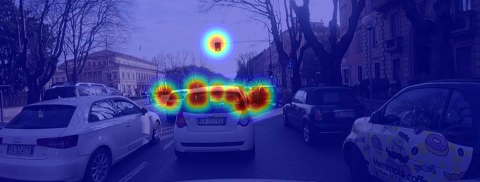}\\
    \hspace{-0.0cm}\includegraphics[width=0.95\columnwidth]{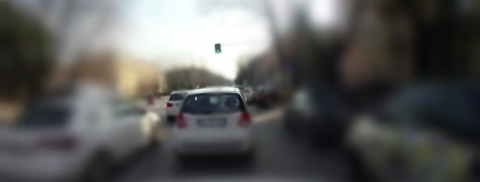}\\
    \end{tabular}
    \caption{The figure depicts a videoclip frame that underwent the foveation process. The attentional map (above) is employed to blur the frame in a way that approximates the foveal vision of the driver\cite{perry2002gaze}. In the foveated frame (below), it can be appreciated how the ratio of high-level information smoothly degrades getting farther from fixation points.}
    \label{fig:visual_assessment}
\end{figure}

%To further validate the predictions of our model from the human perception perspective, the following visual assessment is set up.
To further validate the predictions of our model from the human perception perspective, 50 people with at least 3 years of driving experience were asked to participate in a visual assessment\footnote{These were students (11 females, 39 males) of age between 21 and 26 ($\mu=23.4, \sigma=1.6$) recruited at our University on a voluntary basis through an online form.}. 
% [23, 21, 21, 22, 24, 24, 25, 26, 25, 24, 25, 22, 24, 24, 23, 22, 22, 25, 25, 21, 25, 21, 21, 22, 24, 21, 21, 25, 24, 25, 24, 23, 23, 23, 23, 25, 21, 25, 25, 26, 22, 23, 21, 22, 22, 22, 25, 27, 23, 25]
First, a pool of 400 videoclips (40 seconds long) is sampled from the \drive~dataset. Sampling is weighted such that resulting videoclips are evenly distributed among different scenarios, weathers, drivers and daylight conditions. Also, half of these videoclips contain sub-sequences that were previously annotated as \textit{acting}.\\
%Sampled videoclips are then pre-processed in order to approximate the field of attention of the driver (see below).\\
%
To approximate as realistically as possible the visual field of attention of the driver, sampled videoclips are pre-processed following the procedure in \cite{wang2017central}. As in~\cite{wang2017central} we leverage the \textit{Space Variant Imaging Toolbox}~\cite{perry2002gaze} to implement this phase, setting the parameter that halves the spatial resolution every 2.3$^{\circ}$ to mirror human vision~\cite{wang2017central,larson2009contributions}. The resulting videoclip preserves details near to the fixation points in each frame, whereas the rest of the scene gets more and more blurred getting farther from fixations until only low-frequency contextual information survive. Coherently  with~\cite{wang2017central} we refer to this process as \textit{foveation} (in analogy with human foveal vision). Thus, pre-processed videoclips will be called \textit{foveated videoclips} from now on. To appreciate the effect of this step the reader is referred to Fig.~\ref{fig:visual_assessment}.\\
Foveated videoclips were created by randomly selecting one of the following three fixation maps: the ground truth fixation map (G videoclips), the fixation map predicted by our model (P videoclips) or the average fixation map in the \drive~training set (C videoclips). The latter central baseline allows to take into account the potential preference for a "stable" attentional map (\emph{i.e.} lack of switching of focus). Further details about the creation of foveated videoclips are reported in Sec.~\ref{sup:foveation} of the supplementary material.\\
% To be fair to the ground truth, all the selected videoclips contain less than 0.5\% of frames marked as errors.
%
\begin{figure}[t]
     \centering
     \includegraphics[width=0.7\columnwidth]{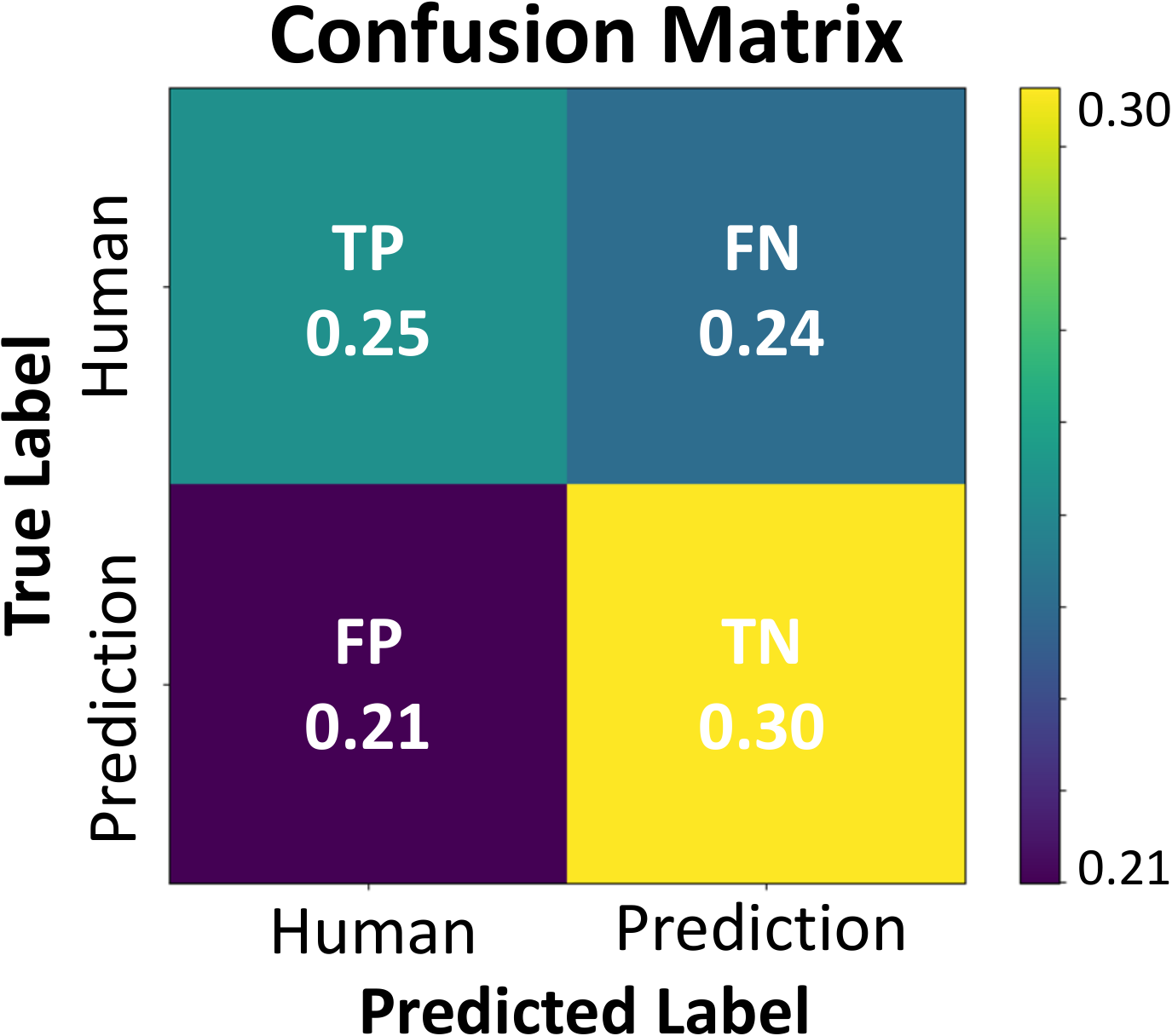}
     \caption{The confusion matrix reports the results of participants' guesses on the source of fixation maps. Overall accuracy is about 55\% which is fairly close to random chance.}
     \label{fig:visual_assessment_confusion_matrix}
\end{figure}
\noindent 
%Once created the set of foveated videoclips as above, the visual assessment could start. 
Each participant was asked to watch five randomly sampled foveated videoclips. After each videoclip, he answered the following question:
\begin{itemize}
    \item Would you say the observed attention behavior comes from a human driver? (yes/no)
\end{itemize}
Each of the 50 participant evaluates five foveated videoclips, for a total of 250 examples.\\
\noindent The confusion matrix of provided answers is reported in Fig.~\ref{fig:visual_assessment_confusion_matrix}. Participants were not particularly good at discriminating between human's gaze and model generated maps, scoring about the 55\% of accuracy which is comparable to random guessing; this suggests our model is capable of producing plausible attentional patterns that resemble a proper driving behavior to a human observer.
%
%%%%%%%%%%%%%%%%%%%%%%%%%%%%%%%%%%%%%%%%%%%%%%%%%%%%%%%%%%%%%%%%%%%%%%
%%%%%%%%%%%%%%%%%%%%%%%%%%%% CONCLUSIONS %%%%%%%%%%%%%%%%%%%%%%%%%%%%%
%%%%%%%%%%%%%%%%%%%%%%%%%%%%%%%%%%%%%%%%%%%%%%%%%%%%%%%%%%%%%%%%%%%%%%
%
\section{Conclusions}
This paper presents a study of human attention dynamics underpinning the driving experience.
Our main contribution is a multi-branch deep network capable of capturing such factors and replicating the driver's focus of attention from raw video sequences. 
The design of our model has been guided by a prior analysis highlighting i) the existence of common gaze patterns across drivers and different scenarios; and ii) a consistent relation between changes in speed, lightning conditions, weather and landscape, and changes in the driver's focus of attention.
Experiments with the proposed architecture and related training strategies yielded state-of-the-art results. To our knowledge, our model is the first able to predict human attention in real-world driving sequences. As the model only input are car-centric videos, it might be integrated with already adopted ADAS technologies. 
% A user study revealed that the attentional patterns predicted by our network made users feel safer than human behavior.
%
%\appendices
\ifCLASSOPTIONcompsoc
  \section*{Acknowledgments}
\else
  \section*{Acknowledgment}
\fi
% This work is partially supported by the \emph{"DriverAttention - Monitoring the car driver's attention with multisensory systems, computer vision and machine learning"} project. 
% We thank Sandro Rubichi for supporting us with psychological insights, and Maximiliano Romani (Ferrari) and Alfredo Reboa (Maserati) for meaningful suggestions and discussions.
We acknowledge the CINECA award under the ISCRA initiative, for the availability of high performance computing resources and support. We also gratefully acknowledge the support of Facebook Artificial Intelligence Research and Panasonic Silicon Valley Lab for the donation of GPUs used for this research.
{\small
\bibliographystyle{ieee}
\bibliography{bibliography}

\begin{thebibliography}{10}\itemsep=-1pt

\bibitem{5206596}
R.~Achanta, S.~Hemami, F.~Estrada, and S.~Susstrunk.
\newblock Frequency-tuned salient region detection.
\newblock In {\em Computer Vision and Pattern Recognition, 2009. CVPR 2009.
  IEEE Conference on}, June 2009.

\bibitem{alletto2016dr}
S.~Alletto, A.~Palazzi, F.~Solera, S.~Calderara, and R.~Cucchiara.
\newblock {Dr(Eye)Ve}: A dataset for attention-based tasks with applications to
  autonomous and assisted driving.
\newblock In {\em Proceedings of the IEEE Conference on Computer Vision and
  Pattern Recognition Workshops}, 2016.

\bibitem{baraldi2016hierarchical}
L.~Baraldi, C.~Grana, and R.~Cucchiara.
\newblock Hierarchical boundary-aware neural encoder for video captioning.
\newblock In {\em Proceedings of the IEEE International Conference on Computer
  Vision}, 2017.

\bibitem{bazzani2017recurrent}
L.~Bazzani, H.~Larochelle, and L.~Torresani.
\newblock Recurrent mixture density network for spatiotemporal visual
  attention.
\newblock In {\em International Conference on Learning Representations (ICLR)},
  2017.

\bibitem{borghi2016poseidon}
G.~Borghi, M.~Venturelli, R.~Vezzani, and R.~Cucchiara.
\newblock Poseidon: Face-from-depth for driver pose estimation.
\newblock In {\em CVPR}, 2017.

\bibitem{borji2016vanishing}
A.~Borji, M.~Feng, and H.~Lu.
\newblock Vanishing point attracts gaze in free-viewing and visual search
  tasks.
\newblock {\em Journal of vision}, 16(14), 2016.

\bibitem{borji2013state}
A.~Borji and L.~Itti.
\newblock State-of-the-art in visual attention modeling.
\newblock {\em IEEE transactions on pattern analysis and machine intelligence},
  35(1), 2013.

\bibitem{CAT2000}
A.~Borji and L.~Itti.
\newblock Cat2000: A large scale fixation dataset for boosting saliency
  research.
\newblock {\em CVPR 2015 workshop on "Future of Datasets"}, 2015.

\bibitem{borji2014look}
A.~Borji, D.~N. Sihite, and L.~Itti.
\newblock What/where to look next? modeling top-down visual attention in
  complex interactive environments.
\newblock {\em IEEE Transactions on Systems, Man, and Cybernetics: Systems},
  44(5), 2014.

\bibitem{bremond2}
R.~Br\'{e}mond, J.-M. Auberlet, V.~Cavallo, L.~D\'{e}sir\'{e}, V.~Faure,
  S.~Lemonnier, R.~Lobjois, and J.-P. Tarel.
\newblock Where we look when we drive: A multidisciplinary approach.
\newblock In {\em Proceedings of Transport Research Arena (TRA'14)}, Paris,
  France, 2014.

\bibitem{mit-saliency-benchmark}
Z.~Bylinskii, T.~Judd, A.~Borji, L.~Itti, F.~Durand, A.~Oliva, and A.~Torralba.
\newblock Mit saliency benchmark, 2015.

\bibitem{bylinskii2016different}
Z.~Bylinskii, T.~Judd, A.~Oliva, A.~Torralba, and F.~Durand.
\newblock What do different evaluation metrics tell us about saliency models?
\newblock {\em arXiv preprint arXiv:1604.03605}, 2016.

\bibitem{3dopNIPS15}
X.~Chen, K.~Kundu, Y.~Zhu, A.~Berneshawi, H.~Ma, S.~Fidler, and R.~Urtasun.
\newblock 3d object proposals for accurate object class detection.
\newblock In {\em NIPS}, 2015.

\bibitem{6871397}
M.~M. Cheng, N.~J. Mitra, X.~Huang, P.~H.~S. Torr, and S.~M. Hu.
\newblock Global contrast based salient region detection.
\newblock {\em IEEE Transactions on Pattern Analysis and Machine Intelligence},
  37(3), March 2015.

\bibitem{mlnet2016}
M.~Cornia, L.~Baraldi, G.~Serra, and R.~Cucchiara.
\newblock {A Deep Multi-Level Network for Saliency Prediction}.
\newblock In {\em International Conference on Pattern Recognition (ICPR)},
  2016.

\bibitem{cornia2016predicting}
M.~Cornia, L.~Baraldi, G.~Serra, and R.~Cucchiara.
\newblock Predicting human eye fixations via an lstm-based saliency attentive
  model.
\newblock {\em arXiv preprint arXiv:1611.09571}, 2016.

\bibitem{Elazary20101338}
L.~Elazary and L.~Itti.
\newblock A bayesian model for efficient visual search and recognition.
\newblock {\em Vision Research}, 50(14), 2010.

\bibitem{fischler1981random}
M.~A. Fischler and R.~C. Bolles.
\newblock Random sample consensus: a paradigm for model fitting with
  applications to image analysis and automated cartography.
\newblock {\em Communications of the ACM}, 24(6), 1981.

\bibitem{fridman2015driver}
L.~Fridman, P.~Langhans, J.~Lee, and B.~Reimer.
\newblock Driver gaze region estimation without use of eye movement.
\newblock {\em IEEE Intelligent Systems}, 31(3), 2016.

\bibitem{frintrop2010computational}
S.~Frintrop, E.~Rome, and H.~I. Christensen.
\newblock Computational visual attention systems and their cognitive
  foundations: A survey.
\newblock {\em ACM Transactions on Applied Perception (TAP)}, 7(1), 2010.

\bibitem{frolich2014will}
B.~Fr\"ohlich, M.~Enzweiler, and U.~Franke.
\newblock Will this car change the lane?-turn signal recognition in the
  frequency domain.
\newblock In {\em 2014 IEEE Intelligent Vehicles Symposium Proceedings}. IEEE,
  2014.

\bibitem{Gao08onthe}
D.~Gao, V.~Mahadevan, and N.~Vasconcelos.
\newblock On the plausibility of the discriminant centersurround hypothesis for
  visual saliency.
\newblock {\em Journal of Vision}, 2008.

\bibitem{gkamas2011guiding}
T.~Gkamas and C.~Nikou.
\newblock Guiding optical flow estimation using superpixels.
\newblock In {\em Digital Signal Processing (DSP), 2011 17th International
  Conference on}. IEEE, 2011.

\bibitem{Goferman:2012:CSD:2360766.2361199}
S.~Goferman, L.~Zelnik-Manor, and A.~Tal.
\newblock Context-aware saliency detection.
\newblock {\em IEEE Trans. Pattern Anal. Mach. Intell.}, 34(10), Oct. 2012.

\bibitem{groner1984looking}
R.~Groner, F.~Walder, and M.~Groner.
\newblock Looking at faces: Local and global aspects of scanpaths.
\newblock {\em Advances in Psychology}, 22, 1984.

\bibitem{Grubm2017}
S.~Grubm{\"u}ller, J.~Plihal, and P.~Nedoma.
\newblock {\em Automated Driving from the View of Technical Standards}.
\newblock Springer International Publishing, Cham, 2017.

\bibitem{henderson2003human}
J.~M. Henderson.
\newblock Human gaze control during real-world scene perception.
\newblock {\em Trends in cognitive sciences}, 7(11), 2003.

\bibitem{dn3}
X.~Huang, C.~Shen, X.~Boix, and Q.~Zhao.
\newblock Salicon: Reducing the semantic gap in saliency prediction by adapting
  deep neural networks.
\newblock In {\em 2015 IEEE International Conference on Computer Vision
  (ICCV)}, Dec 2015.

\bibitem{jain2015car}
A.~Jain, H.~S. Koppula, B.~Raghavan, S.~Soh, and A.~Saxena.
\newblock Car that knows before you do: Anticipating maneuvers via learning
  temporal driving models.
\newblock In {\em Proceedings of the IEEE International Conference on Computer
  Vision}, 2015.

\bibitem{jiang2015salicon}
M.~Jiang, S.~Huang, J.~Duan, and Q.~Zhao.
\newblock Salicon: Saliency in context.
\newblock In {\em The IEEE Conference on Computer Vision and Pattern
  Recognition (CVPR)}, June 2015.

\bibitem{karpathy2014large}
A.~Karpathy, G.~Toderici, S.~Shetty, T.~Leung, R.~Sukthankar, and L.~Fei-Fei.
\newblock Large-scale video classification with convolutional neural networks.
\newblock In {\em Proceedings of the IEEE conference on Computer Vision and
  Pattern Recognition}, 2014.

\bibitem{kingma2014adam}
D.~Kingma and J.~Ba.
\newblock Adam: A method for stochastic optimization.
\newblock In {\em International Conference on Learning Representations (ICLR)},
  2015.

\bibitem{kumar2013learning}
P.~Kumar, M.~Perrollaz, S.~Lefevre, and C.~Laugier.
\newblock Learning-based approach for online lane change intention prediction.
\newblock In {\em Intelligent Vehicles Symposium (IV), 2013 IEEE}. IEEE, 2013.

\bibitem{deepgaze}
M.~K{\"{u}}mmerer, L.~Theis, and M.~Bethge.
\newblock Deep gaze {I:} boosting saliency prediction with feature maps trained
  on imagenet.
\newblock In {\em International Conference on Learning Representations
  Workshops (ICLRW)}, 2015.

\bibitem{kummerer2015information}
M.~K{\"u}mmerer, T.~S. Wallis, and M.~Bethge.
\newblock Information-theoretic model comparison unifies saliency metrics.
\newblock {\em Proceedings of the National Academy of Sciences}, 112(52), 2015.

\bibitem{larson2009contributions}
A.~M. Larson and L.~C. Loschky.
\newblock The contributions of central versus peripheral vision to scene gist
  recognition.
\newblock {\em Journal of Vision}, 9(10), 2009.

\bibitem{dn2}
N.~Liu, J.~Han, D.~Zhang, S.~Wen, and T.~Liu.
\newblock Predicting eye fixations using convolutional neural networks.
\newblock In {\em Computer Vision and Pattern Recognition (CVPR), 2015 IEEE
  Conference on}, June 2015.

\bibitem{lowe1999object}
D.~G. Lowe.
\newblock Object recognition from local scale-invariant features.
\newblock In {\em Computer vision, 1999. The proceedings of the seventh IEEE
  international conference on}, volume~2. Ieee, 1999.

\bibitem{Ma}
Y.-F. Ma and H.-J. Zhang.
\newblock Contrast-based image attention analysis by using fuzzy growing.
\newblock In {\em Proceedings of the Eleventh ACM International Conference on
  Multimedia}, MULTIMEDIA '03, New York, NY, USA, 2003. ACM.

\bibitem{mannan1997fixation}
S.~Mannan, K.~Ruddock, and D.~Wooding.
\newblock Fixation sequences made during visual examination of briefly
  presented 2d images.
\newblock {\em Spatial vision}, 11(2), 1997.

\bibitem{6942210}
S.~Mathe and C.~Sminchisescu.
\newblock Actions in the eye: Dynamic gaze datasets and learnt saliency models
  for visual recognition.
\newblock {\em IEEE Transactions on Pattern Analysis and Machine Intelligence},
  37(7), July 2015.

\bibitem{mauthner2015encoding}
T.~Mauthner, H.~Possegger, G.~Waltner, and H.~Bischof.
\newblock Encoding based saliency detection for videos and images.
\newblock In {\em Proceedings of the IEEE Conference on Computer Vision and
  Pattern Recognition}, 2015.

\bibitem{morris2011lane}
B.~Morris, A.~Doshi, and M.~Trivedi.
\newblock Lane change intent prediction for driver assistance: On-road design
  and evaluation.
\newblock In {\em Intelligent Vehicles Symposium (IV), 2011 IEEE}. IEEE, 2011.

\bibitem{mousavian20163d}
A.~Mousavian, D.~Anguelov, J.~Flynn, and J.~Kosecka.
\newblock 3d bounding box estimation using deep learning and geometry.
\newblock In {\em CVPR}, 2017.

\bibitem{nilsson2016semantic}
D.~Nilsson and C.~Sminchisescu.
\newblock Semantic video segmentation by gated recurrent flow propagation.
\newblock {\em arXiv preprint arXiv:1612.08871}, 2016.

\bibitem{IV}
A.~Palazzi, F.~Solera, S.~Calderara, S.~Alletto, and R.~Cucchiara.
\newblock Where should you attend while driving?
\newblock In {\em IEEE Intelligent Vehicles Symposium Proceedings}, 2017.

\bibitem{pan2016hierarchical}
P.~Pan, Z.~Xu, Y.~Yang, F.~Wu, and Y.~Zhuang.
\newblock Hierarchical recurrent neural encoder for video representation with
  application to captioning.
\newblock In {\em Proceedings of the IEEE Conference on Computer Vision and
  Pattern Recognition}, 2016.

\bibitem{perry2002gaze}
J.~S. Perry and W.~S. Geisler.
\newblock Gaze-contingent real-time simulation of arbitrary visual fields.
\newblock In {\em Human vision and electronic imaging}, volume~57, 2002.

\bibitem{peters2007beyond}
R.~J. Peters and L.~Itti.
\newblock Beyond bottom-up: Incorporating task-dependent influences into a
  computational model of spatial attention.
\newblock In {\em Computer Vision and Pattern Recognition, 2007. CVPR'07. IEEE
  Conference on}. IEEE, 2007.

\bibitem{peters2008applying}
R.~J. Peters and L.~Itti.
\newblock Applying computational tools to predict gaze direction in interactive
  visual environments.
\newblock {\em ACM Transactions on Applied Perception (TAP)}, 5(2), 2008.

\bibitem{posner1985inhibition}
M.~I. Posner, R.~D. Rafal, L.~S. Choate, and J.~Vaughan.
\newblock Inhibition of return: Neural basis and function.
\newblock {\em Cognitive neuropsychology}, 2(3), 1985.

\bibitem{preattention}
N.~Pugeault and R.~Bowden.
\newblock How much of driving is preattentive?
\newblock {\em IEEE Transactions on Vehicular Technology}, 64(12), Dec 2015.

\bibitem{roge2004influence}
J.~Rog{\'e}, T.~P{\'e}bayle, E.~Lambilliotte, F.~Spitzenstetter,
  D.~Giselbrecht, and A.~Muzet.
\newblock Influence of age, speed and duration of monotonous driving task in
  traffic on the driver’s useful visual field.
\newblock {\em Vision research}, 44(23), 2004.

\bibitem{rudoy2013learning}
D.~Rudoy, D.~B. Goldman, E.~Shechtman, and L.~Zelnik-Manor.
\newblock Learning video saliency from human gaze using candidate selection.
\newblock In {\em Proceedings of the IEEE Conference on Computer Vision and
  Pattern Recognition}, 2013.

\bibitem{cogni}
R.~Rukšėnas, J.~Back, P.~Curzon, and A.~Blandford.
\newblock Formal modelling of salience and cognitive load.
\newblock {\em Electronic Notes in Theoretical Computer Science}, 208, 2008.

\bibitem{sardegna2002encyclopedia}
J.~Sardegna, S.~Shelly, and S.~Steidl.
\newblock {\em The encyclopedia of blindness and vision impairment}.
\newblock Infobase Publishing, 2002.

\bibitem{6287326}
B.~Schölkopf, J.~Platt, and T.~Hofmann.
\newblock {\em Graph-Based Visual Saliency}.
\newblock MIT Press, 2007.

\bibitem{bremond1}
L.~Simon, J.~P. Tarel, and R.~Bremond.
\newblock Alerting the drivers about road signs with poor visual saliency.
\newblock In {\em Intelligent Vehicles Symposium, 2009 IEEE}, June 2009.

\bibitem{MatheSminchisescuPAMI2015}
C.~S. Stefan~Mathe.
\newblock Actions in the eye: Dynamic gaze datasets and learnt saliency models
  for visual recognition.
\newblock {\em IEEE Transactions on Pattern Analysis and Machine Intelligence},
  37, 2015.

\bibitem{tatler2007central}
B.~W. Tatler.
\newblock The central fixation bias in scene viewing: Selecting an optimal
  viewing position independently of motor biases and image feature
  distributions.
\newblock {\em Journal of vision}, 7(14), 2007.

\bibitem{Tatler}
B.~W. Tatler, M.~M. Hayhoe, M.~F. Land, and D.~H. Ballard.
\newblock {Eye guidance in natural vision: Reinterpreting salience}.
\newblock {\em Journal of Vision}, 11(5), May 2011.

\bibitem{head2}
A.~Tawari and M.~M. Trivedi.
\newblock Robust and continuous estimation of driver gaze zone by dynamic
  analysis of multiple face videos.
\newblock In {\em Intelligent Vehicles Symposium Proceedings, 2014 IEEE}, June
  2014.

\bibitem{theeuwes2010top}
J.~Theeuwes.
\newblock Top--down and bottom--up control of visual selection.
\newblock {\em Acta psychologica}, 135(2), 2010.

\bibitem{torralba2006contextual}
A.~Torralba, A.~Oliva, M.~S. Castelhano, and J.~M. Henderson.
\newblock Contextual guidance of eye movements and attention in real-world
  scenes: the role of global features in object search.
\newblock {\em Psychological review}, 113(4), 2006.

\bibitem{tran2014learning}
D.~Tran, L.~Bourdev, R.~Fergus, L.~Torresani, and M.~Paluri.
\newblock Learning spatiotemporal features with 3d convolutional networks.
\newblock In {\em 2015 IEEE International Conference on Computer Vision
  (ICCV)}, Dec 2015.

\bibitem{treisman1980feature}
A.~M. Treisman and G.~Gelade.
\newblock A feature-integration theory of attention.
\newblock {\em Cognitive psychology}, 12(1), 1980.

\bibitem{ueda2017eye}
Y.~Ueda, Y.~Kamakura, and J.~Saiki.
\newblock Eye movements converge on vanishing points during visual search.
\newblock {\em Japanese Psychological Research}, 59(2), 2017.

\bibitem{TJunct}
G.~Underwood, K.~Humphrey, and E.~van Loon.
\newblock Decisions about objects in real-world scenes are influenced by visual
  saliency before and during their inspection.
\newblock {\em Vision Research}, 51(18), 2011.

\bibitem{head3}
F.~Vicente, Z.~Huang, X.~Xiong, F.~D. la~Torre, W.~Zhang, and D.~Levi.
\newblock Driver gaze tracking and eyes off the road detection system.
\newblock {\em IEEE Transactions on Intelligent Transportation Systems}, 16(4),
  Aug 2015.

\bibitem{wang2017understanding}
P.~Wang, P.~Chen, Y.~Yuan, D.~Liu, Z.~Huang, X.~Hou, and G.~Cottrell.
\newblock Understanding convolution for semantic segmentation.
\newblock {\em arXiv preprint arXiv:1702.08502}, 2017.

\bibitem{wang2017central}
P.~Wang and G.~W. Cottrell.
\newblock Central and peripheral vision for scene recognition: A
  neurocomputational modeling explorationwang \& cottrell.
\newblock {\em Journal of Vision}, 17(4), 2017.

\bibitem{wang2015saliency}
W.~Wang, J.~Shen, and F.~Porikli.
\newblock Saliency-aware geodesic video object segmentation.
\newblock In {\em Proceedings of the IEEE Conference on Computer Vision and
  Pattern Recognition}, 2015.

\bibitem{wang2015consistent}
W.~Wang, J.~Shen, and L.~Shao.
\newblock Consistent video saliency using local gradient flow optimization and
  global refinement.
\newblock {\em IEEE Transactions on Image Processing}, 24(11), 2015.

\bibitem{wolfe1998visual}
J.~M. Wolfe.
\newblock Visual search.
\newblock {\em Attention}, 1, 1998.

\bibitem{Wolfe89guidedsearch}
J.~M. Wolfe, K.~R. Cave, and S.~L. Franzel.
\newblock Guided search: an alternative to the feature integration model for
  visual search.
\newblock {\em Journal of Experimental Psychology: Human Perception \&
  Performance}, 1989.

\bibitem{yu2015multi}
F.~Yu and V.~Koltun.
\newblock Multi-scale context aggregation by dilated convolutions.
\newblock In {\em ICLR}, 2016.

\bibitem{Zhai}
Y.~Zhai and M.~Shah.
\newblock Visual attention detection in video sequences using spatiotemporal
  cues.
\newblock In {\em Proceedings of the 14th ACM International Conference on
  Multimedia}, MM '06, New York, NY, USA, 2006. ACM.

\bibitem{Zhai:2006}
Y.~Zhai and M.~Shah.
\newblock Visual attention detection in video sequences using spatiotemporal
  cues.
\newblock In {\em Proceedings of the 14th ACM International Conference on
  Multimedia}, MM '06, New York, NY, USA, 2006. ACM.

\bibitem{zhong2013video}
S.-h. Zhong, Y.~Liu, F.~Ren, J.~Zhang, and T.~Ren.
\newblock Video saliency detection via dynamic consistent spatio-temporal
  attention modelling.
\newblock In {\em AAAI}, 2013.

\end{thebibliography}
}
\begin{IEEEbiography}[{\includegraphics[width=1in,height=1.25in,clip,keepaspectratio]{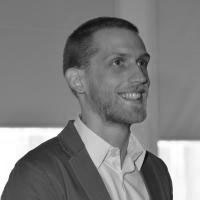}}]{Andrea Palazzi} received the master's degree
in computer engineering from the University of
Modena and Reggio Emilia in 2015. He is currently
PhD candidate within the ImageLab
group in Modena, researching on computer vision and deep learning for automotive applications.
\end{IEEEbiography}

\begin{IEEEbiography}[{\includegraphics[width=1in,height=1.25in,clip,keepaspectratio]{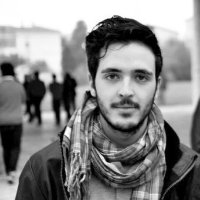}}]{Davide Abati}
received the master's degree
in computer engineering from the University of
Modena and Reggio Emilia in 2015. He is currently
working toward the PhD degree within the ImageLab
group in Modena, researching on computer vision and deep learning for image and video understanding.
\end{IEEEbiography}

\begin{IEEEbiography}[{\includegraphics[width=1in,height=1.25in,clip,keepaspectratio]{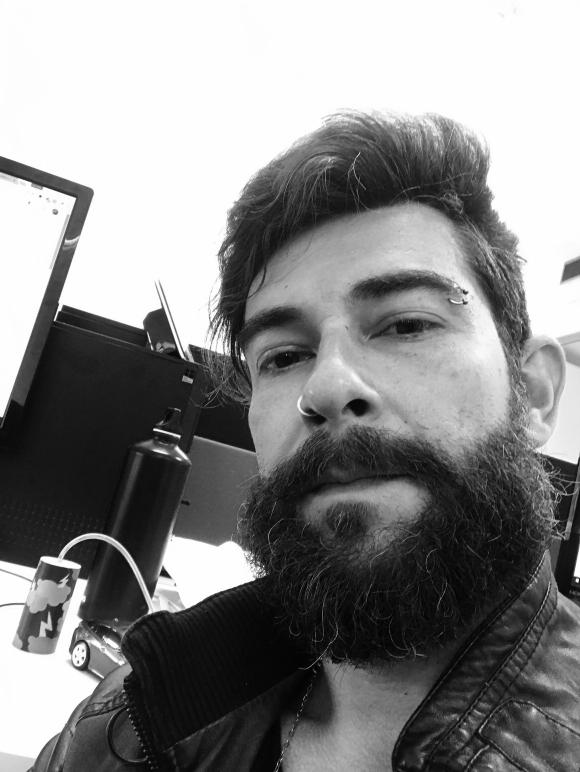}}]{Simone Calderara}
received a computer engineering
master's degree in 2005 and the PhD
degree in 2009 from the University of Modena
and Reggio Emilia, where he is currently an
assistant professor within the Imagelab group.
His current research interests include computer
vision and machine learning applied to human
behavior analysis, visual tracking in crowded scenarios,
and time series analysis for forensic applications.
He is a member of the IEEE.
\end{IEEEbiography}

\begin{IEEEbiography}[{\includegraphics[width=1in,height=1.25in,clip,keepaspectratio]{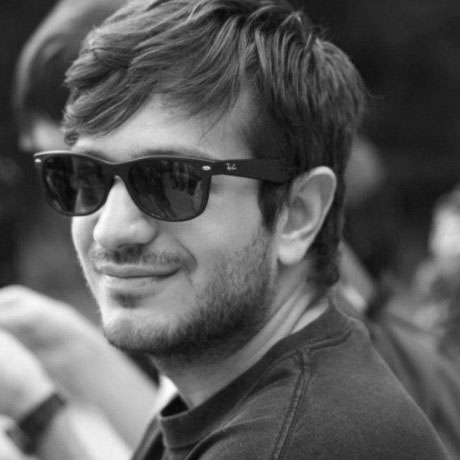}}]{Francesco Solera}
obtained a master's degree
in computer engineering from the University of
Modena and Reggio Emilia in 2013 and a PhD degree in 2017. His research
mainly addresses applied machine
learning and social computer vision.
\end{IEEEbiography}

\begin{IEEEbiography}[{\includegraphics[width=1in,height=1.25in,clip,keepaspectratio]{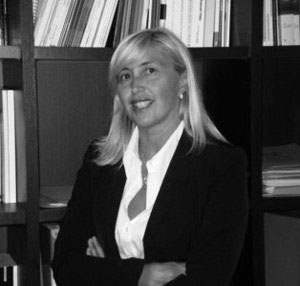}}]{Rita Cucchiara}
received the master's degree in
Electronic Engineering and the PhD degree in
Computer Engineering from the University of Bologna,
Italy, in 1989 and 1992, respectively. Since
2005, she is a full professor at the University of
Modena and Reggio Emilia, Italy, where she
heads the ImageLab group and is Director of the SOFTECH-ICT
research center. She is currently President of the Italian Association of Pattern Recognition, (GIRPR), affilaited with IAPR. She published more than  300 papers on pattern recognition computer vision and multimedia, and in particular in human analysis, HBU and egocentric-vision. The research carried out spans on different application fields, such as video-surveillance, automotive and multimedia big data annotation. Corrently she is AE of IEEE Transactions on Multimedia and serves in the Governing Board of IAPR and in the Advisory Board of the CVF. \end{IEEEbiography}

%%% SUPPLEMENTARY MATERIAL
\clearpage
\twocolumn[
   \begin{center}
      \huge\textbf{Supplementary Material}\\
   \end{center}
]
\noindent Here we provide additional material useful to the understanding
of the paper. Additional multimedia are available at: \url{https://ndrplz.
github.io/dreyeve/}.
\section{DR(eye)VE dataset design}
The following table reports the design the \texttt{DR(eye)VE} dataset. The dataset is composed of 74 sequences of 5 minutes each, recorded under a variety of driving conditions. Experimental design played a crucial role in preparing the dataset to rule out spurious correlation between driver, weather, traffic, daytime and scenario. Here we report the details for each sequence.
\section{Visual Assessment Details}
\label{sup:foveation}
The aim of this section is to provide additional details on the implementation of visual assessment presented in Sec.~\ref{sec:qualitative_assessment} of the paper. Please note that additional videos regarding this section can be found together with other supplementary multimedia at \url{https://ndrplz.github.io/dreyeve/}. Eventually, the reader is referred to \url{https://github.com/ndrplz/dreyeve} for the code used to create foveated videos for visual assessment.\\
\subsection*{Space Variant Imaging System}
Space Variant Imaging System (SVIS) is a MATLAB toolbox that allows to foveate images in real-time\cite{perry2002gaze}, which has been used in a large number of scientific works to approximate human foveal vision since its introduction in 2002. In this frame, the term \textit{foveated imaging} refers to the creation and display of static or video imagery where the resolution varies across the image. In analogy to human foveal vision, the highest resolution region is called the foveation region. In a video, the location of the foveation region can obviously change dynamically. It is also possible to have more than one foveation region in each image.\\

\noindent The foveation process is implemented in the SVIS toolbox as follows: first the the input image is repeatedly low-passed filtered and down-sampled to half of the current resolution by a \textit{Foveation Encoder}. In this way a low-pass pyramid of images is obtained. Then a foveation pyramid is created selecting regions from different resolutions proportionally to the distance from the foveation point. Concretely, the foveation region will be at the highest resolution; first ring around the foveation region will be taken from half-resolution image; and so on. Eventually, a \textit{Foveation Decoder} up-sample, interpolate and blend each layer in the foveation pyramid to create the output foveated image.\\

\noindent The software is open-source and publicly available here: \url{http://svi.cps.utexas.edu/software.shtml}. The  interested reader is referred to the SVIS website for further details.
\begin{table}[t]
    \caption{\drive~ train set: details for each sequence.}
    \begin{tabular}{  c | l  l  l  l  l }
    \textbf{Sequence} & \textbf{Daytime} & \textbf{Weather} & \textbf{Landscape} & \textbf{Driver} & \textbf{Set}\\ \hline
	01 & Evening & Sunny & Countryside & D8 & Train Set \\ 
	02 & Morning & Cloudy & Highway & D2 & Train Set \\ 
	03 & Evening & Sunny & Highway & D3 & Train Set \\ 
	04 & Night & Sunny & Downtown & D2 & Train Set \\ 
	05 & Morning & Cloudy & Countryside & D7 & Train Set \\ 
	06 & Morning & Sunny & Downtown & D7 & Train Set \\ 
	07 & Evening & Rainy & Downtown & D3 & Train Set \\ 
	08 & Evening & Sunny & Countryside & D1 & Train Set \\ 
	09 & Night & Sunny & Highway & D1 & Train Set \\ 
	10 & Evening & Rainy & Downtown & D2 & Train Set \\ 
	11 & Evening & Cloudy & Downtown & D5 & Train Set \\ 
	12 & Evening & Rainy & Downtown & D1 & Train Set \\ 
	13 & Night & Rainy & Downtown & D4 & Train Set \\ 
	14 & Morning & Rainy & Highway & D6 & Train Set \\ 
	15 & Evening & Sunny & Countryside & D5 & Train Set \\ 
	16 & Night & Cloudy & Downtown & D7 & Train Set \\ 
	17 & Evening & Rainy & Countryside & D4 & Train Set \\ 
	18 & Night & Sunny & Downtown & D1 & Train Set \\ 
	19 & Night & Sunny & Downtown & D6 & Train Set \\ 
	20 & Evening & Sunny & Countryside & D2 & Train Set \\ 
	21 & Night & Cloudy & Countryside & D3 & Train Set \\ 
	22 & Morning & Rainy & Countryside & D7 & Train Set \\ 
	23 & Morning & Sunny & Countryside & D5 & Train Set \\ 
	24 & Night & Rainy & Countryside & D6 & Train Set \\ 
	25 & Morning & Sunny & Highway & D4 & Train Set \\ 
	26 & Morning & Rainy & Downtown & D5 & Train Set \\ 
	27 & Evening & Rainy & Downtown & D6 & Train Set \\ 
	28 & Night & Cloudy & Highway & D5 & Train Set \\ 
	29 & Night & Cloudy & Countryside & D8 & Train Set \\ 
	30 & Evening & Cloudy & Highway & D7 & Train Set \\ 
	31 & Morning & Rainy & Highway & D8 & Train Set \\ 
	32 & Morning & Rainy & Highway & D1 & Train Set \\ 
	33 & Evening & Cloudy & Highway & D4 & Train Set \\ 
	34 & Morning & Sunny & Countryside & D3 & Train Set \\ 
	35 & Morning & Cloudy & Downtown & D3 & Train Set \\ 
	36 & Evening & Cloudy & Countryside & D1 & Train Set \\ 
	37 & Morning & Rainy & Highway & D8 & Train Set
	\end{tabular}
\end{table}
\begin{table}[t]
    \caption{\drive~ test set: details for each sequence.}
    \begin{tabular}{  c | l  l  l  l  l }
    \textbf{Sequence} & \textbf{Daytime} & \textbf{Weather} & \textbf{Landscape} & \textbf{Driver} & \textbf{Set}\\ \hline
	38 & Night & Sunny & Downtown & D8 & Test Set \\ 
	39 & Night & Rainy & Downtown & D4 & Test Set \\ 
	40 & Morning & Sunny & Downtown & D1 & Test Set \\ 
	41 & Night & Sunny & Highway & D1 & Test Set \\ 
	42 & Evening & Cloudy & Highway & D1 & Test Set \\ 
	43 & Night & Cloudy & Countryside & D2 & Test Set \\ 
	44 & Morning & Rainy & Countryside & D1 & Test Set \\ 
	45 & Evening & Sunny & Countryside & D4 & Test Set \\ 
	46 & Evening & Rainy & Countryside & D5 & Test Set \\ 
	47 & Morning & Rainy & Downtown & D7 & Test Set \\ 
	48 & Morning & Cloudy & Countryside & D8 & Test Set \\ 
	49 & Morning & Cloudy & Highway & D3 & Test Set \\ 
	50 & Morning & Rainy & Highway & D2 & Test Set \\ 
	51 & Night & Sunny & Downtown & D3 & Test Set \\ 
	52 & Evening & Sunny & Highway & D7 & Test Set \\ 
	53 & Evening & Cloudy & Downtown & D7 & Test Set \\ 
	54 & Night & Cloudy & Highway & D8 & Test Set \\ 
	55 & Morning & Sunny & Countryside & D6 & Test Set \\ 
	56 & Night & Rainy & Countryside & D6 & Test Set \\ 
	57 & Evening & Sunny & Highway & D5 & Test Set \\ 
	58 & Night & Cloudy & Downtown & D4 & Test Set \\ 
	59 & Morning & Cloudy & Highway & D7 & Test Set \\ 
	60 & Morning & Cloudy & Downtown & D5 & Test Set \\ 
	61 & Night & Sunny & Downtown & D5 & Test Set \\ 
	62 & Night & Cloudy & Countryside & D6 & Test Set \\ 
	63 & Morning & Rainy & Countryside & D8 & Test Set \\ 
	64 & Evening & Cloudy & Downtown & D8 & Test Set \\ 
	65 & Morning & Sunny & Downtown & D2 & Test Set \\ 
	66 & Evening & Sunny & Highway & D6 & Test Set \\ 
	67 & Evening & Cloudy & Countryside & D3 & Test Set \\ 
	68 & Morning & Cloudy & Countryside & D4 & Test Set \\ 
	69 & Evening & Rainy & Highway & D2 & Test Set \\ 
	70 & Morning & Rainy & Downtown & D3 & Test Set \\ 
	71 & Night & Cloudy & Highway & D6 & Test Set \\ 
	72 & Evening & Cloudy & Downtown & D2 & Test Set \\ 
	73 & Night & Sunny & Countryside & D7 & Test Set \\ 
	74 & Morning & Rainy & Downtown & D4 & Test Set
    \end{tabular}
\end{table}

\subsection*{Videoclip Foveation}

\begin{figure*}[htp]
\centering
\begin{tabular}{ccc}
\includegraphics[width=0.31\textwidth]{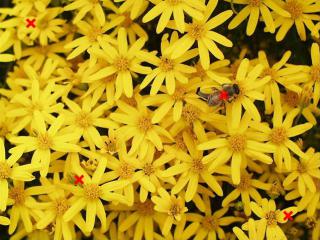} & 
\includegraphics[width=0.31\textwidth]{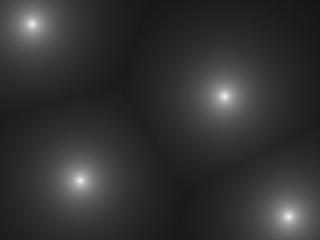} & 
\includegraphics[width=0.31\textwidth]{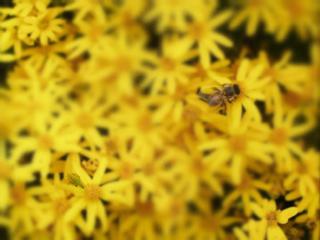}\\
(a) & (b) & (c)
\end{tabular}
\caption{Foveation process using SVIS software is depicted here. Starting from one or more fixation points in a given frame (a), a smooth resolution map is built (b). Image locations with higher values in the resolution map will undergo less blur in the output image (c).}
\label{fig:sup:foveation_bee}
\end{figure*}

\noindent \textbf{From fixation maps back to fixations.} The SVIS toolbox allows to foveate images starting from a list of $(x, y)$ coordinates which represent the foveation points in the given image (please see Fig.~\ref{fig:sup:foveation_bee} for details). However, we do not have this information as in our work we deal with continuous attentional maps rather than discrete points of fixations. To be able to use the same software API we need to regress from the attentional map (either true or predicted) a list of approximated yet plausible fixation locations. To this aim we simply extract the 25 points with highest value in the attentional map. This is justified by the fact that in the phase of dataset creation the ground truth \emph{fixation map} $F_t$ for a frame at time $t$ is built by accumulating projected gaze points in a temporal sliding window of $k=25$ frames, centered in $t$ (see Sec.~\ref{sec:dataset} of the paper). The output of this phase is thus a fixation map we can use as input for the SVIS toolbox.\\

\noindent \textbf{Taking the blurred-deblurred ratio into account.} To the visual assessment purposes, keeping track the amount of blur that a videoclip has undergone is also relevant. Indeed, a certain video may give rise to higher perceived safety only because a more delicate blur allows the subject to see a clearer picture of the driving scene. In order to consider this phenomenon we do the following.\\
Given an input image $I \in \mathbb{R}^{h, w, c}$ the output of the Foveation Encoder is a resolution map $R_{map} \in \mathbb{R}^{h, w, 1}$, taking value in range $[0, 255]$, as depicted in Fig.~\ref{fig:sup:foveation_bee}~(b). Each value indicates the resolution that a certain pixel will have in the foveated image after decoding, where 0 and 255 indicates minimum and maximum resolution respectively.\\
For each video $\mathbf{v}$, we measure video average resolution after foveation as follows:
$$
\mathbf{v}_{res} = \frac{1}{N} \sum_{f=1}^{N}\sum_{i}R_{map}(i, f)
$$
where N is the number of frames in the video ($1000$ in our setting) and $R_{map}(i, f)$ denotes the $i^{th}$ pixel of the resolution map corresponding to the $f^{th}$ frame of the input video. The higher the value of ${v}_{res}$ the more information is preserved in the foveation process. Due to the sparser location of fixations in ground truth attentional maps, these result in much less blurred videoclips. Indeed videos foveated with model predicted attentional maps have in average only the 38\% of the resolution w.r.t. videos foveated starting from ground truth attentional maps. Despite this bias, model predicted foveated videos still gave rise to higher perceived safety to assessment participants.
\section{Perceived safety assessment}
\label{sec:perceived_safety}
\begin{figure*}[bt]
     \centering
     \resizebox{\textwidth}{!}{\begin{tabular}{ccc}
     \includegraphics[height=3.1cm]{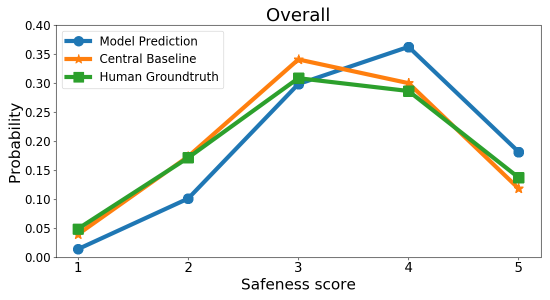} &
     \includegraphics[height=3.1cm]{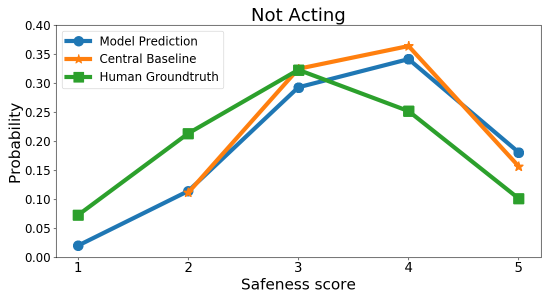} &
     \includegraphics[height=3.1cm]{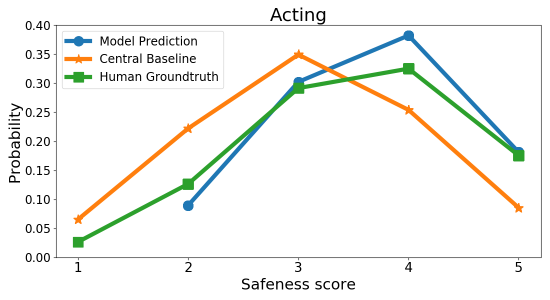}\\
     (a) &
     (b) &
     (c)
     \end{tabular}}
     \caption{Distributions of safeness scores for different map sources, namely Model Prediction, Center Baseline and Human Groundtruth. Considering the score distribution over all foveated videoclips (a) the three distributions may look similar, even though the model prediction still scores slightly better. However, when considering only the foveated videos contaning acting subsequences (b) the model prediction significantly outperforms both center baseline and human groundtruth. Conversely, when the videoclips did not contain acting subsequences (\textit{i.e.} the car was mostly going straight) the fixation map from human driver is the one perceived as less safe, while both model prediction and center baseline perform similarly.}
     \label{fig:visual_assessment_score_distributions}
\end{figure*}

The assessment of predicted fixation maps described in Sec~\ref{sec:qualitative_assessment} has also been carried out for validating the model in terms of perceived safety. Indeed, partecipants were also asked to answer the following question:
\begin{itemize}
    \item If you were sitting in the same car of the driver whose attention behavior you just observed, how safe would you feel? (rate from 1 to 5)
\end{itemize}
The aim of the question is to measure the comfort level of the observer during a driving experience when suggested to focus at specific locations in the scene. 
The underlying assumption is that the observer is more likely to feel safe if he agrees that the suggested focus is lighting up the right portion of the scene, that is what he thinks it is worth looking in the current driving scene. Conversely, if the observer wishes to focus at some specific location but he cannot retrieve details there, he is going to feel uncomfortable.\\
The answers provided by subjects, summarized in Fig.~\ref{fig:visual_assessment_score_distributions}, indicate that perceived safety for videoclips foveated using the attentional maps predicted by the model is generally higher than for the ones foveated using either human or central baseline maps. Nonetheless the central bias baseline proves to be extremely competitive, in particular in non-acting videoclips in which it scores similarly to the model prediction. It is worth noticing that in this latter case both kind of automatic predictions outperform human ground truth by a significant margin (Fig.~\ref{fig:visual_assessment_score_distributions}b). Conversely, when we consider only the foveated videoclips containing acting subsequences, the human ground truth is perceived as much safer than the baseline, despite still scores worse than our model prediction (Fig.~\ref{fig:visual_assessment_score_distributions}c). These results hold despite due to the localization of the fixations the average resolution of the predicted maps is only the 38\% of the resolution of ground truth maps (\textit{i.e.} videos foveated using prediction map feature much less information). We did not measure significant difference in perceived safety across the different drivers in the dataset ($\sigma^2=0.09$).
\begin{figure}[ht]
     \centering
     \includegraphics[width=0.8\columnwidth]{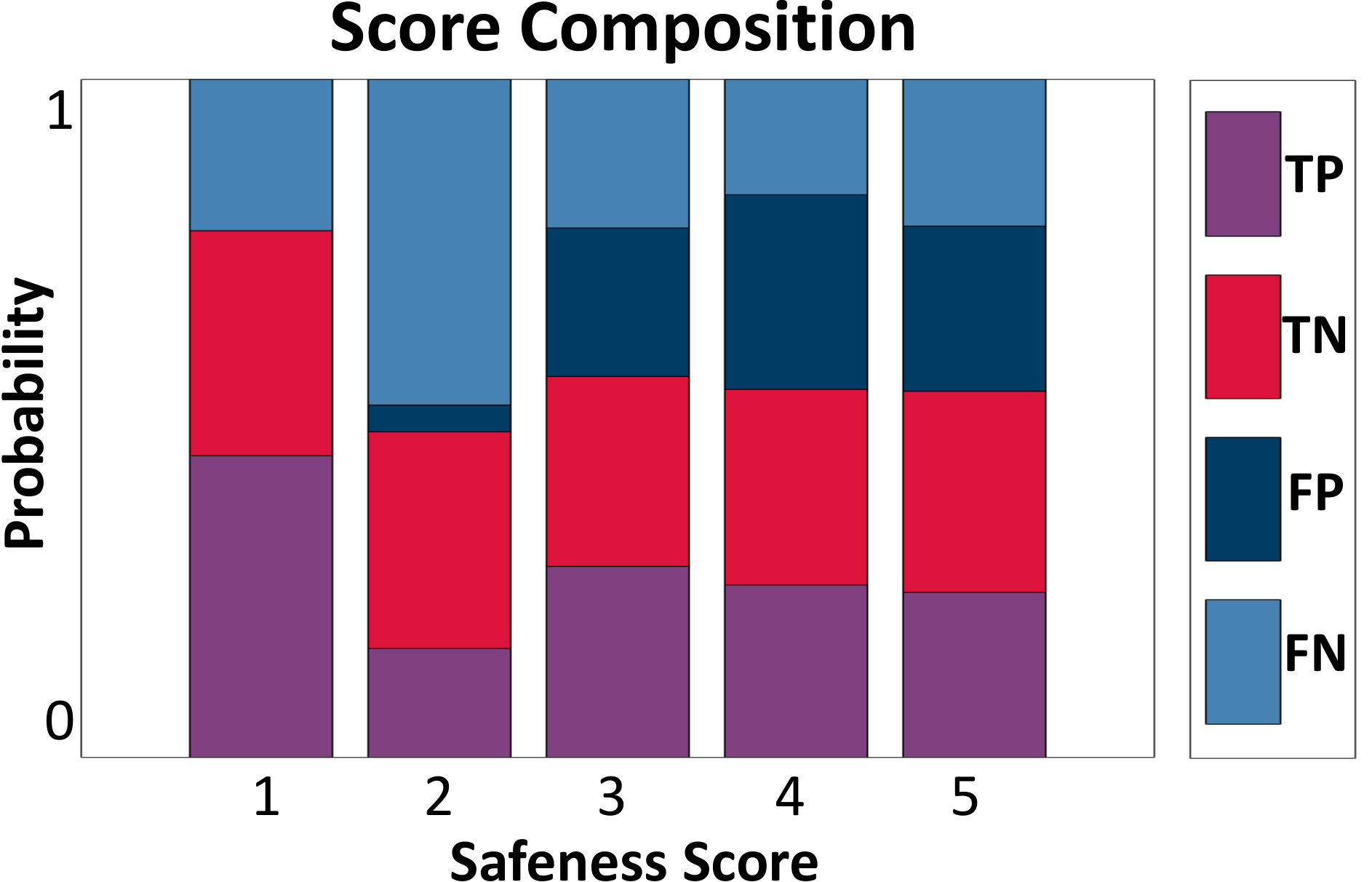}
     \caption{The stacked bar graph represents the ratio of TP, TN, FP and FN composing each score. The increasing score of FP -- participants falsely thought the attentional map came from a human driver -- highlights that participants were tricked into believing that "safer" clips came from humans.
     }
     \label{fig:visual_assessment_confusion_composition}
\end{figure}
We report in Fig~\ref{fig:visual_assessment_confusion_composition} the composition of each score in terms of answers to the other visual assessment question (\quotes{Would you say the observed attention behavior comes from a human driver? (yes/no)}). 
This analysis aims to measure participants' bias towards human driving ability.
Indeed, increasing trend of false positives towards higher scores suggests that participants were tricked into believing that \quotes{safer} clips came from humans. The reader is referred to Fig.~\ref{fig:visual_assessment_confusion_composition} for further details.
\begin{figure}[t]
    \centering
    \includegraphics[width=0.5\textwidth]{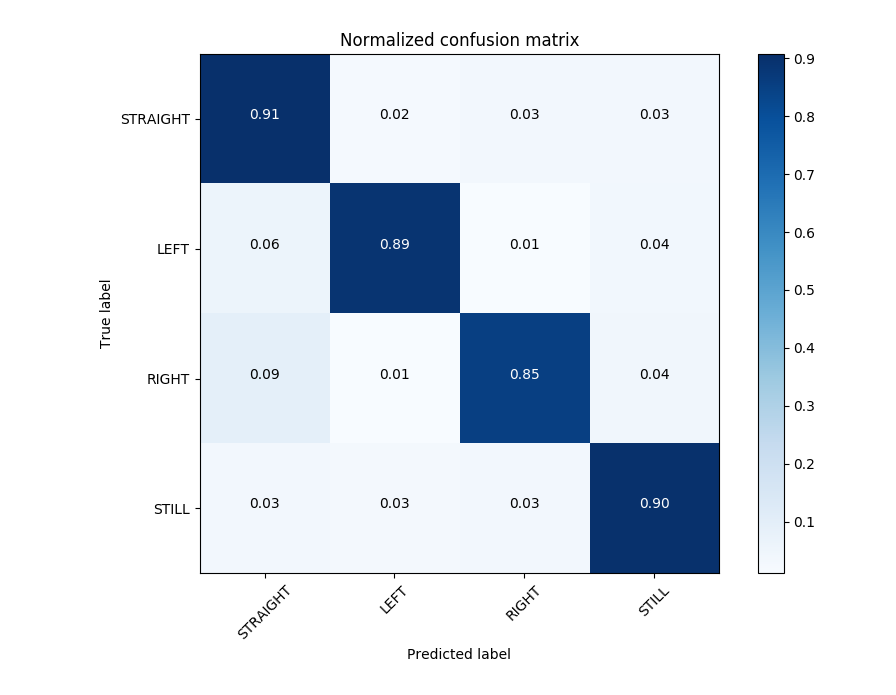}    \caption{Confusion matrix for SVM classifier trained to distinguish driving actions from network activations. The accuracy is generally high, which corroborates the assumption that the model benefits from learning an internal representation of the different driving sub-tasks.}
    \label{fig:svm_confusion_matrix}
\end{figure}
\section{Do subtasks help in FoA prediction?}
The driving task is inherently composed of many subtasks, such as turning or merging in traffic, looking for parking and so on. While such fine-grained subtasks are hard to discover (and probably to emerge during learning) due to scarcity, here we show how the proposed model has been able to leverage on more common subtask to get to the final prediction. These subtasks are: turning left/right, going straight, being still. We gathered automatic annotation through GPS information released with the dataset. We then train a linear SVM classifier to distinguish the above 4 different actions starting from the activations of the last layer of \texttt{multi-path} model, unrolled in a feature vector. The SVM classifier scores a 90\% of accuracy on the test set (5000 uniformely sampled videoclips), supporting the fact that network activations are highly discriminative for distinguishing the different driving subtasks. Please refer to Fig.~\ref{fig:svm_confusion_matrix} for further details. Code to replicate this result is available at \url{https://github.com/ndrplz/dreyeve} along with the code of all other experiments in the paper.
\section{Segmentation}
In this section we report exemplar cases that particularly benefit from the segmentation branch. In Fig.~\ref{fig:sup:segmentation_helps} we can appreciate that, among the three branches, only the semantic one captures the real gaze, that is focused on traffic lights and street signs.
\bgroup
\setlength{\tabcolsep}{.16667em}
\begin{figure*}[htp]
\centering
\begin{tabular}{ccccc}
\textbf{RGB} & \textbf{flow} & \textbf{segmentation} & \textbf{gt}\\
\includegraphics[width=0.24\textwidth]{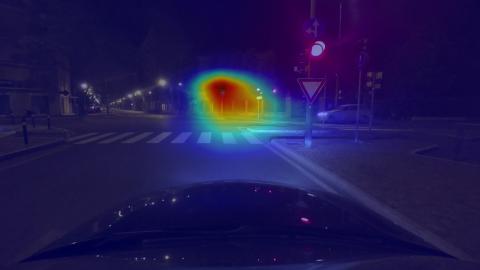}& 
\includegraphics[width=0.24\textwidth]{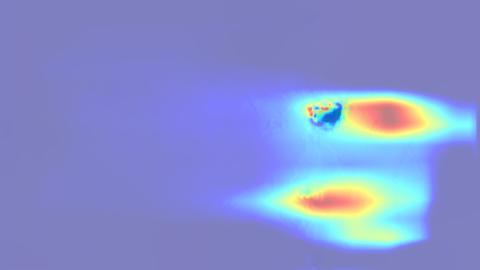}& 
\includegraphics[width=0.24\textwidth]{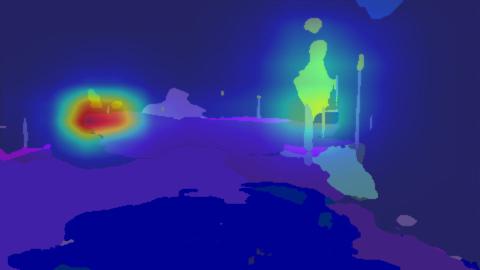}& 
\includegraphics[width=0.24\textwidth]{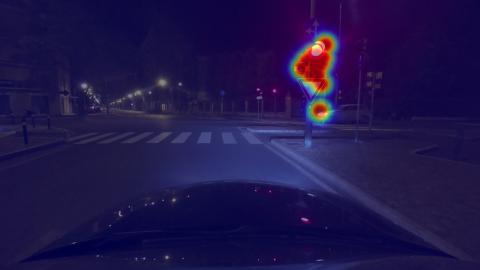}\\
\includegraphics[width=0.24\textwidth]{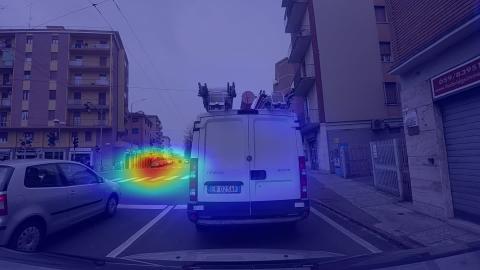}& 
\includegraphics[width=0.24\textwidth]{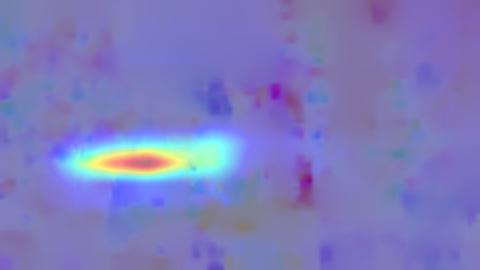}& 
\includegraphics[width=0.24\textwidth]{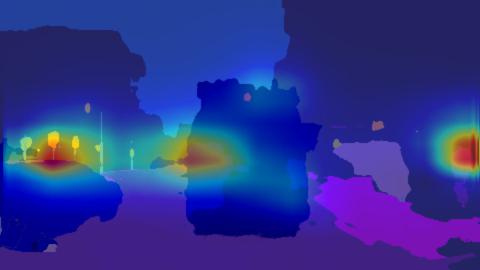}& 
\includegraphics[width=0.24\textwidth]{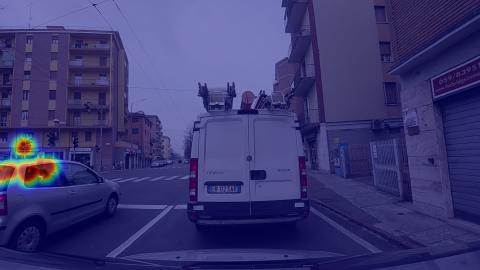}
\end{tabular}
\caption{Some examples of the beneficial effect of the semantic segmentation branch. In the two cases depicted here, the car is stopped at a crossroad. While the RGB branch remains biased towards the road vanishing point and the optical flow branch focuses on moving objects, the semantic branch tends to highlight traffic lights and signals, coherently with the human behavior.}
\label{fig:sup:segmentation_helps}
\end{figure*}
\section{Ablation Study}
\bgroup
\setlength{\tabcolsep}{.16667em}
\begin{figure*}[htp]
\centering
\begin{tabular}{cccccc}
\textbf{Input frame} & \textbf{GT} & \textbf{\texttt{multi-branch}} & RGB & FLOW & SEG \\
\includegraphics[width=0.16\textwidth]{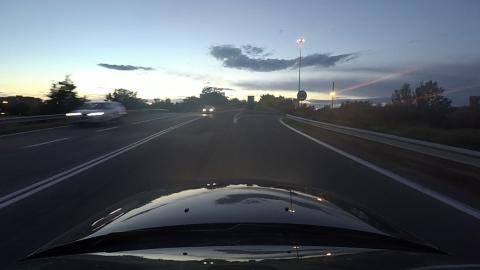}& \includegraphics[width=0.16\textwidth]{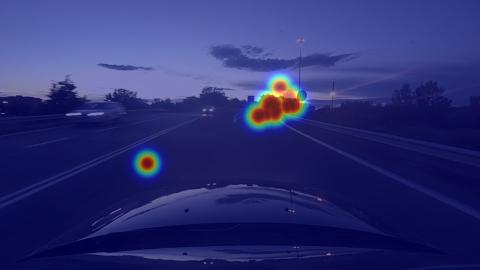}& \includegraphics[width=0.16\textwidth]{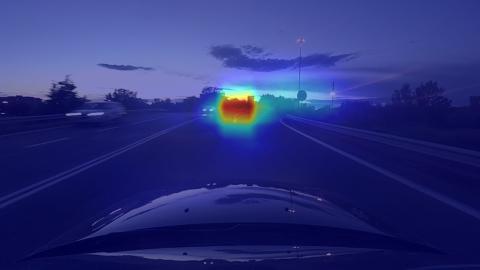}& \includegraphics[width=0.16\textwidth]{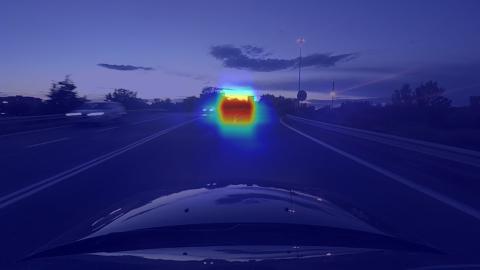}& \includegraphics[width=0.16\textwidth]{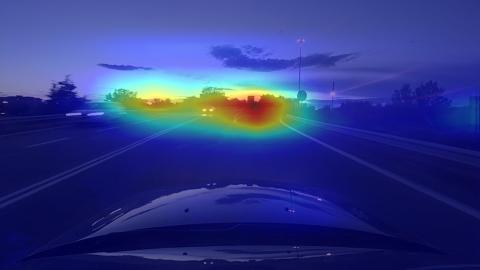}&
\includegraphics[width=0.16\textwidth]{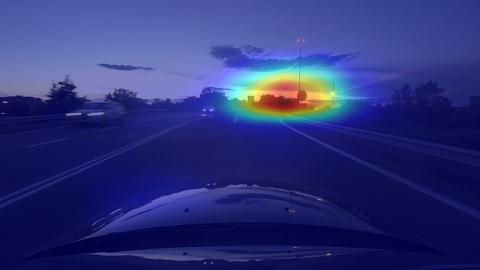}\\
\includegraphics[width=0.16\textwidth]{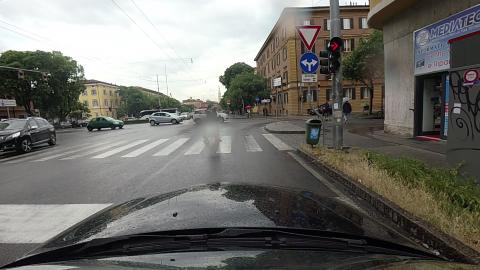}& \includegraphics[width=0.16\textwidth]{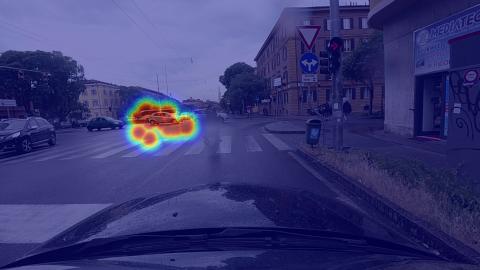}& \includegraphics[width=0.16\textwidth]{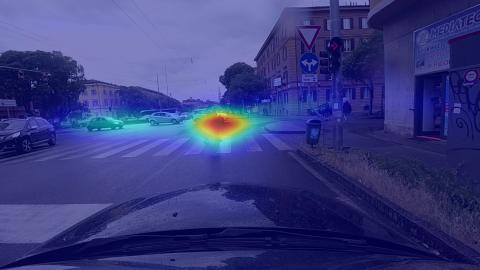}& \includegraphics[width=0.16\textwidth]{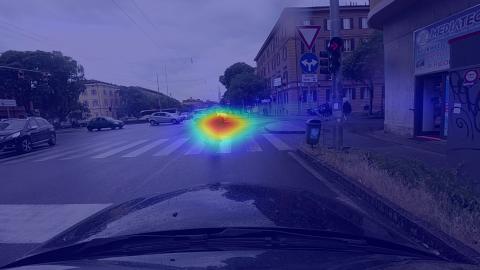}& \includegraphics[width=0.16\textwidth]{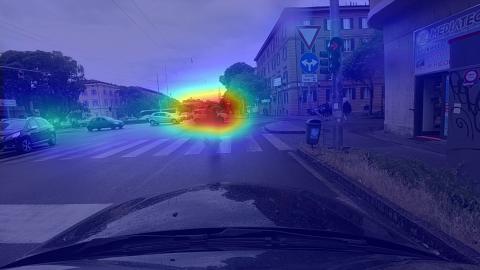}&
\includegraphics[width=0.16\textwidth]{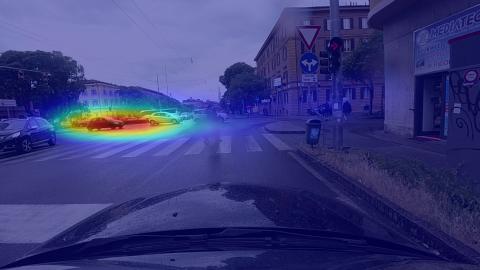}\\
\includegraphics[width=0.16\textwidth]{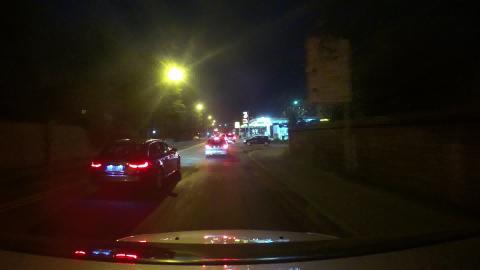}& \includegraphics[width=0.16\textwidth]{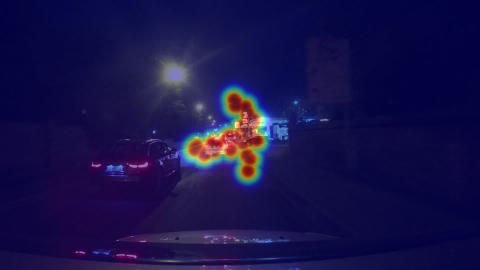}& \includegraphics[width=0.16\textwidth]{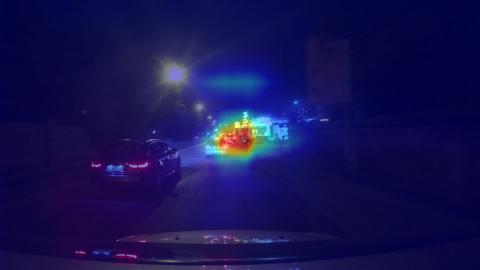}& \includegraphics[width=0.16\textwidth]{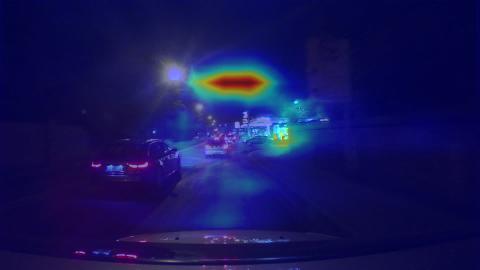}& \includegraphics[width=0.16\textwidth]{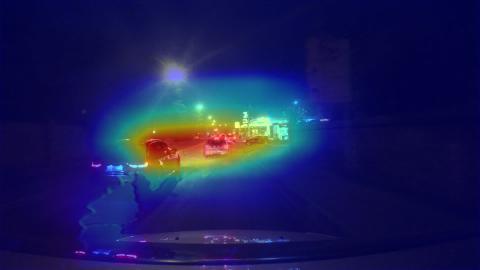}&
\includegraphics[width=0.16\textwidth]{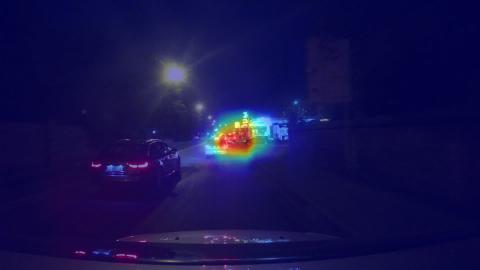}\\
\includegraphics[width=0.16\textwidth]{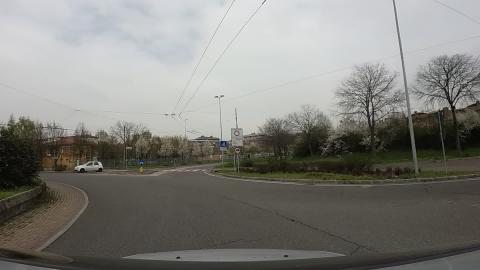}& \includegraphics[width=0.16\textwidth]{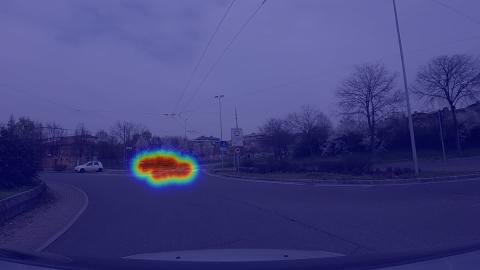}& \includegraphics[width=0.16\textwidth]{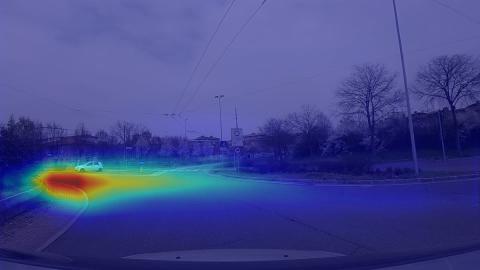}& \includegraphics[width=0.16\textwidth]{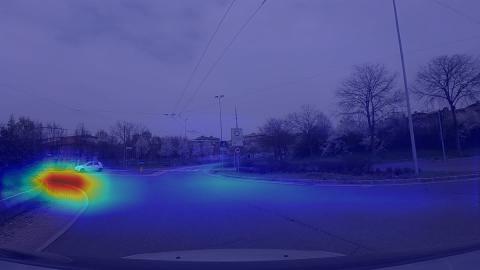}& \includegraphics[width=0.16\textwidth]{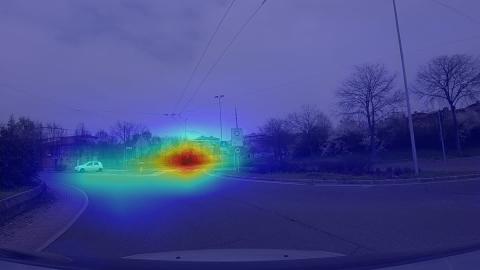}&
\includegraphics[width=0.16\textwidth]{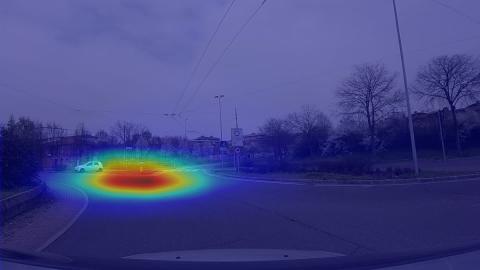}\\
\end{tabular}
\caption{Example cases that qualitatively show how each branch contribute to the final prediction. Best viewed on screen.}
\label{fig:sup:branch_ablation}
\end{figure*}
\egroup
\noindent In Fig.~\ref{fig:sup:branch_ablation} we showcase several examples depicting the contribution of each branch of the \texttt{multi-branch} model in predicting the visual focus of attention of the driver. As expected, the RGB branch is the one that more heavily influences the overall network output.
\section{Error analysis for non-planar homographic projection}
\label{sup:bound}
\begin{figure*}[t]
\centering
\begin{tabular}{ccc}
\includegraphics[width=0.4\textwidth]{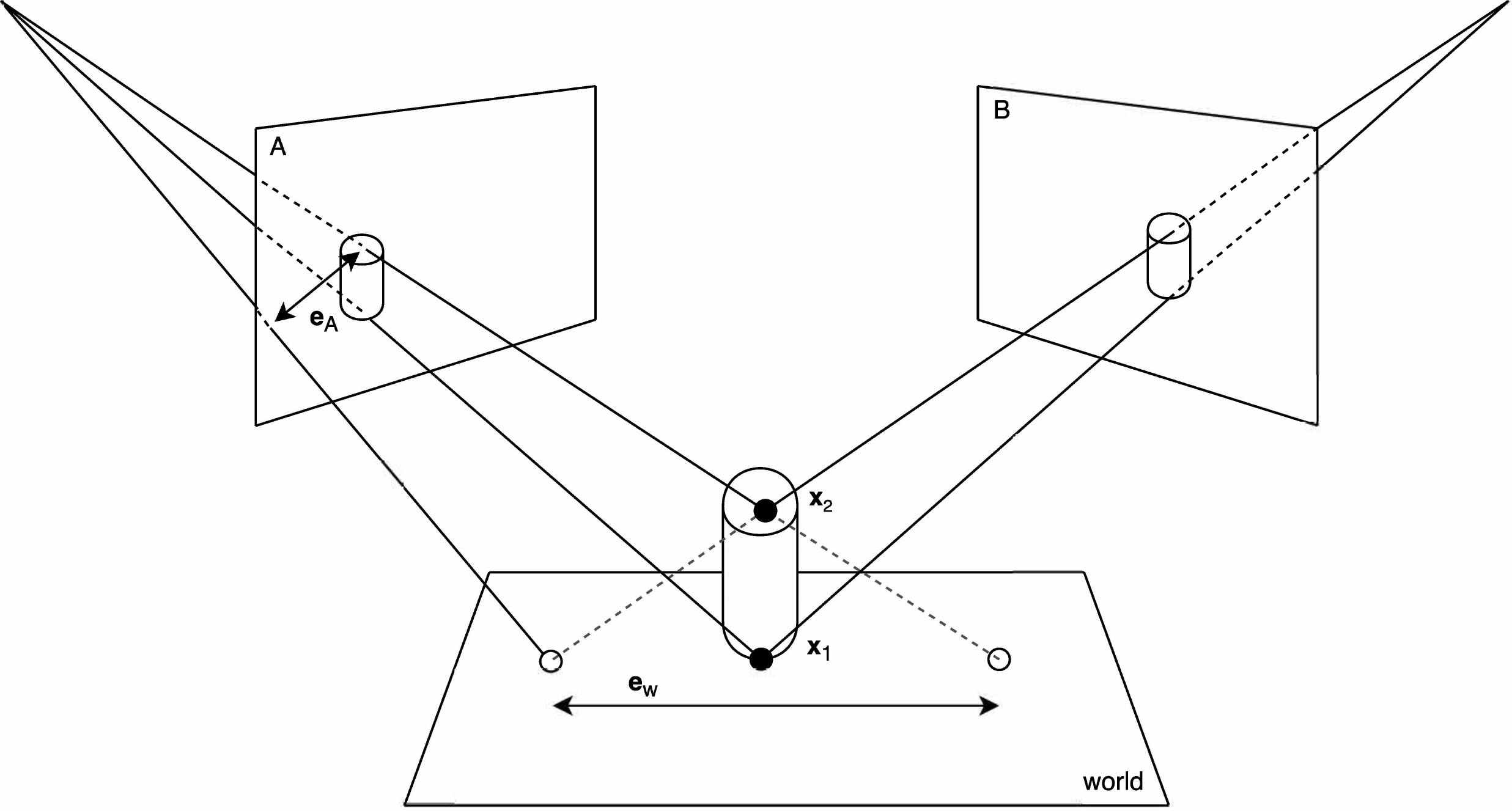}
&&
\includegraphics[width=0.35\textwidth]{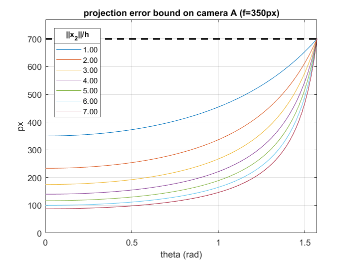}\\
(a)&&(b)
\end{tabular}
\caption{(a) Two image planes capture a 3D scene from different viewpoints and (b) a use case of the bound derived below.}
\label{fig:geometry_3d}
\end{figure*}

A homography $H$ is a projective transformation from a plane $A$ to another plane $B$ such that the collinearity property is preserved during the mapping. In real world applications, the homography matrix $H$ is often computed through an overdetermined set of image coordinates lying on the same implicit plane, aligning points on the plane in one image with points on the plane in the other image. If the input set of points is approximately lying on the true implicit plane, then $H$ can be efficiently recovered through least square projection minimization.

Once the transformation $H$ has been either defined or approximated from data, to map an image point $\bf x$ from the first image to the respective point $H\bf x$ in the second image, the basic assumption is that $\bf x$ actually lies on the implicit plane.
In practice this assumption is widely violated in real world applications, when the process of mapping is automated and the content of the mapping is not known a-priori.

\subsection{The geometry of the problem}
In Fig.~\ref{fig:geometry_3d} we show the generic setting of two cameras capturing the same 3D plane. To construct an erroneous case study, we put a cylinder on top of the plane. Points on the implicit 3D world plane can be consistently mapped across views with an homography transformation and retain their original semantic. As an example, the point $\bf x_1$ is the center of the cylinder base both in world coordinates and across different views. Conversely, the point $\bf x_2$ on the top of the cylinder cannot be consistently mapped from one view to the other. To see why, suppose we want to map $\bf x_2$ from view $B$ to view $A$. Since the homography assumes $\bf x_2$ to also be on the implicit plane, its inferred 3D position is far from the true top of the cylinder and is depicted with the leftmost empty circle in Fig.~\ref{fig:geometry_3d}. When this point gets reprojected to view $A$, its image coordinates are unaligned with the correct position of the cylinder top in that image. We call this offset the \emph{reprojection error} on plane $A$, or $\bf e_A$. Analogously, a reprojection error on plane $B$ could be computed with an homographic projection of point $\bf x_2$ from view $A$ to view $B$.\\

The reprojection error is useful to measure the perceptual misalignment of projected points with their intended locations, but due to the (re)projections involved is not an easy tool to work with. Moreover, the very same point can produce different reprojection errors when measured on $A$ and on $B$. A related error also arising in this setting is the \emph{metric error} $\bf e_W$, or the displacement in world space of the projected image points at the intersection with the implicit plane. This measure of error is of particular interest because it is view-independent, does not depend on the rotation of the cameras with respect to the plane and is zero if and only if the reprojection error is also zero.

\subsection{Computing the metric error}
Since the metric error does not depend on the mutual rotation of the plane with the camera views, we can simplify Fig.~\ref{fig:geometry_3d} by retaining only the optical centers $A$ and $B$ from all cameras and by setting, without loss of generality, the reference system on the projection of the 3D point on the plane. This second step is useful to factor out the rotation of the world plane, which is unknown in the general setting. The only assumption we make is that the non-planar point $\bf x_2$ can be seen from both camera views. This simplification is depicted in Fig.~\ref{fig:geometry_3d_simple_a}(a), where we have also named several important quantities such as the distance $h$ of ${\bf x}_2$ from the plane.\\
\begin{figure*}[t]
\centering
\begin{tabular}{ccccc}
\includegraphics[width=0.25\textwidth]{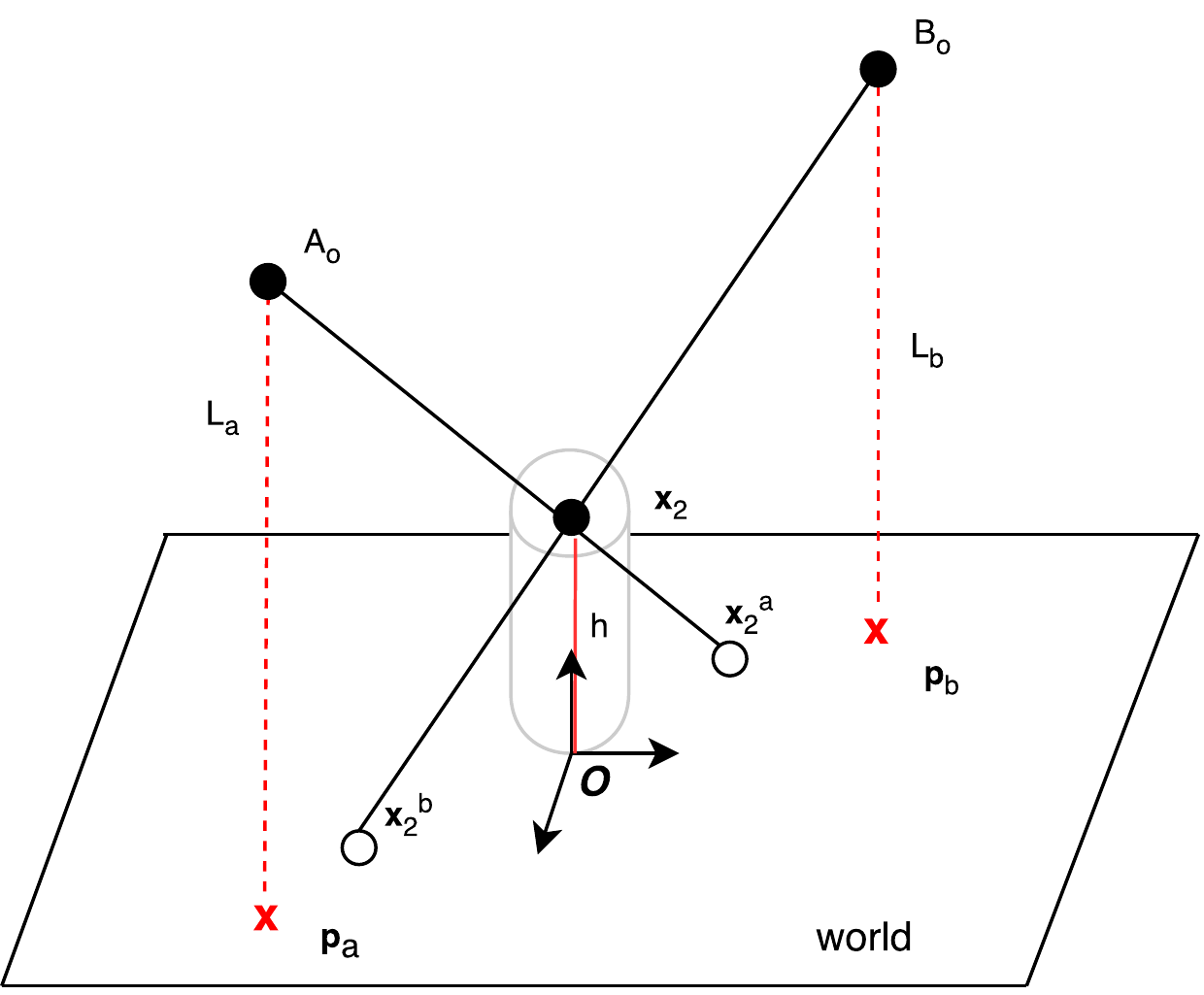} &\quad\quad\quad&
\includegraphics[width=0.25\textwidth]{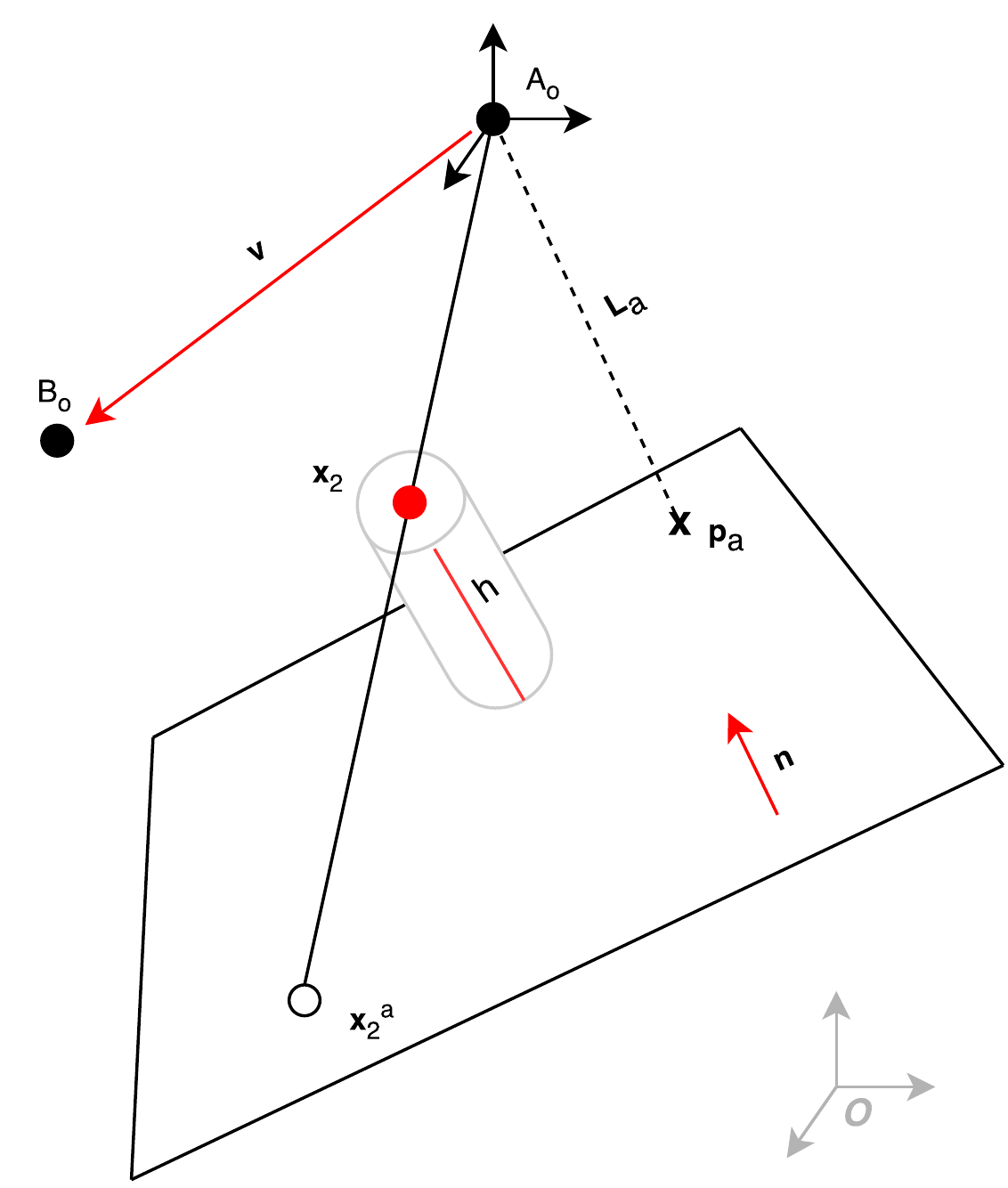}&\quad\quad\quad&\includegraphics[width=0.25\textwidth]{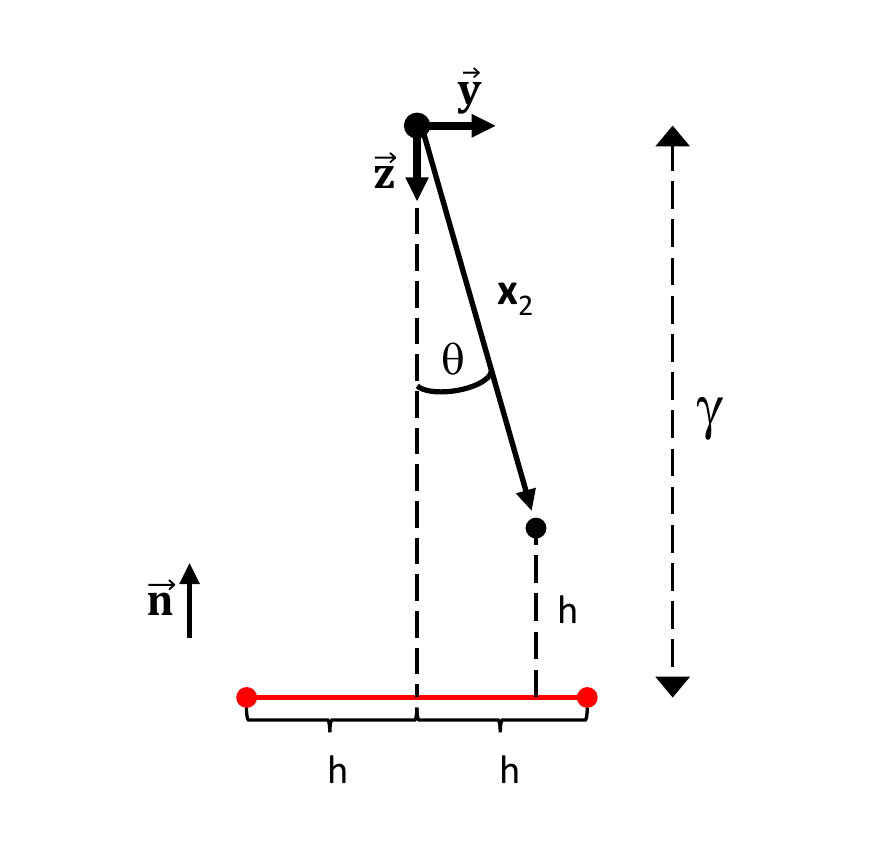}\\
(a) & &(b) && (c)
\end{tabular}
\caption{(a) By aligning the reference system with the plane normal, centered on the projection of the non-planar point onto the plane, the metric error is the magnitude of the difference between the two vectors $\vec{{\bf x}_2^a}$ and $\vec{{\bf x}_2^b}$. The red lines help to highlight the similarity of inner and outer triangles having ${\bf x}_x^a$ as a vertex. (b) The geometry of the problem when the reference system is placed off the plane in an arbitrary position (gray) or, specifically, on one of the camera (black). (c) The simplified setting in which we consider the projection of the metric error $\| {\bf e}_w \|$ on the camera plane of $A$.}
\label{fig:geometry_3d_simple_a}
\end{figure*}

In Fig.~\ref{fig:geometry_3d_simple_a}(a), the metric error can be computed as the magnitude of the difference between the two vectors relating points ${\bf x}_2^a$ and ${\bf x}_2^b$ to the origin:
\begin{equation}
{\bf e}_w = {\bf x}_2^a - {\bf x}_2^b.
\end{equation}
The aforementioned points are at the intersection of the lines connecting the optical center of the cameras with the 3D point ${\bf x}_2$ and the implicit plane. An easy way to get such points is through their magnitude and orientation. As an example, consider the point ${\bf x}_2^a$. Starting from ${\bf x}_2^a$ the following two similar triangles can be built:
\begin{equation}
\Delta {\bf x}_2^a {\bf p}_a A \sim \Delta {\bf x}_2^a {\bf O}{\bf x}_2.
\end{equation}

\noindent Since they are similar, \emph{i.e.} they share the same shape, we can measure the distance of ${\bf x}_2^a$ from the origin. More formally,
\begin{equation}
\frac{L_a}{\|{\bf p}_a\|+\|{\bf x}_2^a\|} = \frac{h}{\|{\bf x}_2^a\|},
\end{equation}
from which we can recover
\begin{equation}
\|{\bf x}_2^a\| = \frac{h\|{\bf p}_a\|}{L_a-h}.
\end{equation}
The orientation of the ${\bf x}_2^a$ vector can be obtained directly from the orientation of the ${\bf p}_a$ vector, which is known and equal to
\begin{equation}
\vv{{\bf x}_2^a} = - \vv{{\bf p}_a} = - \frac{{\bf p}_a}{\|{\bf p}_a\|}.
\end{equation}
Eventually, with the magnitude and orientation in place, we can locate the vector pointing to ${\bf x}_2^a$:
\begin{equation}
{\bf x}_2^a = \|{\bf x}_2^a\| \vv{{\bf x}_2^a} = -\frac{h}{L_a-h}{\bf p}_a.
\end{equation}
Similarly, ${\bf x}_2^b$ can also be computed. The metric error can thus be described by the following relation:
\begin{equation}
\label{eq:error}
{\bf e}_w = h\left(\frac{{\bf p}_b}{L_b-h} - \frac{{\bf p}_a}{L_a-h}\right).
\end{equation}
The error ${\bf e}_w$ is a vector, but a convenient scalar can be obtained by using the preferred norm.

\subsection{Computing the error on a camera reference system}

When the plane inducing the homography remains unknown, the bound and the error estimation from the previous section cannot be directly applied. A more general case is obtained if the reference system is set off the plane, and in particular, on one of the cameras. The new geometry of the problem is shown in Fig.~\ref{fig:geometry_3d_simple_a}(b), where the reference system is placed on camera $A$. In this setting, the metric error is a function of four independent quantities (highlighted in red in the figure): i) the point ${\bf x}_2$, ii) the distance of such point from the inducing plane $h$, iii) the plane normal $\vv{\bf n}$ and iv) the distance between the cameras ${\bf v}$, which is also equal to the position of camera $B$.\\

\noindent To this end, starting from Eq. \eqref{eq:error}, we are interested in expressing ${\bf p}_b$, ${\bf p}_a$, $L_b$ and $L_a$ in terms of this new reference system. Since ${\bf p}_a$ is the projection of $A$ on the plane it can also be defined as
\begin{equation}
\label{eq:pa}
{\bf p}_a = A - (A - {\bf p}_k)\vv{\bf n}\otimes \vv{\bf n} = A + \boldsymbol\alpha \vv{\bf n}\otimes \vv{\bf n},
\end{equation}
where $\vv{\bf n}$ is the plane normal, ${\bf p}_k$ is an arbitrary point on the plane that we set to ${\bf x}_2-h\otimes \vv{\bf n}$, \emph{i.e.} the projection of ${\bf x}_2$ on the plane. To ease the readability of the following equations, $\boldsymbol\alpha = -(A-{\bf x}_2-h\otimes \vv{\bf n})$. Now, if ${\bf v}$ describes the distance from  $A$ to $B$, we have
\begin{equation}
\label{eq:pb}
\begin{aligned}
    {\bf p}_b &= A + {\bf v} - (A + {\bf v} - {\bf p}_k)\vv{\bf n}\otimes \vv{\bf n}\\
    &=A + \boldsymbol\alpha \vv{\bf n}\otimes \vv{\bf n} + {\bf v} - {\bf v}\vv{\bf n}\otimes \vv{\bf n}.
\end{aligned}
\end{equation}
Through similar reasoning, $L_a$ and $L_b$ are also rewritten as follows:
\begin{equation}
\label{eq:la_b}
    \begin{aligned}
    L_a &= A-{\bf p}_a = -\boldsymbol\alpha \vv{\bf n}\otimes \vv{\bf n}\\
    L_b &= B - {\bf p}_b = -\boldsymbol\alpha \vv{\bf n}\otimes \vv{\bf n} + {\bf v} \vv{\bf n}\otimes \vv{\bf n}.
    \end{aligned}
\end{equation}
Eventually, by substituting Eq. \eqref{eq:pa}-\eqref{eq:la_b} in Eq. \eqref{eq:error} and by fixing the origin on the location of camera $A$ so that $A = (0, 0, 0)^T$, we have:
\begin{equation}
\label{eq:final_error}
{\bf e}_w = \frac{h\otimes {\bf v}}{({\bf x}_2-{\bf v})\vv{\bf n}}\left(\vv{\bf n}\otimes \vv{\bf n}(1 - \frac{1}{1-\frac{h}{h-{\bf x}_2\vv{\bf n}}}) - I\right).
\end{equation}
Notably, the vector ${\bf v}$ and the scalar $h$ both appear as multiplicative factors in Eq. \eqref{eq:final_error}, so that if any of them goes to zero, then the magnitude of the metric error ${\bf e}_w$ also goes to zero.\\
If we assume that $h\neq 0$, we can go one step further and obtain a formulation were ${\bf x}_2$ and ${\bf v}$ are always divided by $h$, suggesting that what really matters is not the absolute position of ${\bf x}_2$ or camera $B$ with respect to camera $A$ but rather how many times further ${\bf x}_2$ and camera $B$ are from $A$ than ${\bf x}_2$ from the plane. Such relation is made explicit below:
\begin{align}
\label{eq:final_error_x2_h}
{\bf e}_w = &h\underbrace{\frac{\|{\bf v}\|/h}{(\frac{\|{\bf x}_2\|}{h}\otimes\vv{\bf x}_2-\frac{\|{\bf v}\|}{h}\otimes 
\vv{\bf v})\vv{\bf n}}}_{Q}\otimes \\
& \vv{\bf v}\Bigg(\vv{\bf n}\otimes \vv{\bf n}\underbrace{(1 - \frac{1}{1-\frac{1}{1-\frac{\|{\bf x}_2\|}{h}\otimes\vv{\bf x}_2\vv{\bf n}}})}_Z - I\Bigg).
\end{align}

\subsection{Working towards a bound.} Let $M=\|{\bf x}_2\|/\|{\bf v}\||\cos\theta|$, being $\theta$ the angle between $\vv{\bf x}_2$ and $\vv{\bf n}$, and let $\beta$ be the angle between $\vv{\bf v}$ and $\vv{\bf n}$. Then $Q$ can be rewritten as $Q=(M-\cos\beta)^{-1}$. Note that under the assumption that $M\geq2$, $Q\leq1$ always holds. Indeed for $M\geq2$ to hold, we need to require $\|{\bf x}_2\|\geq2\|{\bf v}\|/|\cos\theta|$. Next, consider the scalar $Z$: it is easy to verify that if $|\|{\bf x}_2\|/h\otimes\vv{\bf x}_2\vv{\bf n}| > 1$, then $|Z| \leq 1$. Since both $\vv{\bf x}_2$ and $\vv{\bf n}$ are versors, the magnitude of their dot product is at most one. It follows that $|Z|<1$ if and only if $\|{\bf x}_2\| > h$. Now we are left with a versor $\vv{\bf v}$ that multiplies the difference of two matrices. If we compute such product we obtain a new vector with magnitude less or equal to one, $\vv{\bf v}\vv{\bf n}\otimes \vv{\bf n}$, and the versor $\vv{\bf v}$. The difference of such vectors is at most 2. Summing up all the presented considerations, we have that the magnitude of the error is bounded as follows.

\begin{observation}
\label{obs:bound}
If $\|{\bf x}_2\|\geq2\|{\bf v}\|/|\cos\theta|$ and $\|{\bf x}_2\| > h$, then
$\|{\bf e}_w\| \leq 2h$.
\end{observation}

We now aim to derive a projection error bound from the above presented metric error bound. In order to do so, we need to introduce the focal length of the camera $f$. For simplicity, we'll assume that $f=f_x=f_y$.
First, we simplify our setting without loosing the upper bound constraint. To do so, we consider the worst case scenario, in which the mutual position of the plane and the camera maximizes the projected error:
\begin{itemize}
\item the plane rotation is so that $\vv{\bf n}\parallelsum z$;
\item the error segment is just in front of the camera;
\item the plane rotation along the $z$ axis is such that the parallel component of the error w.r.t. the $x$ axis is zero (this allows us to express the segment end points with simple coordinates without loosing generality);
\item the camera $A$ falls on the middle point of the error segment.
\end{itemize}
In the simplified scenario depicted in Fig.~\ref{fig:geometry_3d_simple_a}(c), the projection of the error is maximized.
%This way, our setting simplifies to the one depicted in Fig.~\ref{fig:geometric_simple}, that, despite being simple, represents the scenario under which the error projection is maximized.
In this case, the two points we want to project are ${\bf p}_1 = [0, -h, \gamma]$ and ${\bf p}_2 = [0, h, \gamma]$ (we consider the case in which $||{\bf e}_w|| = 2h$, see Observation.~\ref{obs:bound}) where $\gamma$ is the distance of the camera from the plane.
Considering the focal length $f$ of camera $A$, ${\bf p}_1$ and ${\bf p}_2$ are projected as follows:
\begin{align}
% primo punto
K_A{\bf p}_1 = \begin{bmatrix}
    f & 0 & 0 & 0\\
    0 & f & 0 & 0\\
    0 & 0 & 1 & 0
    \end{bmatrix} 
    \begin{bmatrix}
    0\\
    -h\\
    \gamma
    \end{bmatrix}
    = \begin{bmatrix}
    0 \\
    -fh\\
    \gamma
    \end{bmatrix}
    \rightarrow
    \begin{bmatrix}
    0 \\
    -\frac{fh}{\gamma}
    \end{bmatrix}
    \\
    K_A{\bf p}_2 = \begin{bmatrix}
    f & 0 & 0 & 0\\
    0 & f & 0 & 0\\
    0 & 0 & 1 & 0
    \end{bmatrix} 
    \begin{bmatrix}
    0\\
    h\\
    \gamma
    \end{bmatrix}
    = \begin{bmatrix}
    0 \\
    fh\\
    \gamma
    \end{bmatrix}
    \rightarrow
    \begin{bmatrix}
    0 \\
    \frac{fh}{\gamma}
    \end{bmatrix}
\end{align}
Thus, the magnitude of the projection $\|{\bf e}_a\|$ of the metric error $\|{\bf e}_w\|$ is bounded by $\frac{2fh}{\gamma}$.\\
Now, we notice that $\gamma = h + {\bf x}_2\vv{\bf n } = h + \|{\bf x}_2\|cos(\theta)$, so
\begin{equation}
\label{eq:bound_px}
\|{\bf e}_a\| \leq \frac{2fh}{\gamma} = \frac{2fh}{h + \|{\bf x}_2\|cos(\theta)} = \frac{2f}{1+\frac{\|{\bf x}_2\|}{h}cos(\theta)}
\end{equation}
Notably, the right term of the equation is maximized when $\cos(\theta) = 0$ (since when $cos(\theta) < 0$ the point is behind the camera, which is impossible in our setting). Thus, we obtain that $\|{\bf e}_a\| \leq 2f$.\\

\noindent Fig.~\ref{fig:geometry_3d}(b) shows a use case of the bound in Eq.~\ref{eq:bound_px}. It shows values of $\theta$ up to $pi/2$, where the presented bound simplifies to $\|{\bf e}_a\| \leq 2f$ (dashed black line). 
In practice, if we require i) $\theta \leq \pi/3$ and ii) that the camera-object distance $\|{\bf x}_2\|$ is at least three times the plane-object distance $h$, and if we let $f=350$px, then the error is always lower than 200px, which translate to a precision up to 20\% of an image at 1080p resolution.

\section{The effect of random cropping}
\label{sup:shifts}
In order to advocate for the peculiar training strategy illustrated in Sec.~\ref{sec:single_foa_branch}, involving two streams processing both resized clips and randomly cropped clips, we perform an additional experiment as follows.
We first re-train our \texttt{multi-branch} architecture following the same procedures explained in the main paper, except for the cropping of input clips and groundtruth maps, which is always central rather than random.
% Subsequently, we run additional tests on \emph{acting} subsequences in order to validate the intuition that a model trained with random crops is more resilient to central bias. 
At test time we shift each input clip in the range $[-800, 800]$ pixels (negative and positive shifts indicate left and right translations respectively). After the translation, we apply mirroring to fill borders on the opposite side of the shift direction. We perform the same operation on groundtruth maps and report the mean $D_{KL}$ of the \texttt{multi-branch} model when trained with random and central crops, as a function of the translation size, in Fig.~\ref{fig:shifts}.
The figure highlights how random cropping consistently outperforms central cropping. Importantly, the gap in performance increases with amount of shift applied, from a $27.86\%$ relative $D_{KL}$ difference when no translation is performed to $258.13\%$ and $216.17\%$ increases for $-800$ px and $800$ px translations, suggesting the model trained with central crops is not robust to positional shifts.
\begin{figure}[t]
    \centering
    \includegraphics[width=0.8\columnwidth]{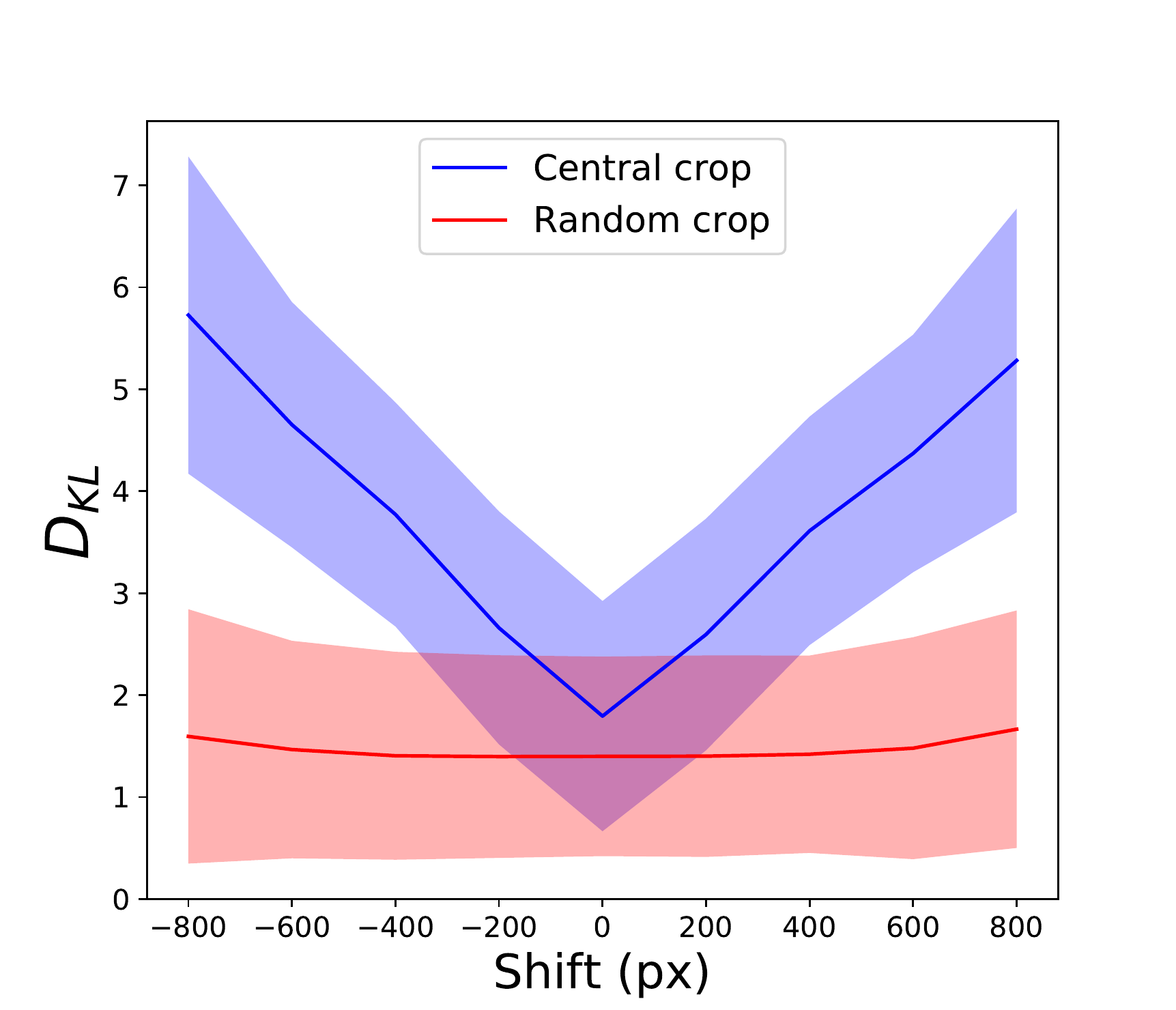}
    \caption{Performances of the \texttt{multi-branch} model trained with random and central crop policies, in presence of horizontal shifts in input clips. Colored background regions highlight the area within the standard deviation of the measured divergence. Recall that a lower value of $D_{KL}$ means a more accurate prediction.}
    \label{fig:shifts}
\end{figure}
\section{The effect of padded convolutions in learning a central bias}
\label{sup:padding}
Learning a globally localized solution seems theoretically impossible when using a fully convolutional architecture. Indeed, convolutional kernels are applied uniformly along all spatial dimension of the input feature map. Conversely, a globally localized solution requires knowing where kernels are applied during convolutions. We argue that a convolutional element can know its absolute position if there are latent statistics contained in the activations of the previous layer. 
In what follows, we show how the common habit of padding feature maps before feeding them to convolutional layers, in order to maintain borders, is an underestimated source of spatially localized statistics. Indeed, padded values are always constant, and unrelated to the input feature map. Thus, a convolutional operation, depending on its receptive field, can localize itself in the feature map by looking for statistics biased by the padding values. 
To validate this claim, we design a toy experiment in which a fully convolutional neural network is tasked to regress a white central square on a black background, when provided with a uniform or a noisy input map (in both cases, the target is independent from the input). We position the square (bias element) at the center of the output as it is the furthest position from borders, \emph{i.e.} where the bias originates.
We perform the experiment with several networks featuring the same number of parameters yet different receptive fields\footnote{a simple way to decrease the receptive field without changing the number of parameters is to replace two convolutional layers featuring $C$ output channels into a single one featuring $2C$ output channels.}. Moreover, to advocate for the random cropping strategy employed in the training phase of our network (recall that it was introduced to prevent a saliency-branch to regress a central bias), we repeat each experiment employing such strategy during training. 
All models were trained to minimize the mean squared error between target maps and predicted maps, by means of Adam optimizer\footnote{the code to reproduce this experiment is publicly released at \url{https://github.com/DavideA/can_i_learn_central_bias/}.}.
The outcome of such experiments, in terms of regressed prediction maps and loss function value at convergence, are reported in Fig.~\ref{fig:crops_uniform} and Fig.~\ref{fig:crops_noise}.
As shown by the top-most region of both figures, despite the uncorrelation between input and target all models can learn a central biased map. Moreover, the receptive field plays a crucial role, as it controls the amount of pixels able to \quotes{localize themselves} within the predicted map. As the receptive field of the network increases, the responsive area shrinks to the groundtruth area, and loss value lowers reaching zero. For an intuition of why this is the case, we refer the reader to Fig.~\ref{fig:receptive_field}.
Conversely,  as clearly emerges from the bottom-most region of both figures, random cropping prevents the model to regress a biased map, regardless the receptive field of the network. The reason underlying this phenomenon is that padding is applied after the crop, so its relative position with respect to the target depends on the crop location, which is random.
\begin{figure*}[tb]
\centering
\begin{tabular}{cc}
%% UNIFORM
\hline
\multicolumn{2}{c}{TRAINING WITHOUT RANDOM CROPPING}\\\hline\noalign{\vskip 2mm}
\begin{tabular}{ccccc}
Input & Target & RF=114 & RF=106 & RF=98\\ 
\includegraphics[width=0.07\textwidth]{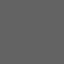}&
\includegraphics[width=0.07\textwidth]{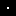}&
\includegraphics[width=0.07\textwidth]{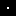}&
\includegraphics[width=0.07\textwidth]{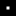}&
\includegraphics[width=0.07\textwidth]{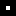}\\
\end{tabular}&
\begin{tabular}{c}
\includegraphics[width=0.45\textwidth]{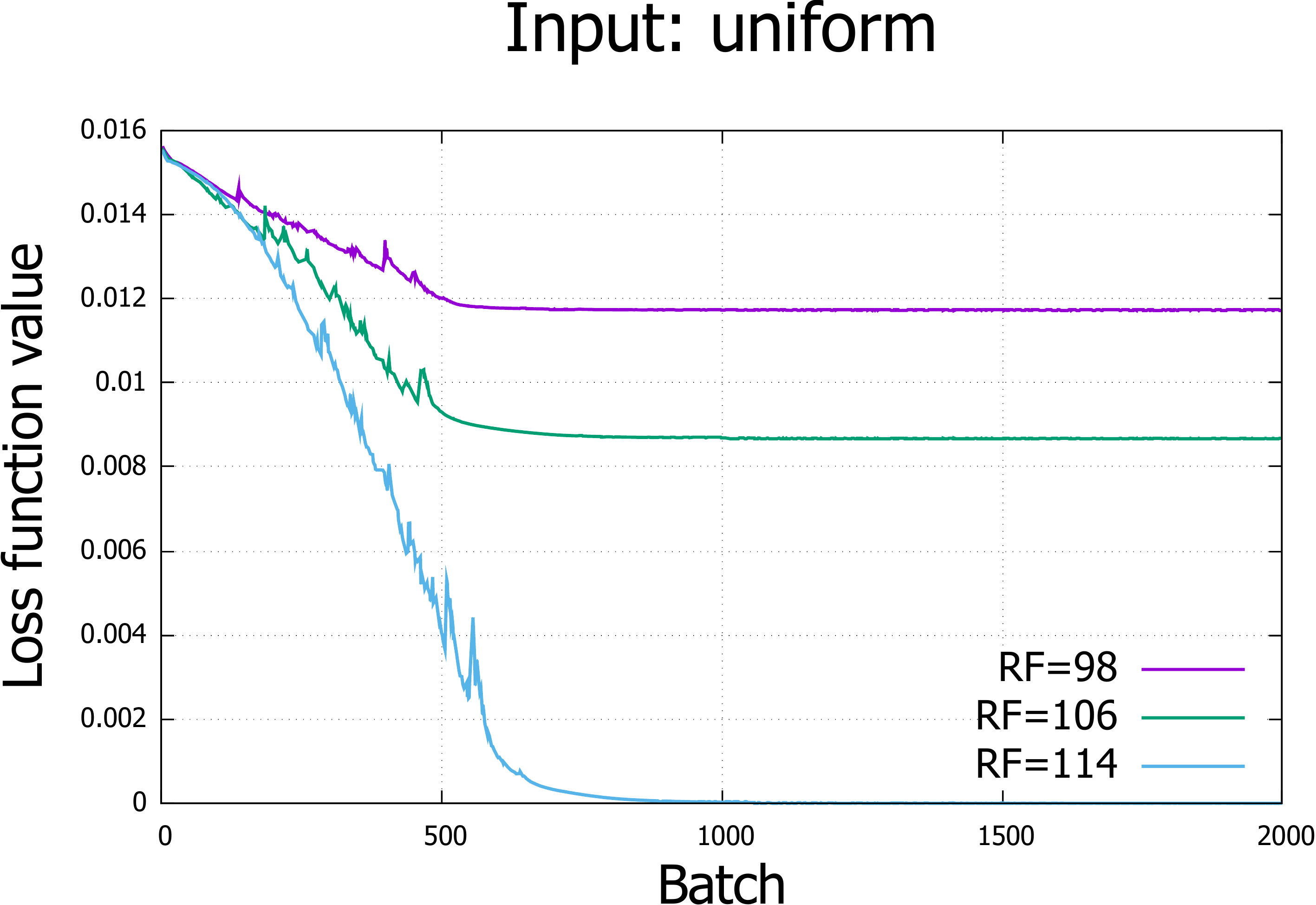}
\end{tabular}\\
\hline
\multicolumn{2}{c}{TRAINING WITH RANDOM CROPPING}\\\hline\noalign{\vskip 2mm}
\begin{tabular}{ccccc}
Input & Target & RF=114 & RF=106 & RF=98\\ 
\includegraphics[width=0.07\textwidth]{legacy/rebuttal_letter/img/bias_padding_img/outputs/input_uniform.png}&
\includegraphics[width=0.07\textwidth]{legacy/rebuttal_letter/img/bias_padding_img/outputs/target.png}&
\includegraphics[width=0.07\textwidth]{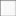}&
\includegraphics[width=0.07\textwidth]{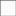}&
\includegraphics[width=0.07\textwidth]{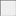}\\
\end{tabular}&
\begin{tabular}{c}
\includegraphics[width=0.45\textwidth]{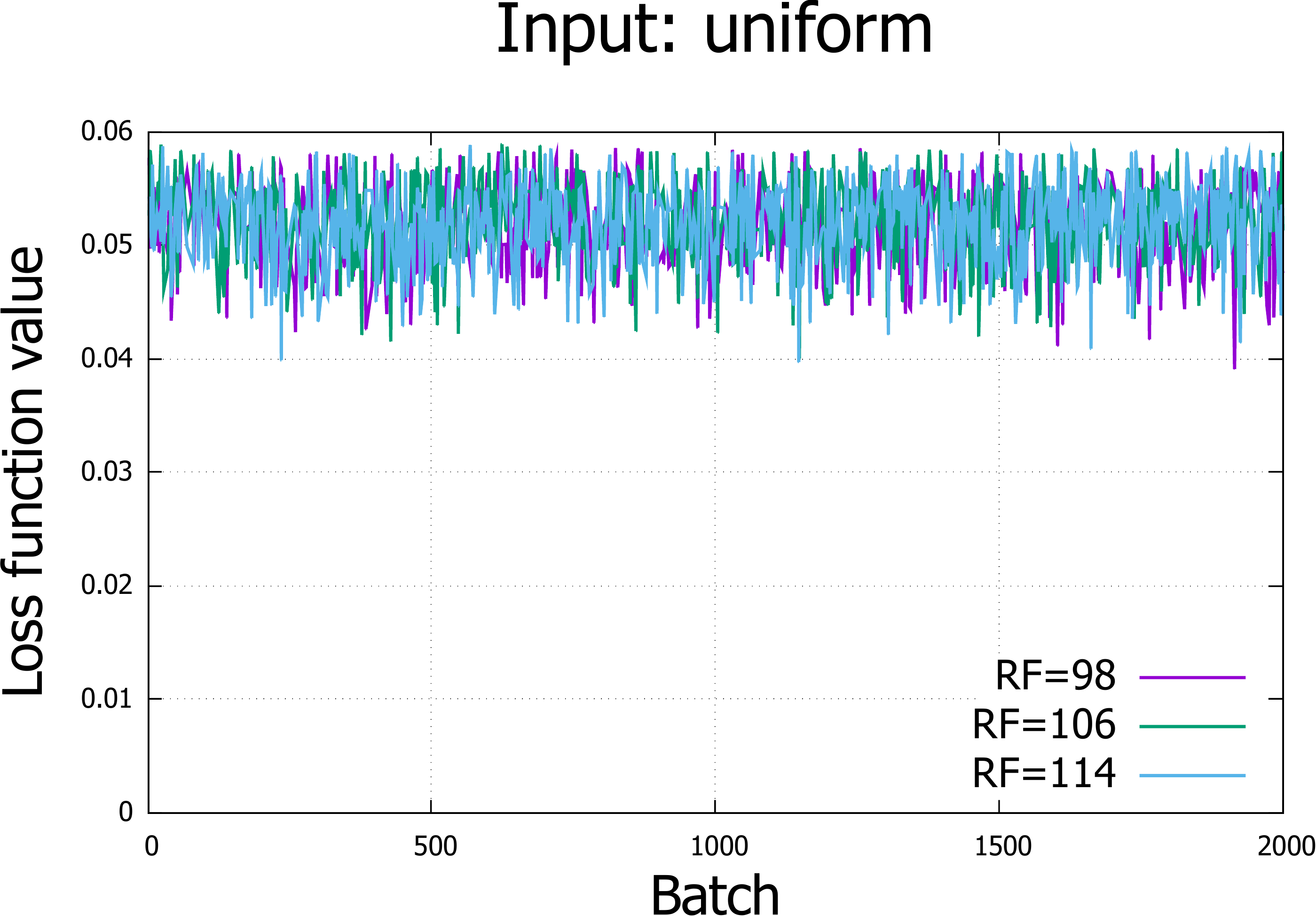}
\end{tabular}
\end{tabular}
\caption{To show the beneficial effect of random cropping in preventing a convolutional network to learn a biased map, we train several models to regress a fixed map from uniform input images. We argue that, in presence of padded convolutions and big receptive fields, the relative location of the groundtruth map with respect to image borders is fixed, and proper kernels can be learned to localize the output map. We report output solutions and loss functions for different receptive fields. As the receptive field grows, the portion of the image accessing borders grows and the solution improves (reaching lower loss values). Please note that in this setting the input image is 128x128, while the output map is 32x32 and the central bias is 4x4. Therefore, the minimum receptive field required to solve the task is $2*(32/2-4/2)*(128/32)=112$. Conversely, when trained by randomly cropping images and groundtruth maps, the reliable statistic (in this case, the relative location of the map with respect to image borders) ceases to exist, making the training process hopeless.}
\label{fig:crops_uniform}
\end{figure*}
\begin{figure*}[tb]
\centering
\begin{tabular}{cc}
%% UNIFORM
\hline
\multicolumn{2}{c}{TRAINING WITHOUT RANDOM CROPPING}\\\hline\noalign{\vskip 2mm}
\begin{tabular}{ccccc}
Input & Target & RF=114 & RF=106 & RF=98\\ 
\includegraphics[width=0.07\textwidth]{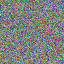}&
\includegraphics[width=0.07\textwidth]{legacy/rebuttal_letter/img/bias_padding_img/outputs/target.png}&
\includegraphics[width=0.07\textwidth]{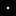}&
\includegraphics[width=0.07\textwidth]{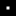}&
\includegraphics[width=0.07\textwidth]{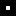}\\
\end{tabular}&
\begin{tabular}{c}
\includegraphics[width=0.45\textwidth]{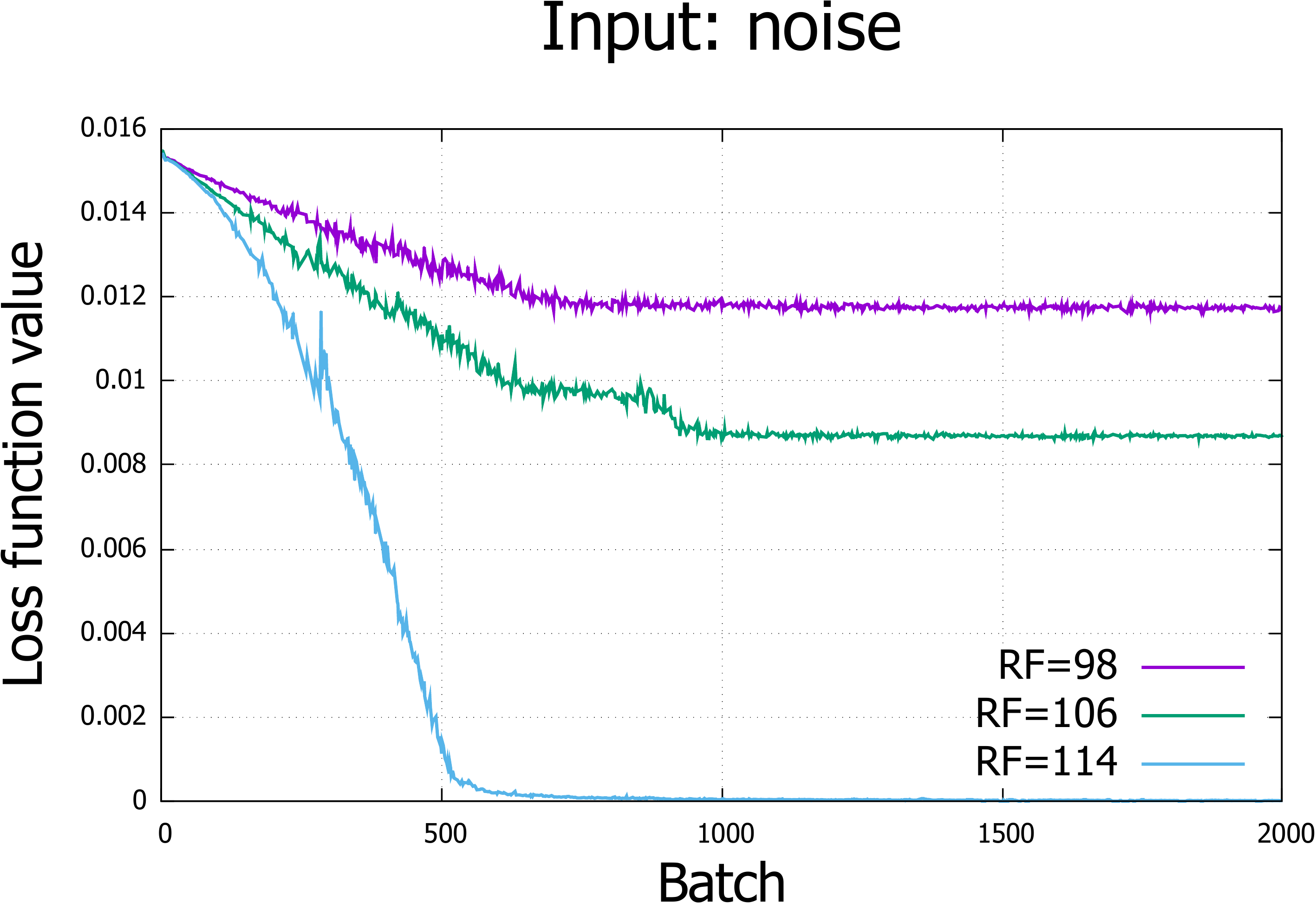}
\end{tabular}\\
\hline
\multicolumn{2}{c}{TRAINING WITH RANDOM CROPPING}\\\hline\noalign{\vskip 2mm}
\begin{tabular}{ccccc}
Input & Target & RF=114 & RF=106 & RF=98\\ 
\includegraphics[width=0.07\textwidth]{legacy/rebuttal_letter/img/bias_padding_img/outputs/input_noise.png}&
\includegraphics[width=0.07\textwidth]{legacy/rebuttal_letter/img/bias_padding_img/outputs/target.png}&
\includegraphics[width=0.07\textwidth]{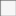}&
\includegraphics[width=0.07\textwidth]{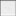}&
\includegraphics[width=0.07\textwidth]{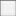}\\
\end{tabular}&
\begin{tabular}{c}
\includegraphics[width=0.45\textwidth]{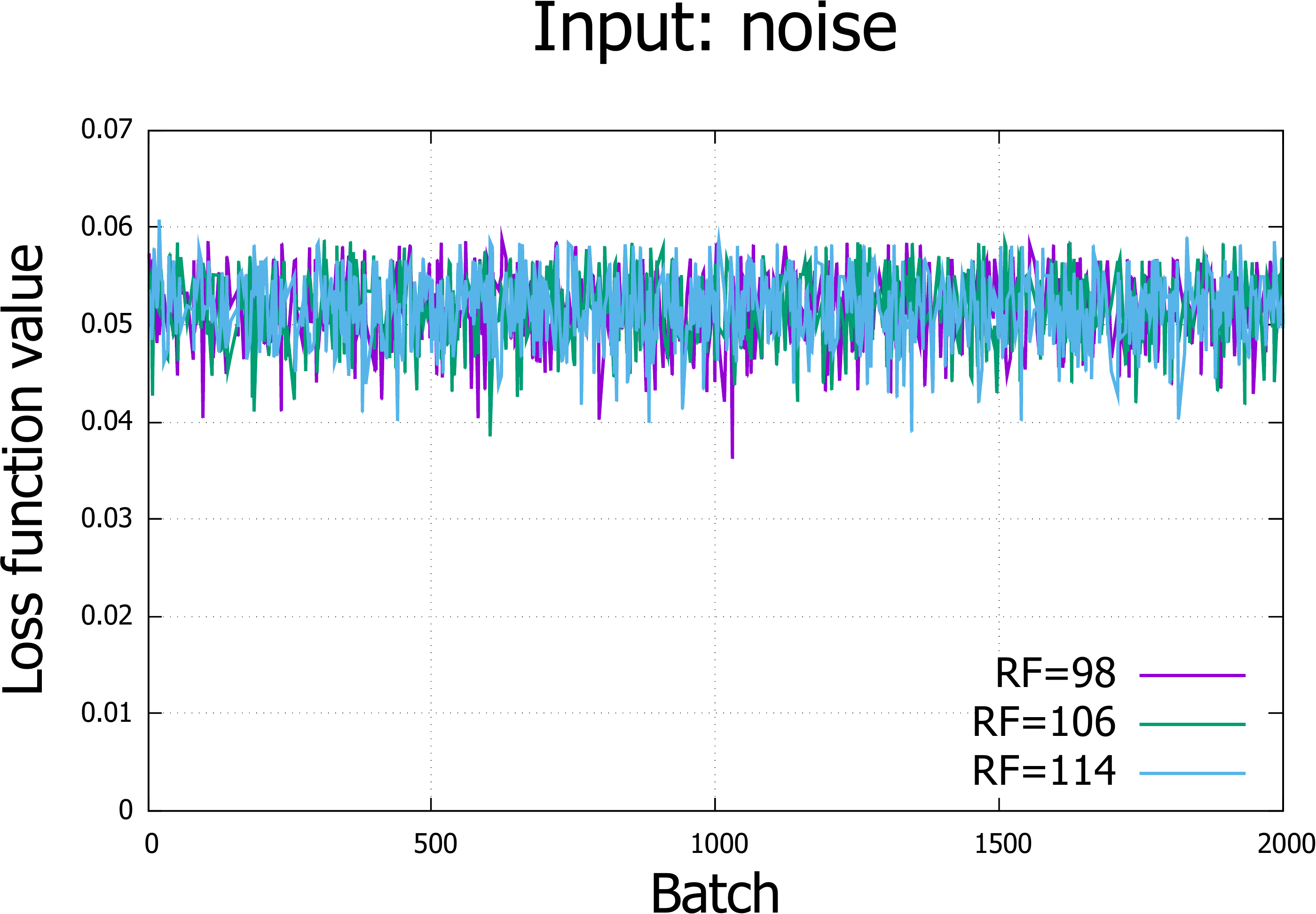}
\end{tabular}
\end{tabular}
\caption{This figure illustrates the same content as Fig.~\ref{fig:crops_uniform}, but reports output solutions and loss functions obtained from noisy inputs. See Fig.~\ref{fig:crops_uniform} for further details.}
\label{fig:crops_noise}
\end{figure*}
\begin{figure*}[bt]
\centering
\includegraphics[width=0.7\textwidth]{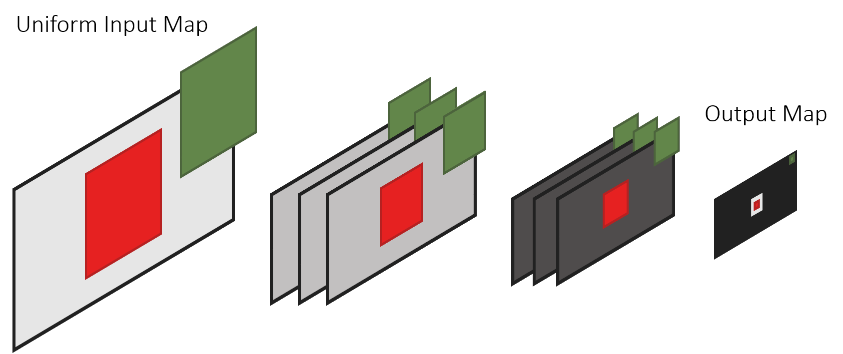}
\caption{Importance of the receptive field in the task of regressing a central bias exploiting padding: the receptive field of the red pixel in the output map has no access to padding statistics, so it will be activated. On the contrary, the green pixel's receptive field exceeds image borders: in this case, padding is a reliable anchor to break the uniformity of feature maps.}
\label{fig:receptive_field}
\end{figure*}
\section{On Fig.~\ref{fig:what_gt}}
We represent in Fig.~\ref{fig:thresholds} the process employed for building the histogram in Fig.\ref{fig:what_gt} (in the paper). Given a segmentation map of a frame and the corresponding ground-truth fixation map, we collect pixel classes within the area of fixation thresholded at different levels: as the threshold increases, the area shrinks to the real fixation point. A better visualization of the process can be found at \url{https://ndrplz.github.io/dreyeve/}.
\bgroup
\setlength{\tabcolsep}{.16667em}
\begin{figure*}[btp]
\centering
\begin{tabular}{cc}
image & segmentation\\
\includegraphics[width=0.4\textwidth]{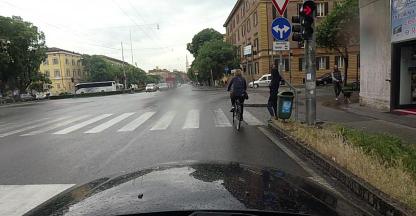}&
\includegraphics[width=0.4\textwidth]{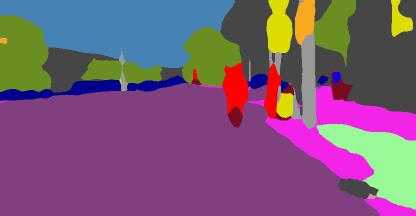}
\end{tabular}
\begin{tabular}{ccc}
\multicolumn{3}{c}{different fixation map thresholds}\\
\includegraphics[width=0.27\textwidth]{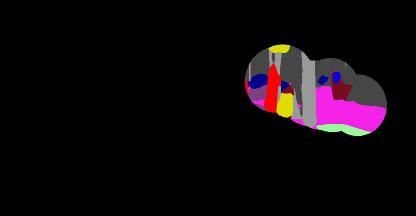}&
\includegraphics[width=0.27\textwidth]{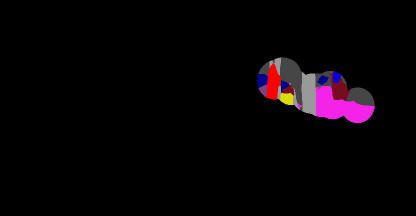}&
\includegraphics[width=0.27\textwidth]{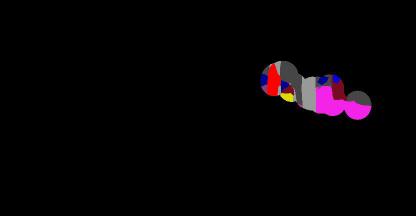}\\
\includegraphics[width=0.27\textwidth]{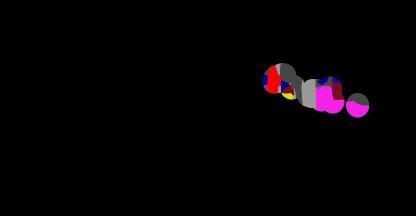}&
\includegraphics[width=0.27\textwidth]{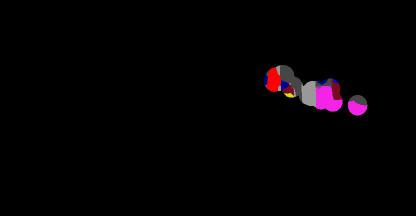}&
\includegraphics[width=0.27\textwidth]{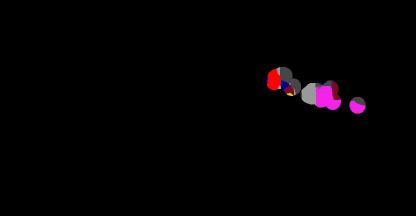}\\
\includegraphics[width=0.27\textwidth]{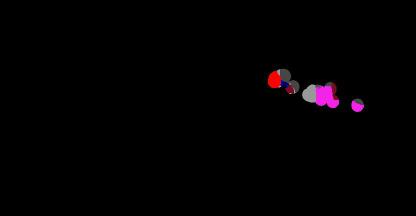}&
\includegraphics[width=0.27\textwidth]{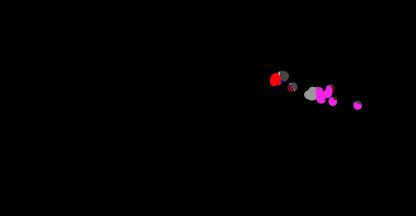}&
\includegraphics[width=0.27\textwidth]{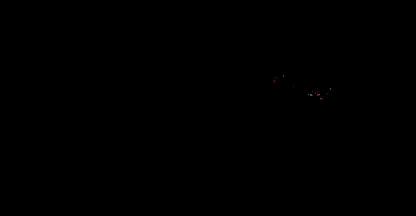}
\end{tabular}
\caption{Representation of the process employed to count class occurrences to build Fig.~\ref{fig:what_gt}. See text and paper for more details.}
\label{fig:thresholds}
\end{figure*}
\egroup
\section{Failure cases}
In Fig.~\ref{fig:sup:failure_cases} we report several clips in which our architecture fails to capture the groundtruth human attention.
\begin{figure*}[htp]
\centering
\resizebox{\textwidth}{!}{% <------ Don't forget this %
\begin{tabular}{ccc}
\textbf{Input frame} & \textbf{GT} & \textbf{\texttt{multi-branch}}\\
\includegraphics[width=0.33\textwidth]{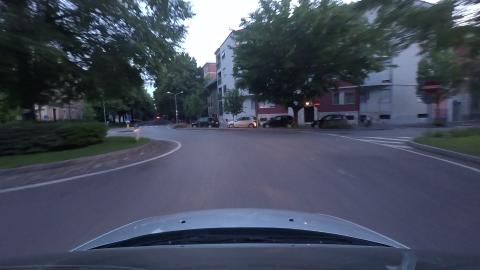}& \includegraphics[width=0.33\textwidth]{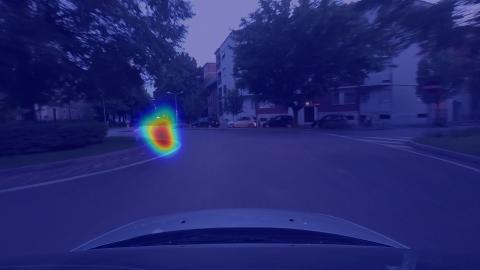}&
\includegraphics[width=0.33\textwidth]{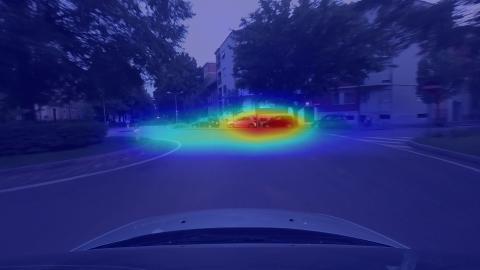}\\
\includegraphics[width=0.33\textwidth]{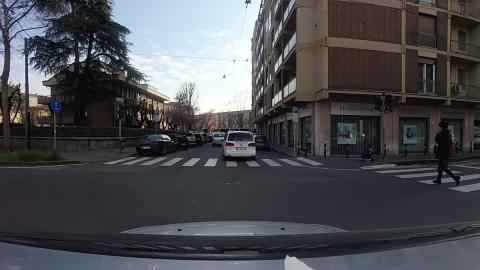}& \includegraphics[width=0.33\textwidth]{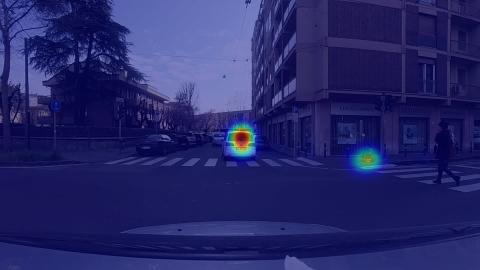}&
\includegraphics[width=0.33\textwidth]{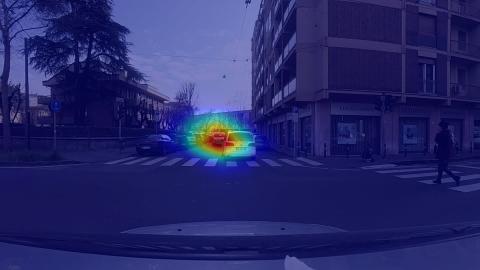}\\
\includegraphics[width=0.33\textwidth]{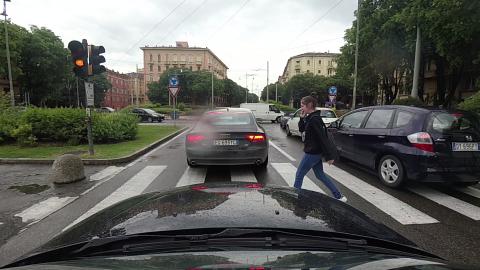}& \includegraphics[width=0.33\textwidth]{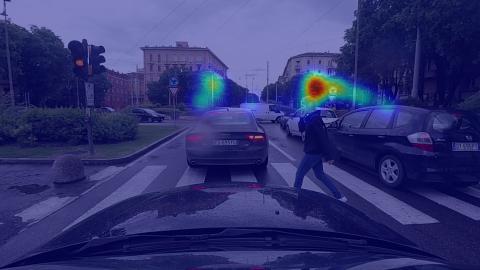}&
\includegraphics[width=0.33\textwidth]{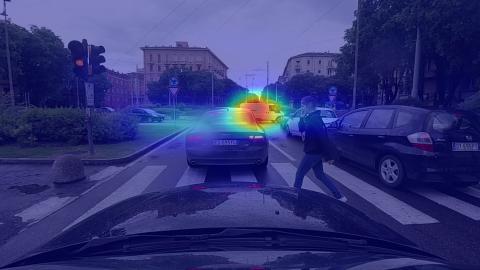}\\
\includegraphics[width=0.33\textwidth]{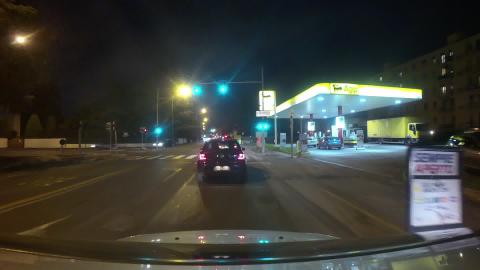}& \includegraphics[width=0.33\textwidth]{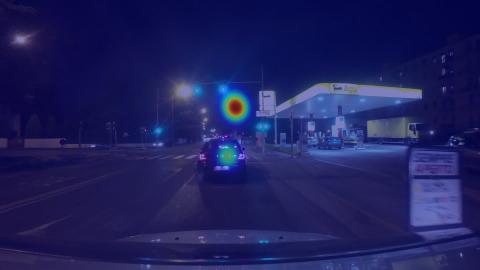}&
\includegraphics[width=0.33\textwidth]{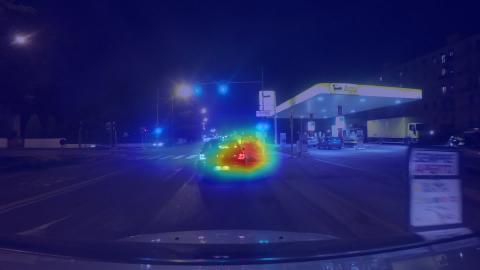}\\
\includegraphics[width=0.33\textwidth]{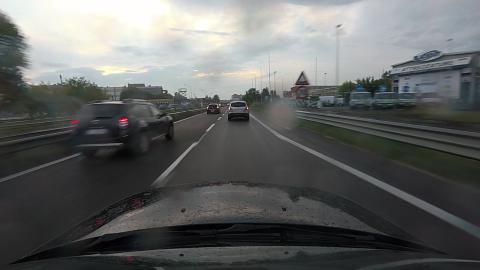}& \includegraphics[width=0.33\textwidth]{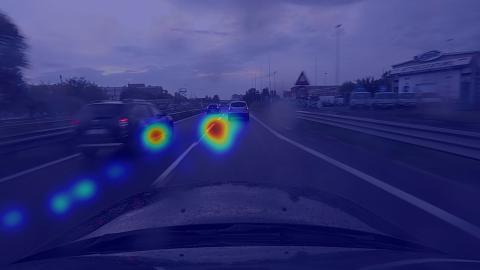}&
\includegraphics[width=0.33\textwidth]{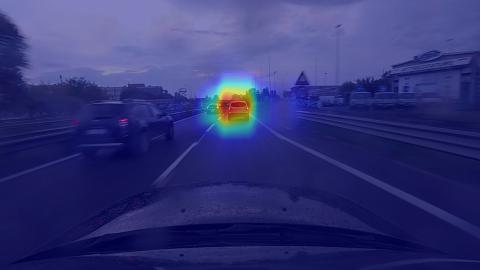}
\end{tabular}}
\caption{Some failure cases of our \texttt{multi-branch} architecture.}
\label{fig:sup:failure_cases}
\end{figure*}

\end{document}